\documentclass[10pt,twocolumn,letterpaper]{article}

\usepackage[pagenumbers]{cvpr} 

\usepackage{graphicx}
\usepackage{amsmath}
\usepackage{amssymb}
\usepackage{booktabs}
\usepackage{stfloats}
\usepackage{multirow}
\usepackage{multicol}
\usepackage{flushend}
\usepackage[export]{adjustbox}

\usepackage[pagebackref,breaklinks,colorlinks]{hyperref}
\usepackage{blindtext}

\usepackage[capitalize]{cleveref}
\crefname{section}{Sec.}{Secs.}
\Crefname{section}{Section}{Sections}
\Crefname{table}{Table}{Tables}
\crefname{table}{Tab.}{Tabs.} 

\begin{document}

\title{Decoupled Cross-Scale Cross-View Interaction for Stereo Image  \\ Enhancement in The Dark}

\author{Huan Zheng$^1$, Zhao Zhang$^{1*}$, Jicong Fan$^2$, Richang Hong$^1$, Yi Yang$^3$, and Shuicheng Yan$^4$\vspace{2mm}\\
	$^1$Hefei University of Technology, China\\
	$^2$The Chinese University of Hong Kong (Shenzhen), China\\
	$^3$Zhejiang University, China\\
	$^4$Sea AI Lab, Singapore
}

\twocolumn[{
	\renewcommand\twocolumn[1][]{#1}
	\maketitle
	\begin{center}
		\captionsetup{type=figure}
		\vspace{-8mm}
\begin{minipage}{1.01\linewidth}
	\centering
			\rotatebox{90}{\scriptsize{~~~~~~~~~~~~RGB Image}}
			\includegraphics[width=0.09\linewidth,height=0.15\linewidth, frame]{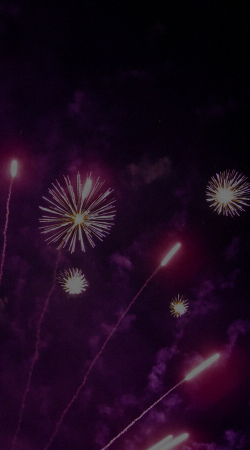}
			\includegraphics[width=0.09\linewidth,height=0.15\linewidth, frame]{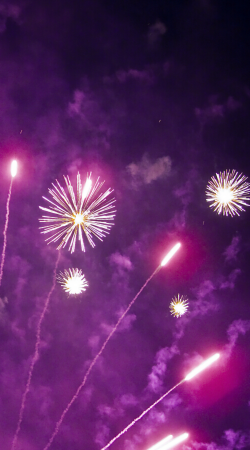}
			\includegraphics[width=0.09\linewidth,height=0.15\linewidth, frame]{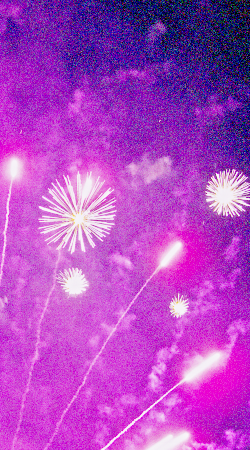}
			\includegraphics[width=0.09\linewidth,height=0.15\linewidth, frame]{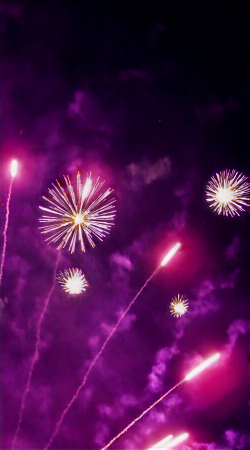}
			\includegraphics[width=0.09\linewidth,height=0.15\linewidth, frame]{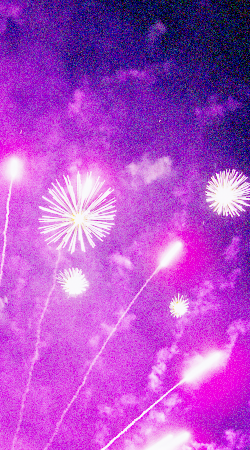}
			\includegraphics[width=0.09\linewidth,height=0.15\linewidth, frame]{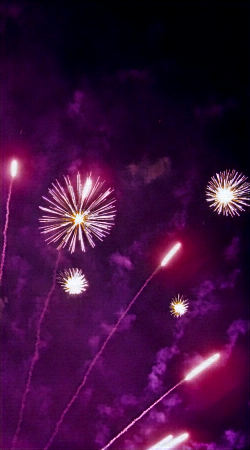}
			\includegraphics[width=0.09\linewidth,height=0.15\linewidth, frame]{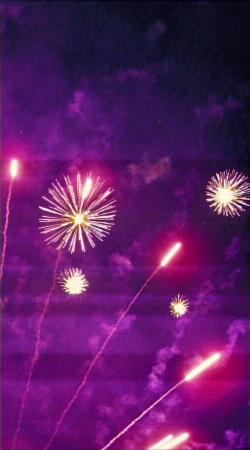}
			\includegraphics[width=0.09\linewidth,height=0.15\linewidth, frame]{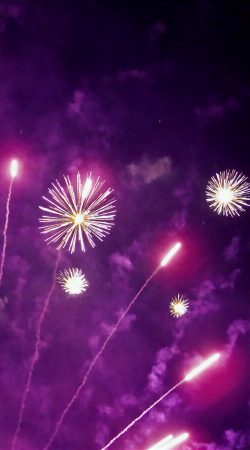}
			\includegraphics[width=0.09\linewidth,height=0.15\linewidth, frame]{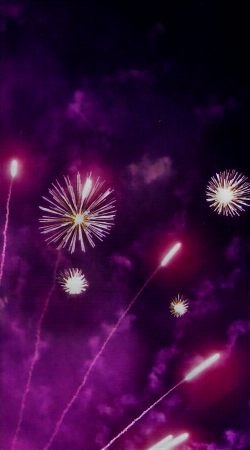}
			\includegraphics[width=0.09\linewidth,height=0.15\linewidth, frame]{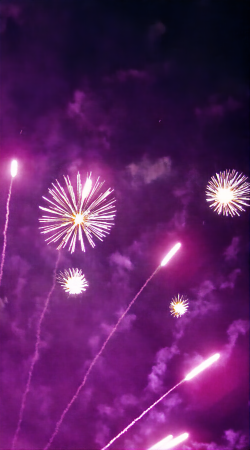}\\
			\rotatebox{90}{\scriptsize{~~~~~~~~~~~~Error Map}}
			\includegraphics[width=0.09\linewidth,height=0.15\linewidth, frame]{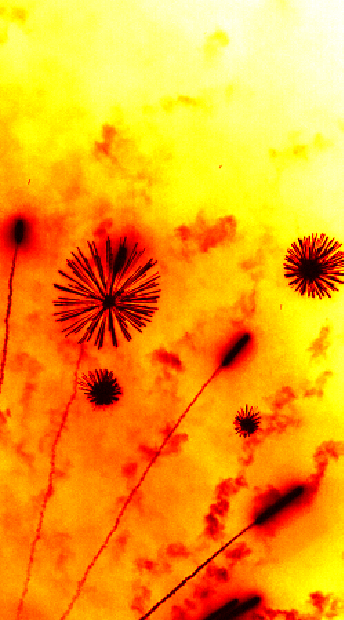}
			\includegraphics[width=0.09\linewidth,height=0.15\linewidth, frame]{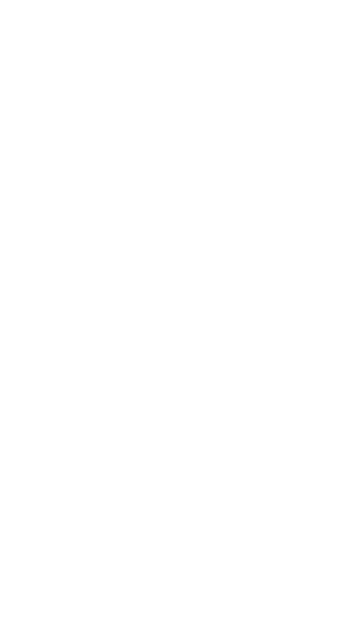}
			\includegraphics[width=0.09\linewidth,height=0.15\linewidth, frame]{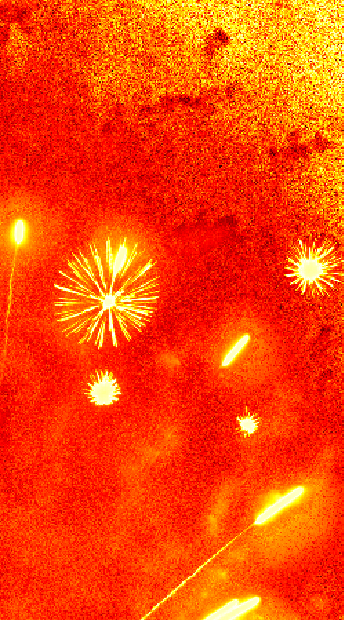}
			\includegraphics[width=0.09\linewidth,height=0.15\linewidth, frame]{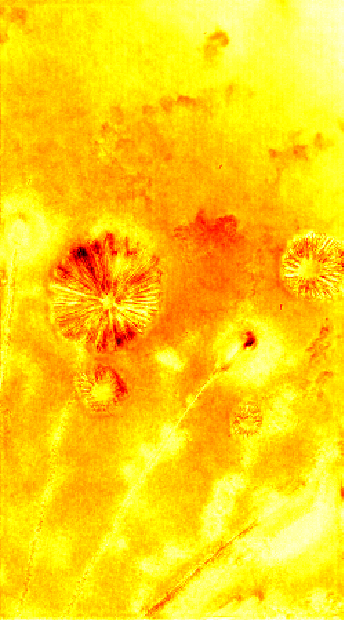}
			\includegraphics[width=0.09\linewidth,height=0.15\linewidth, frame]{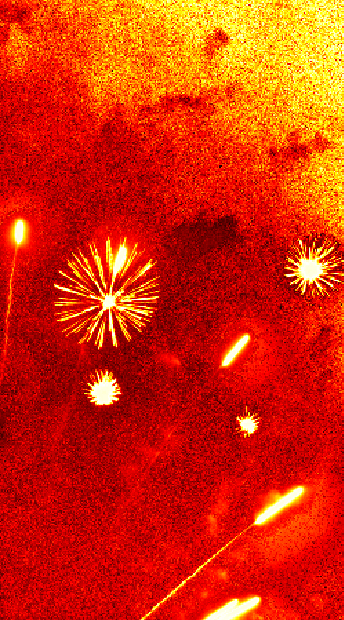}
			\includegraphics[width=0.09\linewidth,height=0.15\linewidth, frame]{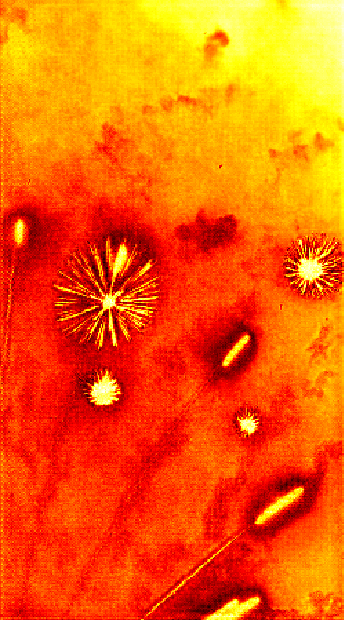}
			\includegraphics[width=0.09\linewidth,height=0.15\linewidth, frame]{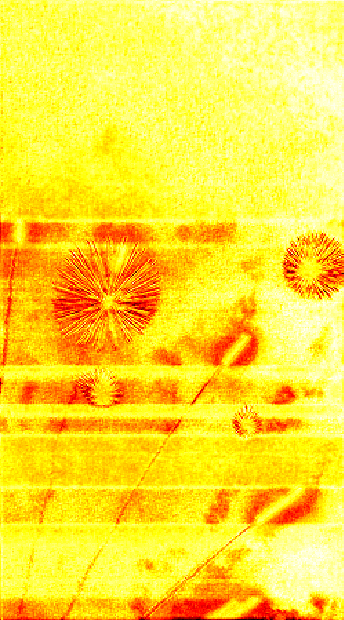}
			\includegraphics[width=0.09\linewidth,height=0.15\linewidth, frame]{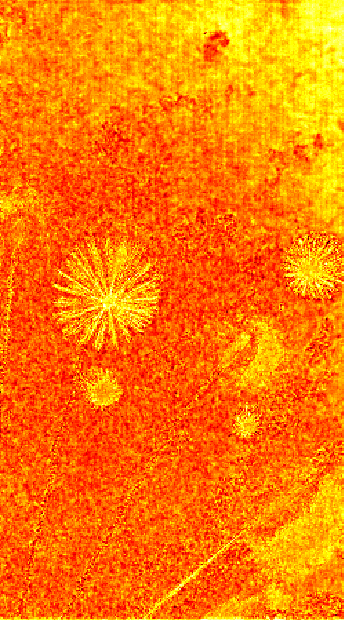}
			\includegraphics[width=0.09\linewidth,height=0.15\linewidth, frame]{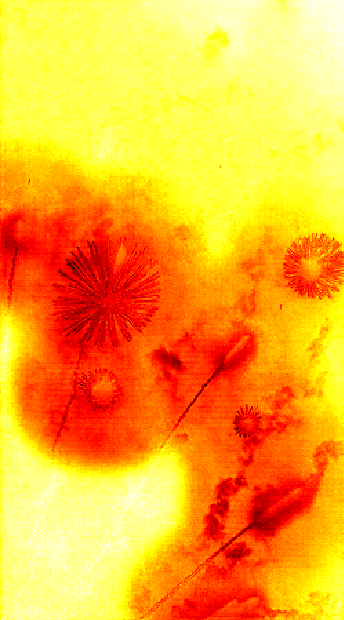}
			\includegraphics[width=0.09\linewidth,height=0.15\linewidth, frame]{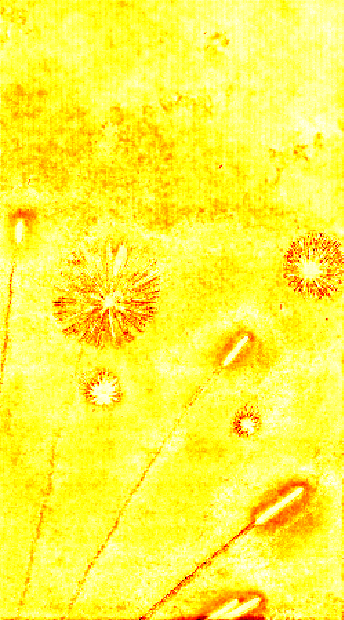}
		\end{minipage}\\
		\begin{minipage}{1\linewidth}
			\vspace{2pt}
			\hspace{4mm}
			\begin{minipage}{0.09\linewidth} \centering \footnotesize Input\\(PSNR/SSIM) \end{minipage}\hspace{1pt}
			\begin{minipage}{0.09\linewidth} \centering \footnotesize GT\\($+\infty$/1.000) \end{minipage}\hspace{1pt}
			\begin{minipage}{0.09\linewidth} \centering \footnotesize ZeroDCE\\(10.19/0.409) \end{minipage}\hspace{1pt}
			\begin{minipage}{0.09\linewidth} \centering	\footnotesize iPASSR\\(20.12/0.634) \end{minipage}\hspace{1pt}
			\begin{minipage}{0.09\linewidth} \centering	\footnotesize ZeroDCE++\\(9.71/0.385) \end{minipage}\hspace{1pt}
			\begin{minipage}{0.09\linewidth} \centering \footnotesize DVENet\\(16.97/0.557) \end{minipage}\hspace{1pt}
			\begin{minipage}{0.09\linewidth} \centering \footnotesize NAFSSR\\(24.35/0.825) \end{minipage}\hspace{1pt}
			\begin{minipage}{0.09\linewidth} \centering	\footnotesize NAFSSR-F\\(27.46/0.852) \end{minipage}\hspace{1pt}
			\begin{minipage}{0.09\linewidth} \centering	\footnotesize SNR\\(15.59/0.537) \end{minipage}\hspace{1pt}
			\begin{minipage}{0.09\linewidth} \centering	\footnotesize \textbf{DCI-Net}\\(31.33/0.937) \end{minipage}
		\end{minipage}
		\vspace{-3.5mm}
		\captionof{figure}{Visualization of the enhanced images and the corresponding error maps of each method based on Flickr1024 dataset, including NAFSSR \cite{Chu2022NAFSSRSI}, NAFSSR-F \cite{Chu2022NAFSSRSI}, iPASSRNet \cite{Wang2021SymmetricPA}, SNR \cite{Xu2022SNRAwareLI}, DVENet \cite{Huang2022LowLightSI}, ZeroDCE \cite{guo2020zero}, ZeroDCE++ \cite{li2021learning} and our DCI-Net. Whiter and brighter pixels in the error maps indicate smaller errors. It is clear that other compared methods obtain darker pixels in the error maps than our DCI-Net, which means that our method is capable of preserving consistent color and recovering the textures more accurately.}
		\label{fig:1}
	\end{center}
}]

\begin{abstract}
\vspace{-3.5mm}
Low-light stereo image enhancement (LLSIE) is a relatively new task to enhance the quality of visually unpleasant stereo images captured in dark condition. However, current methods achieve inferior performance on detail recovery and illumination adjustment. We find it is because: 1) the insufficient single-scale inter-view interaction makes the cross-view cues unable to be fully exploited; 
2) lacking long-range dependency leads to the inability to deal with the spatial long-range effects caused by illumination degradation.
To alleviate such limitations, we propose a LLSIE model termed Decoupled Cross-scale Cross-view Interaction Network (DCI-Net). Specifically, we present a decoupled interaction module (DIM) that aims for sufficient dual-view information interaction. DIM decouples the dual-view information exchange into discovering multi-scale cross-view correlations and further exploring cross-scale information flow. 
Besides, we present a spatial-channel information mining block (SIMB) for intra-view feature extraction, and the benefits are twofold. One is the long-range dependency capture to build spatial long-range relationship, and the other is expanded channel information refinement that enhances information flow in channel dimension. Extensive experiments on Flickr1024, KITTI 2012, KITTI 2015 and Middlebury datasets show that our method obtains better illumination adjustment and detail recovery, and achieves SOTA performance compared to other related methods. 
Our codes, datasets and models will be publicly available. 

\end{abstract}

\vspace{-6mm}
\section{Introduction}
Single image processing and understanding have made great achievements across a wide range of application areas, such as image classification \cite{he2016deep}, object detection \cite{ren2015faster} and semantic segmentation \cite{long2015fully}. Recently, with the growing application of dual cameras, stereo vision has attracted much attention in various fields, e.g., mobile phones and autonomous driving cars \cite{li2019stereo}. However, the stereo images captured in dark environments usually suffer from low-contrast, weak illumination and various noise \cite{zhang2022deep}. As a consequence, there are obviously negative effects on subsequent high-vision tasks. Hence, low-light stereo image enhancement (LLSIE) is proposed to enhance dark stereo images \cite{huang2022low}. To be specific, LLSIE is a task of enhancing the illumination and recovering the hidden details in the dark, via utilizing the stereo images from left and right views.

\begin{figure*}[t]
	\centering  
	\includegraphics[width=0.88\linewidth]{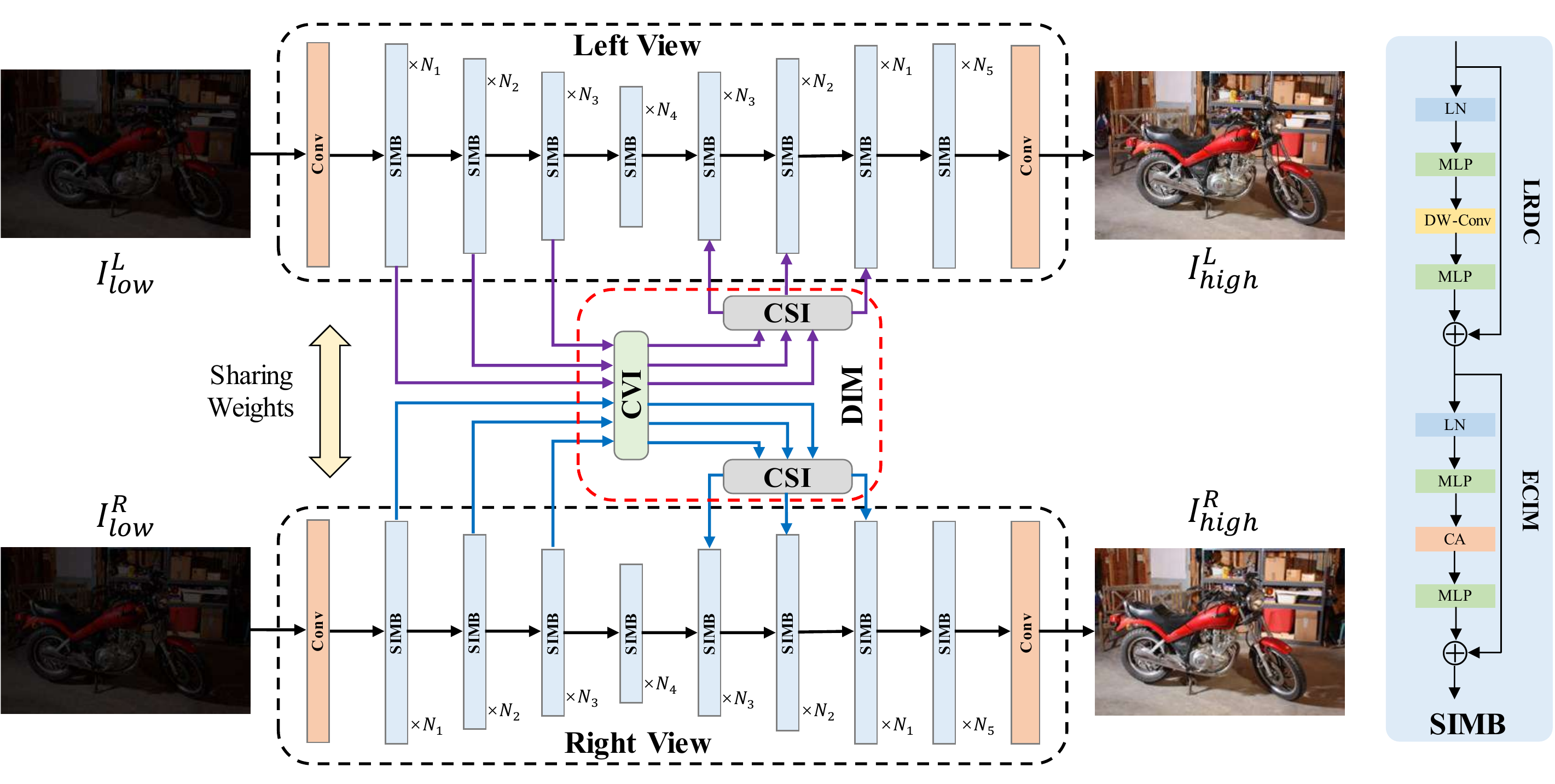}
	\vspace{-2mm} 
	\caption{The overall framework of our proposed DCI-Net for LLSIE, which contains two weights-shared branches to process left and right views respectively. Besides, DCI-Net includes two main modules, i.e., DIM and SIMB. To be specific, DIM completes sufficient inter-view information interaction and flow across two views, which includes cross-view interaction (CVI) and cross-scale interaction (CSI); SIMB enhances intra-view feature representation, whose structure is shown on the right. }
	\label{fig:2}
	\vspace{-3.5mm}
\end{figure*}

Compared with LLSIE, single low-light image enhancement (LLIE) methods aim to refine the illumination of single-view images in the dark, which can be divided into traditional and deep learning-based ones \cite{li2021low}. Traditional LLIE methods build prior-based optimization models to adjust the illumination and enhance the contrast \cite{guo2016lime}. While these methods are relatively simple or highly rely on the hand-crafted priors, which may cause low-quality enhanced results \cite{Liu2021BenchmarkingLI}. With the great development of deep learning, convolutional neural network (CNN)-based methods have achieved impressive performance in various low-level vision tasks \cite{zheng2022deep, Zhao2022FCLGANAL, Dong2016ImageSU, Feng2022MIPI2C, Wei2022SGINetTS, Zheng2021GCMNetTE}. More importantly, CNN-based deep models also show superior capability for LLIE \cite{wang2019underexposed, xu2020learning, liu2021retinex, wang2021low, wei2018deep, li2018lightennet, Chen2018LearningTS, zhang2022frc}. These deep LLIE methods use CNN as backbones to establish a neural network to learn a map from low-light image to normal-light image.

Stereo image restoration is the task of recovering high-quality stereo images from diverse degradations. In comparison to single image restoration, parallax in stereo image pairs is a key point for stereo image restoration. Some recent methods have been proposed to restore the lost information using the correlations between two views \cite{Huang2022LowLightSI, Wang2019LearningPA, Wang2021SymmetricPA, Chu2022NAFSSRSI}. iPASSRNet is firstly proposed for stereo image super-resolution via exploring symmetry cues between two views \cite{Wang2021SymmetricPA}. Inspired by iPASSRNet, DVENet is presented, which is the most representative method exploring the LLSIE task \cite{Huang2022LowLightSI}. However, the generated illumination-improved images are still unsatisfactory for both low-light enhancement and stereo image restoration methods, as shown in Fig. \ref{fig:1}. Therefore, we ask: \textbf{\textit{what makes the enhanced stereo images suffer from undesired and inaccurate contents}}? We attempt to answer this question from two respects: 

\begin{enumerate}
	\vspace{-2mm} 
	\item [(1)] \textbf{Insufficient dual-view information interaction}. The single LLIE methods do not consider the relationship between two views as stereo image at all. In contrast, stereo image restoration methods clearly need to exploit the cues between stereo image pairs. Nevertheless, current dual-view information interaction strategies are still weak. Because existing methods only explore the cross-view correlations at single scale, while ignoring the cues at different scales and missing cross-scale interaction, causing the inability of achieving sufficient cross-scale cross-view interaction. 
	\vspace{-2mm} 
	\item [(2)]  \textbf{Lack of long-range dependency in intra-view learning}. Current methods usually adopt CNNs with a kernel size of 3$\times$3 to build a neural network for image restoration. However, small kernel size design may limit the learning ability of CNNs, because the convolutional operation can only extract information from small regions, and prevents capturing long-range dependencies. Poor illumination has a great influence on the entire image. To handle the spatial long-range effects caused by the degradation, it is important to build the long-range relationship for LLSIE. 
	\vspace{-2mm} 
\end{enumerate}

In this paper, we therefore explore effective strategies to facilitate interaction and enhance the stereo images in the dark, and propose a decoupling strategy to complete cross-scale cross-view information interaction. The main contributions of this paper are summarized as follows: 
\begin{enumerate}
	\vspace{-2mm} 
	\item [(1)] \textbf{DCI-Net: LLSIE by Decoupled Cross-scale Cross-view Interaction Network}. We propose DCI-Net to address the issue of weak cross-view information interaction for LLSIE. Specifically, DCI-Net aims at improving the enhancement process by refining both intra-view feature extraction and cross-view information interaction. Experiments on Flickr1024, KITTI 2012, KITTI 2015 and Middlebury datasets demonstrate that our method can better adjust the illumination, recover the details and obtain SOTA performance. 
	
	\vspace{-2mm} 
	\item [(2)] \textbf{Decoupled Interaction Module (DIM)}. To enable sufficient dual-view information interaction, namely, cross-scale cross-view information interaction, DIM decouples the above process into two levels, i.e., cross-view interactions at multiple scales and further cross-scale interaction. The first level aims at discovering multi-scale cross-view cues, and the second level focuses on exploring cross-scale information flow for further interaction. Hence, DIM can make full use of the correlations between stereo image pairs.
	
	\vspace{-2mm} 	
	\item [(3)] \textbf{Spatial-channel Information Mining Block (SIMB)}. We design the novel module SIMB for intra-view feature extraction. To be specific, SIMB is based on the structure of vision transformer (ViT) so that it can possess a strong learning ability, but our core idea departs from ViT. Instead of using multi-head self-attention, large-kernel design is incorporated into the process of long-range dependency capture (LRDC) for discovering spatial long-range correlations. In addition, expanded channel information refinement (ECIR) is also developed for enhancing channel information flow.	
	\vspace{-2mm} 
\end{enumerate}

\section{Related Work}
We briefly review the recent progress on single low-light image enhancement and stereo image restoration. 

\subsection{Low-light single image enhancement}
For traditional LLIE methods, we mainly introduce the retinex-based and histogram equalization (HE)-based ones. Inspired by the retinex theory \cite{land1977retinex}, retinex models are proposed to decompose the low-light images and reconstruct the normal-light images \cite{Ren2020LR3MRL, Li2018StructureRevealingLI, Cai2017AJI}. HE-based methods aim at adjusting the dynamic range of the low-light image to enhance the contrast \cite{Lee2013ContrastEB}. Deep LLIE methods can be further fallen into end-to-end and retinex-based modes. To be specific, end-to-end method directly learns a map to reverse the illumination degradation, which takes the low-light image as input and directly outputs the enhanced normal-light image \cite{ren2019low, yang2021band, li2021learning, guo2020zero, Xu2022SNRAwareLI, Ma2022TowardFF, Zheng2021AdaptiveUT}; Retinex-based deep methods decouple an image into the illumination map and reflectance map \cite{zhang2019kindling, Wu2022URetinexNetRD, liu2021retinex, wei2018deep}. Differently, deep retinex-based methods employ deep neural networks for image decomposition, in comparison to the traditional retinex-based methods.

\vspace{-2mm} 
	
\subsection{Stereo image restoration}
Recently, stereo vision has been attracting much attention. A few stereo image restoration methods are also studied. For stereo image super-resolution, Jeon et al. \cite{jeon2018enhancing} proposed the first work that uses the shift operation to compensate for the parallax between two views. Wang et al. \cite{Wang2021SymmetricPA} developed a novel parallax attention to make full use of the correlations between stereo image pairs. The latest state-of-the-art method is NAFSSR, which is the champion of the NTIRE 2022 Stereo Image Super-resolution Challenge \cite{Chu2022NAFSSRSI, Wang2022NTIRE2C}. For stereo image deraining, Zhang et al. \cite{Zhang2022BeyondMD} incorporated semantic priors into deraining process for better rain removal. For stereo debluring, Zhou et al. \cite{zhou2019davanet} presented a depth-aware and view aggregated method. Li et al. \cite{Li2022LearningDA} delivered a novel stereo image debluring model by exploring the dual-pixel alignment. For stereo image dehazing, Nie et al. \cite{Nie2022StereoRD} proposed SRDNet which aims at better exploiting the stereo information from two views. There are also few works for LLSIE \cite{Huang2022LowLightSI, liao4240548no, jung2020multi}. Specifically, DVENet \cite{Huang2022LowLightSI} is the most representative method, which incorporates retinex theory into the overall framework in a coarse-to-fine manner. Besides, parallax attention model is also used to explore the correlations between two views.

\section{Proposed Method}
We introduce the proposed DCI-Net in detail in this section. We first illustrate the overall architecture of DCI-Net. Then, we describe the detailed structures of the designed modules. Finally, the used loss functions are discussed.

\subsection{Overall framework}
An overview of the proposed DCI-Net is shown in Fig. \ref{fig:2}. Clearly, our model takes a pair of low-light stereo images as input, enhances the illumination of both views, and outputs the enhanced normal-light stereo images. The pipeline of DCI-Net can be divided into three stages: shallow feature extraction, deep feature extraction and stereo image reconstruction. To be specific, we use two convolutional layers in the head and tail, where the first one extracts shallow features and the last one reconstructs the enhanced normal-light stereo images. Given a pair of low-light stereo images, the above processes can be formulated as follows: 
\begin{equation}
	I_{high}^L, I_{high}^R = {\rm H_{SR}(H_{DF}(H_{SF}(}I_{low}^L, I_{low}^R{\rm )))},
\end{equation}
where $I_{low}^L$, $I_{low}^R$, $I_{high}^L$ and $I_{high}^R$ denote the low-light left-view image, low-light right-view image, enhanced left-view image and right-view image, ${\rm H_{SF}}(\cdot)$, ${\rm H_{DF}}(\cdot)$ and ${\rm H_{SR}}(\cdot)$ denote the transformations for shallow feature extraction, deep feature extraction and stereo image reconstruction respectively. Deep feature extraction can be further fallen into intra-view feature extraction and dual-view interaction. For dual-view interaction, we deliver a decoupled interaction module (DIM) to explore synchronous cross-view and cross-scale interaction. For intra-view feature extraction, we construct a spatial-channel information mining block (SIMB)-based U-Net to obtain stronger feature representation. It is worth noting that the weights of shallow feature extraction, intra-view feature extraction in deep feature extraction and stereo image reconstruction are always shared.

\subsection{Decoupled interaction module (DIM)}
Different from single image processing, one key point of LLSIE is exploring the correlations between two views to promote illumination enhancement. Hence, methods for low-light single image enhancement do not well in enhancing the stereo images, since they only consider one view. Some previous attempts have done to discover and exploit the cues between a pair of stereo images \cite{jeon2018enhancing, Chu2022NAFSSRSI, Wang2021SymmetricPA, Huang2022LowLightSI, wang2019learning}. Nevertheless, these methods lack considering the cross-view interaction at different scales, which makes the cross-scale interaction be ignored. Note that a lot of existing studies on CNN and ViT show the importance of multi-scale information interaction \cite{wang2020deep, gu2022multi}. To alleviate this issue, we propose DIM to decouple cross-scale cross-view information interaction into studying inter-view correlations at multiple scales, and further cross-scale interaction.

\begin{figure}[t]
	\centering
	\includegraphics[width=0.88\linewidth]{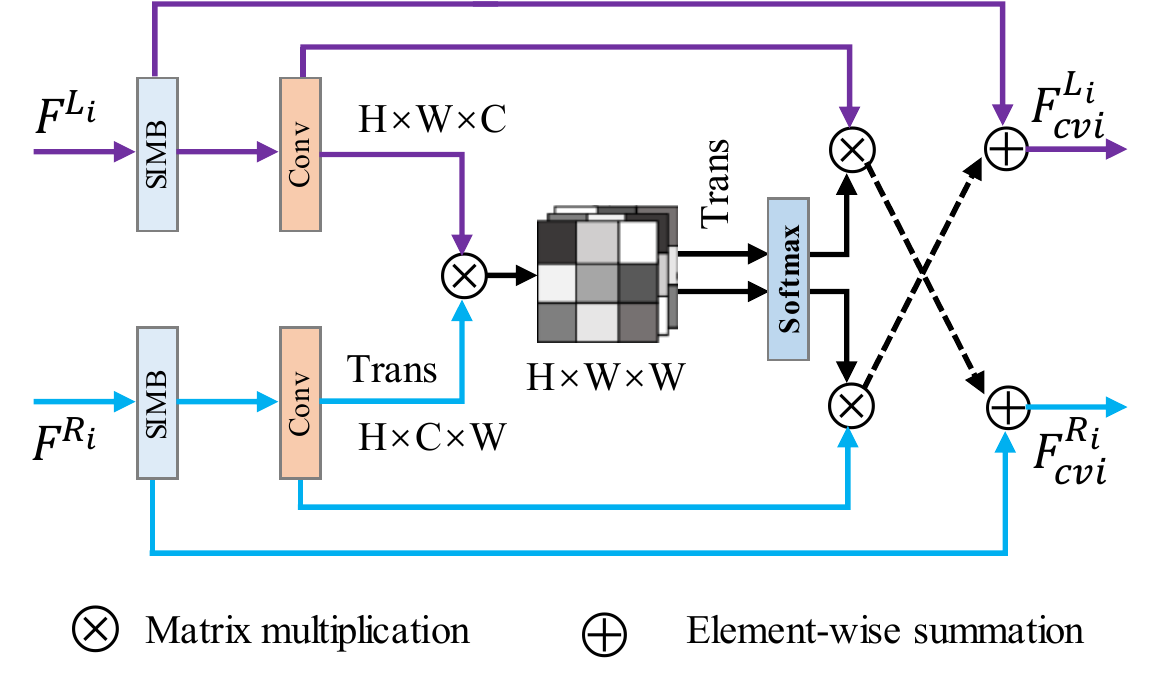}
	\vspace{-4mm} 
	\caption{The detailed process of cross-view interaction in DIM. To be specific, CVI explores the cues between both left and right views at multiple scales. Note that only the process at single scale is shown as example.}
	\label{fig:3}
	\vspace{-4mm} 
\end{figure}

\textbf{Cross-view interaction at multiple scales}. Previous methods only use the correlations at a single scale \cite{jeon2018enhancing, Chu2022NAFSSRSI, Wang2021SymmetricPA, Huang2022LowLightSI, wang2019learning}. In contrast, we explore cross-view cues at multiple scales. We incorporate a cross-view interaction (CVI) module into different scales for discovering the multi-scale cross-view cues. The detailed structure of CVI is shown in Fig. \ref{fig:3}. Given the input stereo feature maps from the $i$-th stage, CVI uses the matrix multiplication to compute the cross-view correlations for information interaction. This process can be formulated as follows:
\begin{equation}
	M_{cor} = LR^{T},
\end{equation}
where $L\in \mathbb{R}^{H\times W\times C}$ and $R\in \mathbb{R}^{H\times C\times W}$ denote the input stereo feature maps, and $M_{cor}\in \mathbb{R}^{H\times W\times W}$ denotes the correlations matrix. It is noted that there are only horizontal shifts for stereo images. Hence, we mainly pay attention to the horizontal correlations between two views.

\begin{figure}[t]
	\centering
	\includegraphics[width=0.9\linewidth]{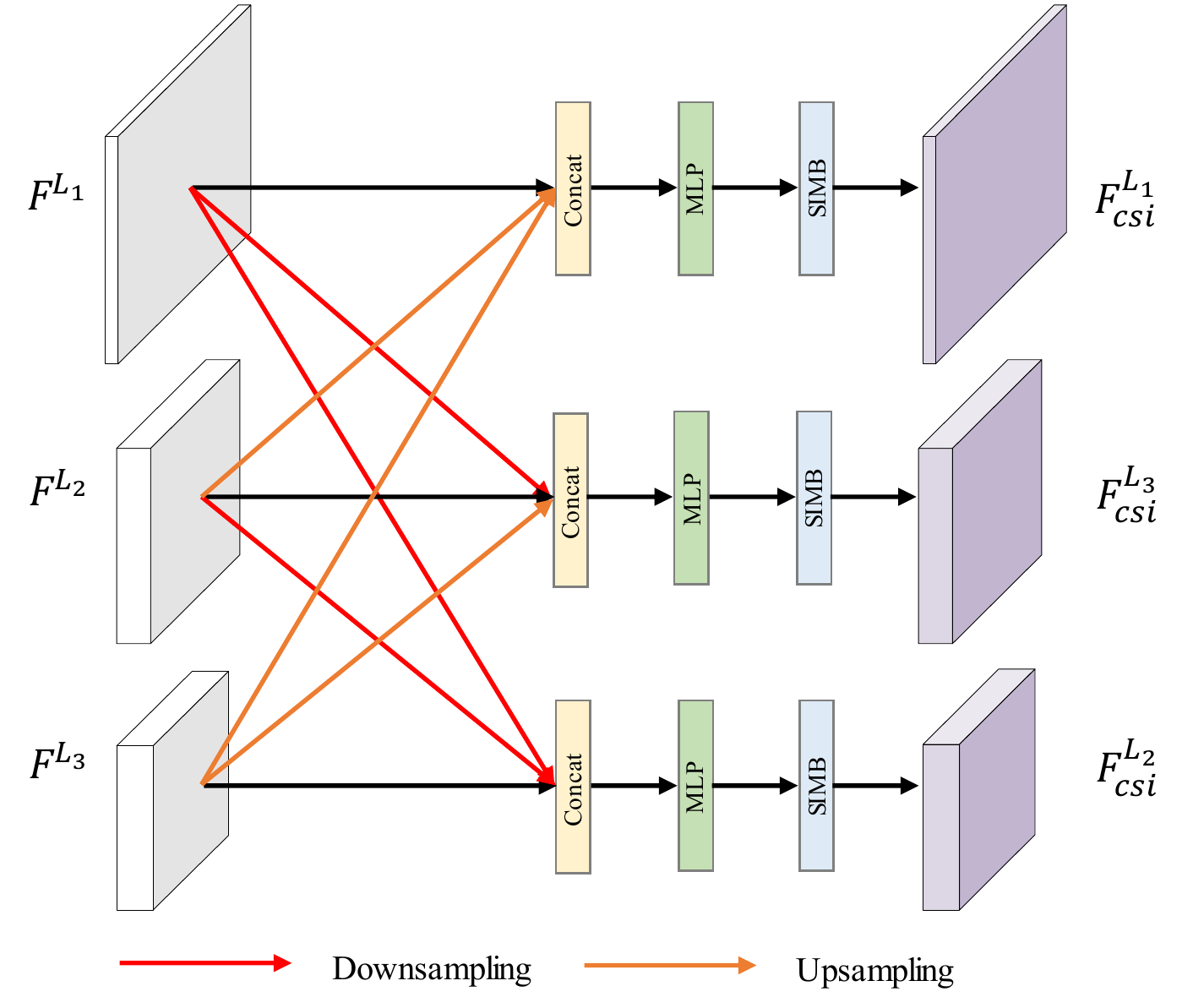}
	\vspace{-2mm} 
	\caption{The detailed process of cross-scale interaction in DIM. Specifically, CSI completes the interaction among different scales. Note that only the process of left view is shown.}
	\label{fig:4}
	\vspace{-3mm} 
\end{figure}

\textbf{Cross-scale interaction}. CVI has extracted the multi-scale correlations between two views, but the information interaction among different scales is still missing. Cross-scale interaction (CSI) is therefore presented to handle this issue by enhancing the cross-scale information flow. The overall structure of CSI is shown in Fig. \ref{fig:4}. Given the left-view multi-scale feature maps $F^{L_1}$, $F^{L_2}$ and $F^{L_3}$ obtained by CVI as an example, CSI firstly uses scaling operations to densely concatenate them, and then utilizes a multilayer perceptron (MLP) to process the concatenated feature maps. The purpose is twofold, one is to reduce channels and the other is to exchange and fuse information in channel dimension. In the end, a spatial-channel information mining block is incorporated into the tail for further spatial information interaction. The above processes can be formulated as
\begin{equation}
	\hspace{-0.6em}F_{csi}^{L_1}, F_{csi}^{L_2}, F_{csi}^{L_3} = {\rm SIMB(MLP(DC(}F^{L_1}, F^{L_2}, F^{L_3}{\rm )))}
\end{equation}
where $\rm DC(\cdot)$ denotes the densely concatenate operation in CSI, and $F_{csi}^{L_1}$, $F_{csi}^{L_2}$ and $F_{csi}^{L_3}$ denote the cross-scale interacted feature maps.

\begin{figure*}[t]
	\centering  
	\begin{minipage}{1\linewidth}
		\centering
		\rotatebox{90}{\scriptsize{~~~~~~~~RGB Image (\textcolor{magenta}{Left})}}
		\includegraphics[width=0.09\linewidth,height=0.15\linewidth, frame]{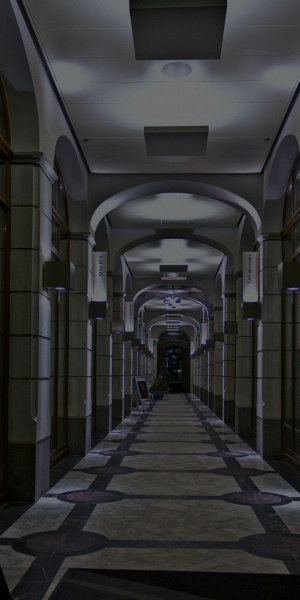}
		\includegraphics[width=0.09\linewidth,height=0.15\linewidth, frame]{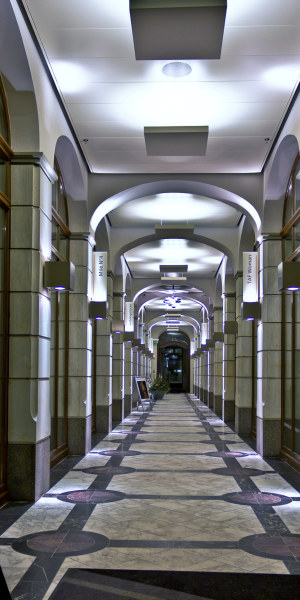}
		\includegraphics[width=0.09\linewidth,height=0.15\linewidth, frame]{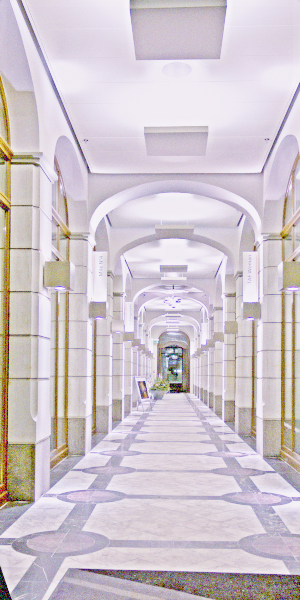}
		\includegraphics[width=0.09\linewidth,height=0.15\linewidth, frame]{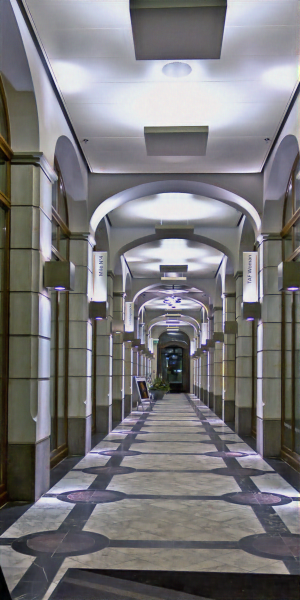}
		\includegraphics[width=0.09\linewidth,height=0.15\linewidth, frame]{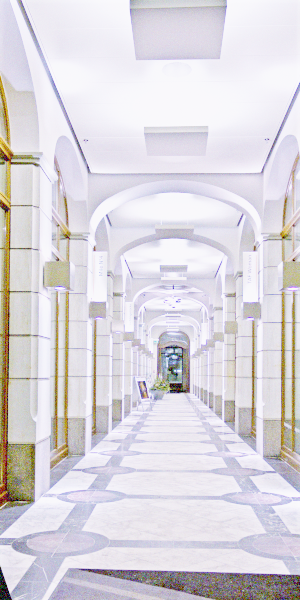}
		\includegraphics[width=0.09\linewidth,height=0.15\linewidth, frame]{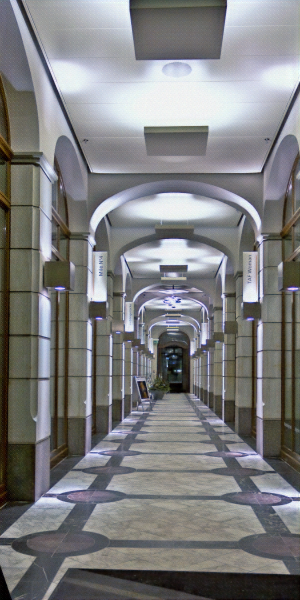}
		\includegraphics[width=0.09\linewidth,height=0.15\linewidth, frame]{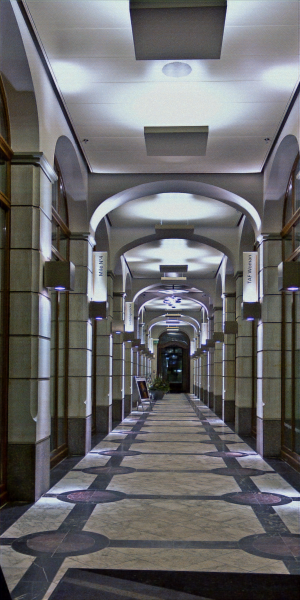}
		\includegraphics[width=0.09\linewidth,height=0.15\linewidth, frame]{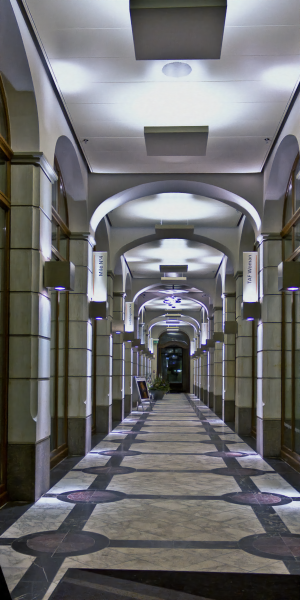}
		\includegraphics[width=0.09\linewidth,height=0.15\linewidth, frame]{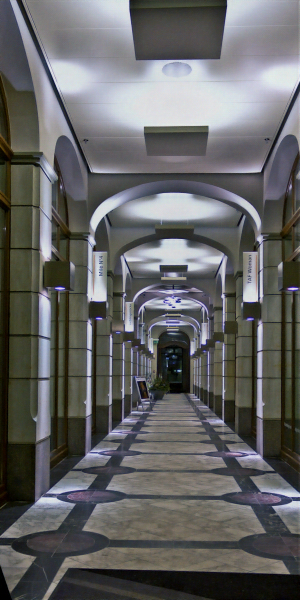}
		\includegraphics[width=0.09\linewidth,height=0.15\linewidth, frame]{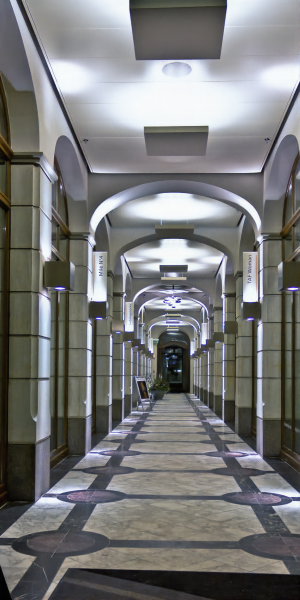}\\
		\rotatebox{90}{\scriptsize{~~~~~~~~Error Map (\textcolor{magenta}{Left})}}
		\includegraphics[width=0.09\linewidth,height=0.15\linewidth, frame]{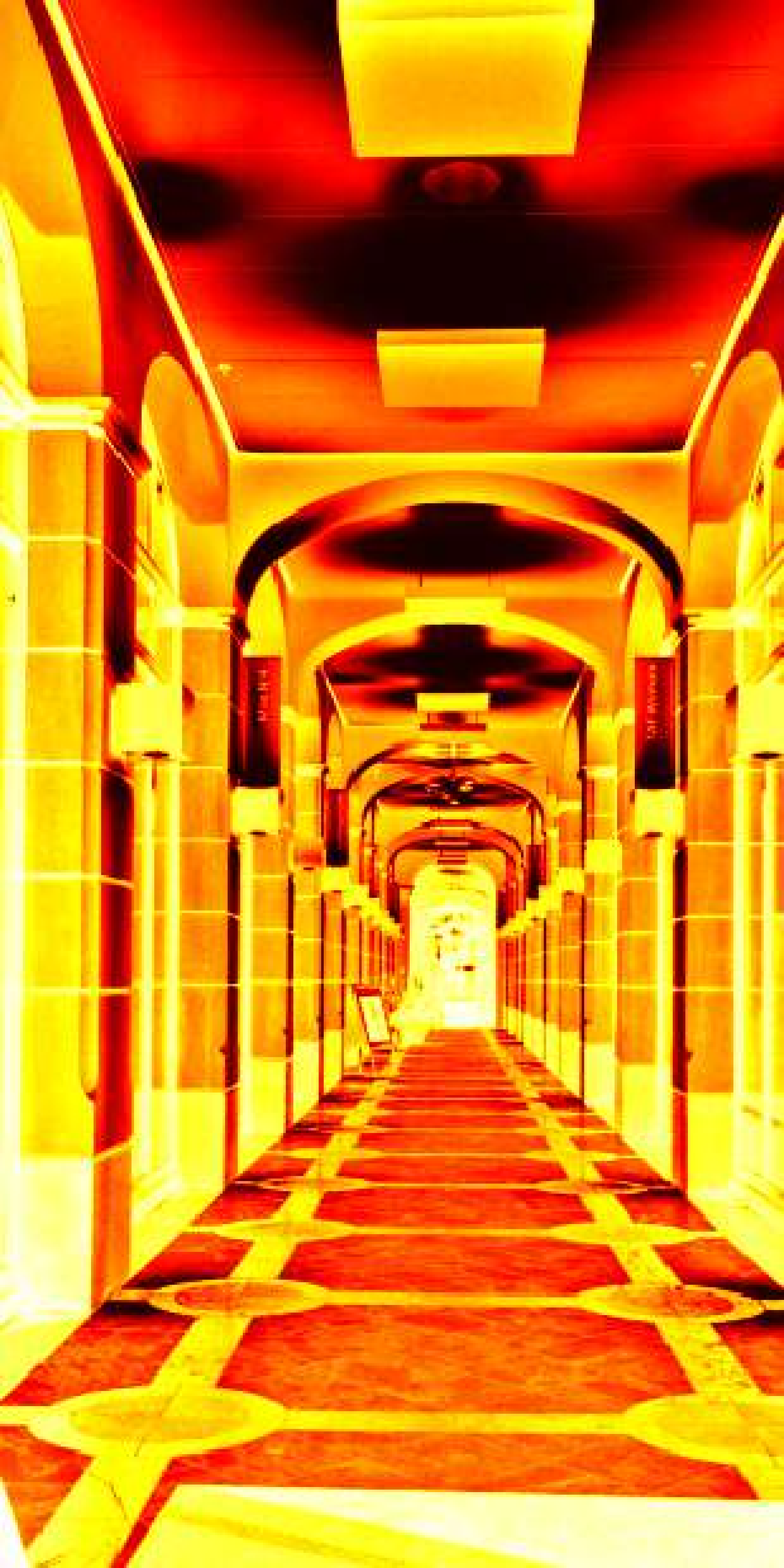}
		\includegraphics[width=0.09\linewidth,height=0.15\linewidth, frame]{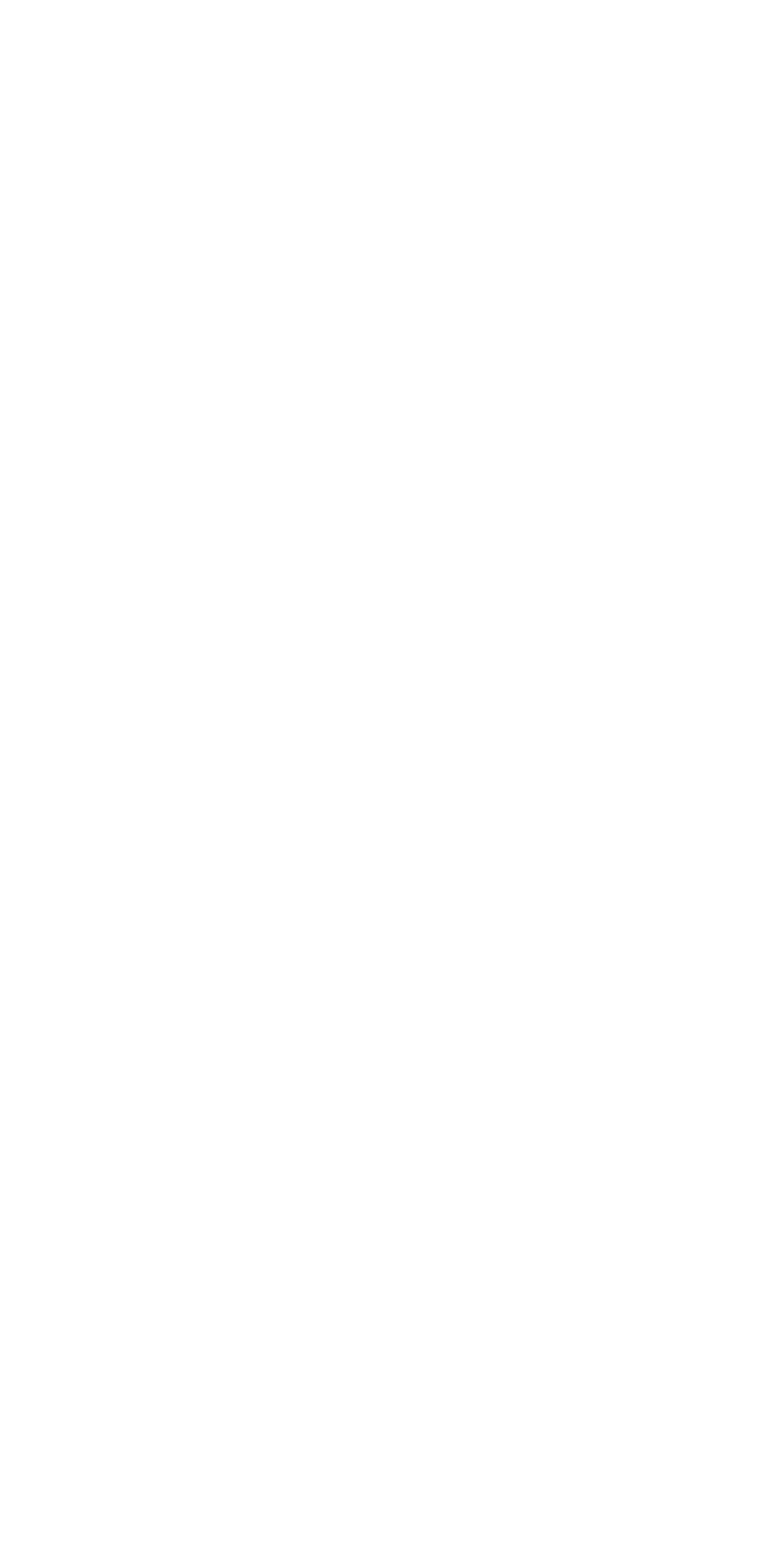}
		\includegraphics[width=0.09\linewidth,height=0.15\linewidth, frame]{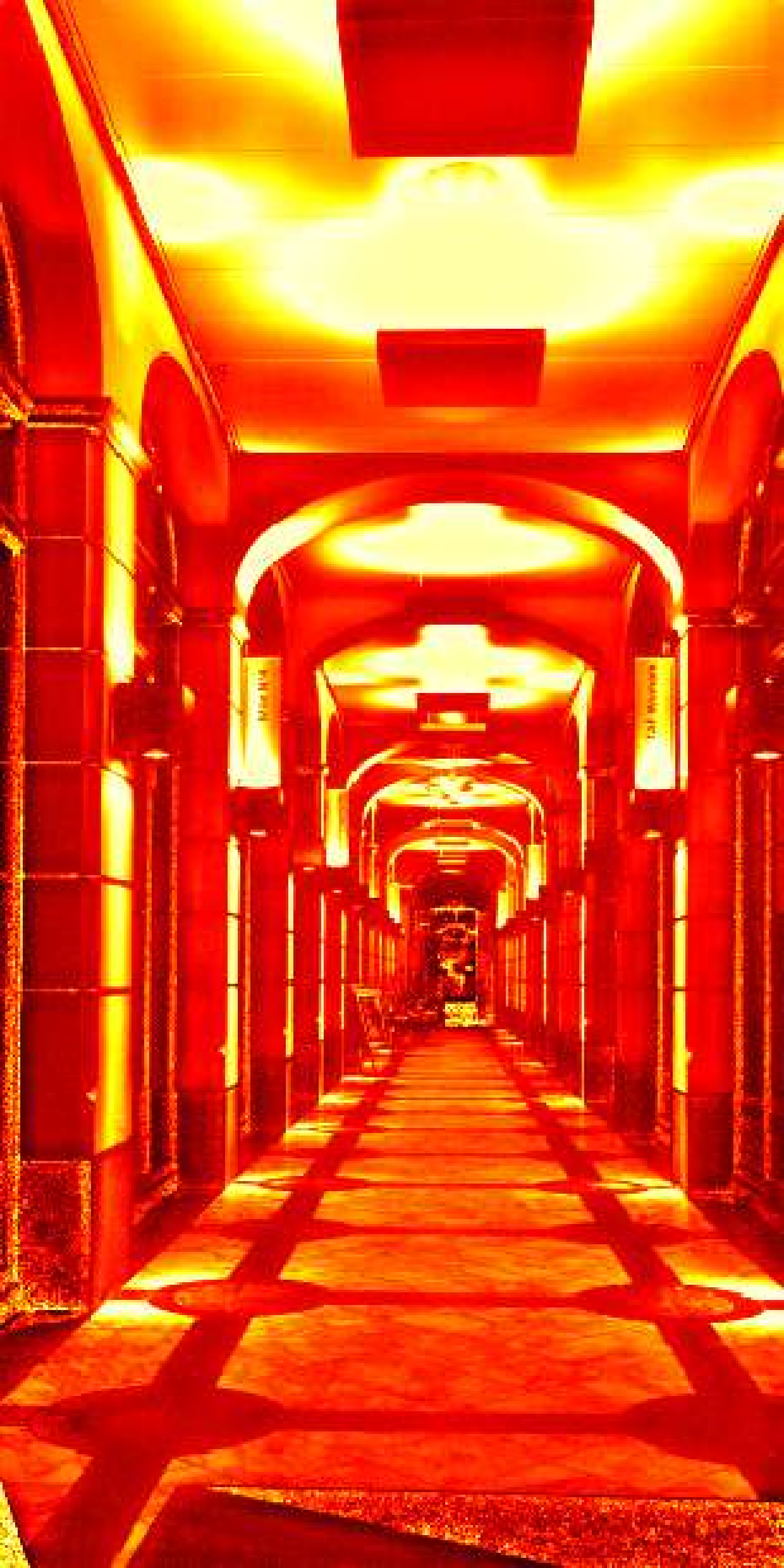}
		\includegraphics[width=0.09\linewidth,height=0.15\linewidth, frame]{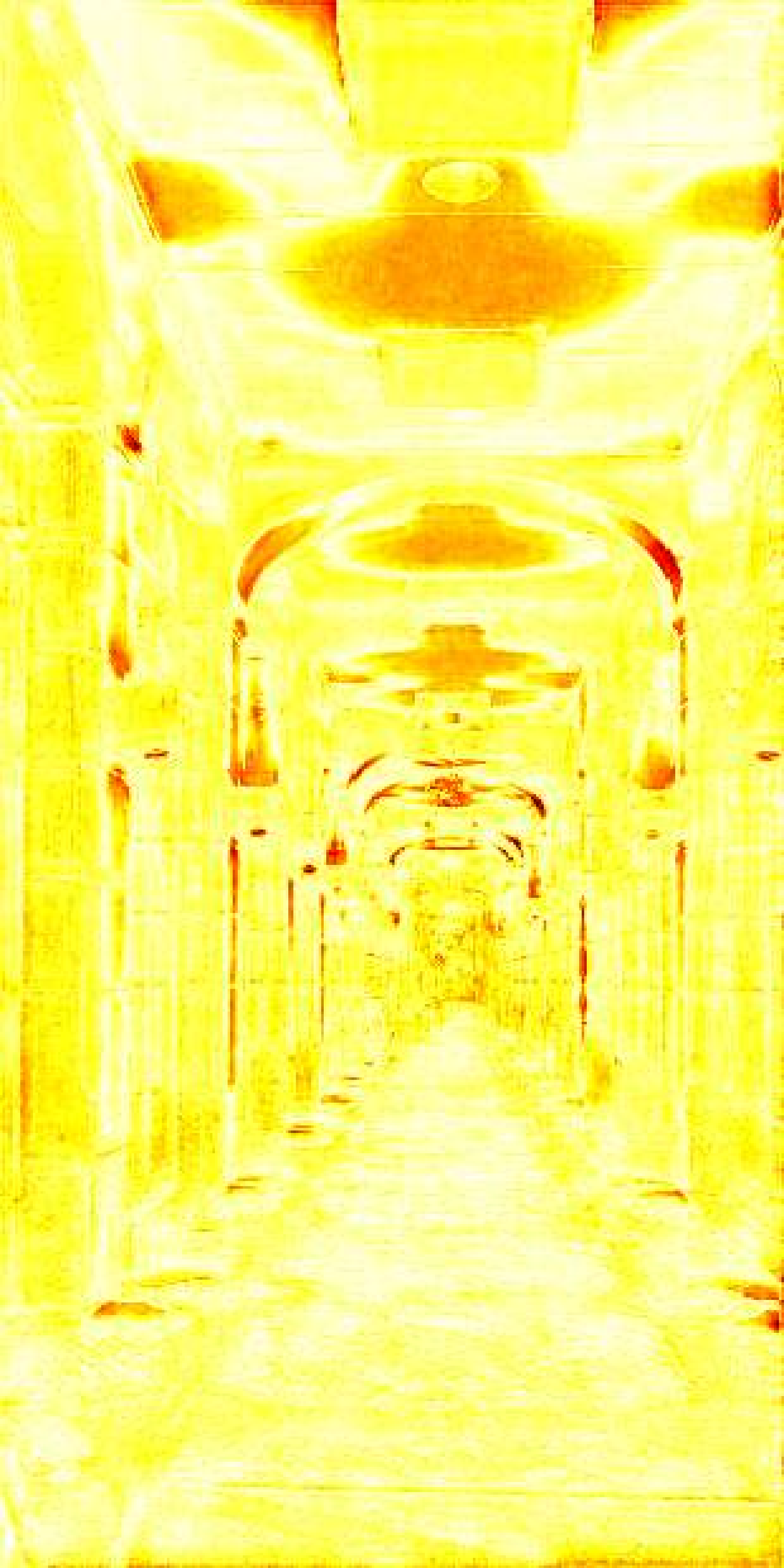}
		\includegraphics[width=0.09\linewidth,height=0.15\linewidth, frame]{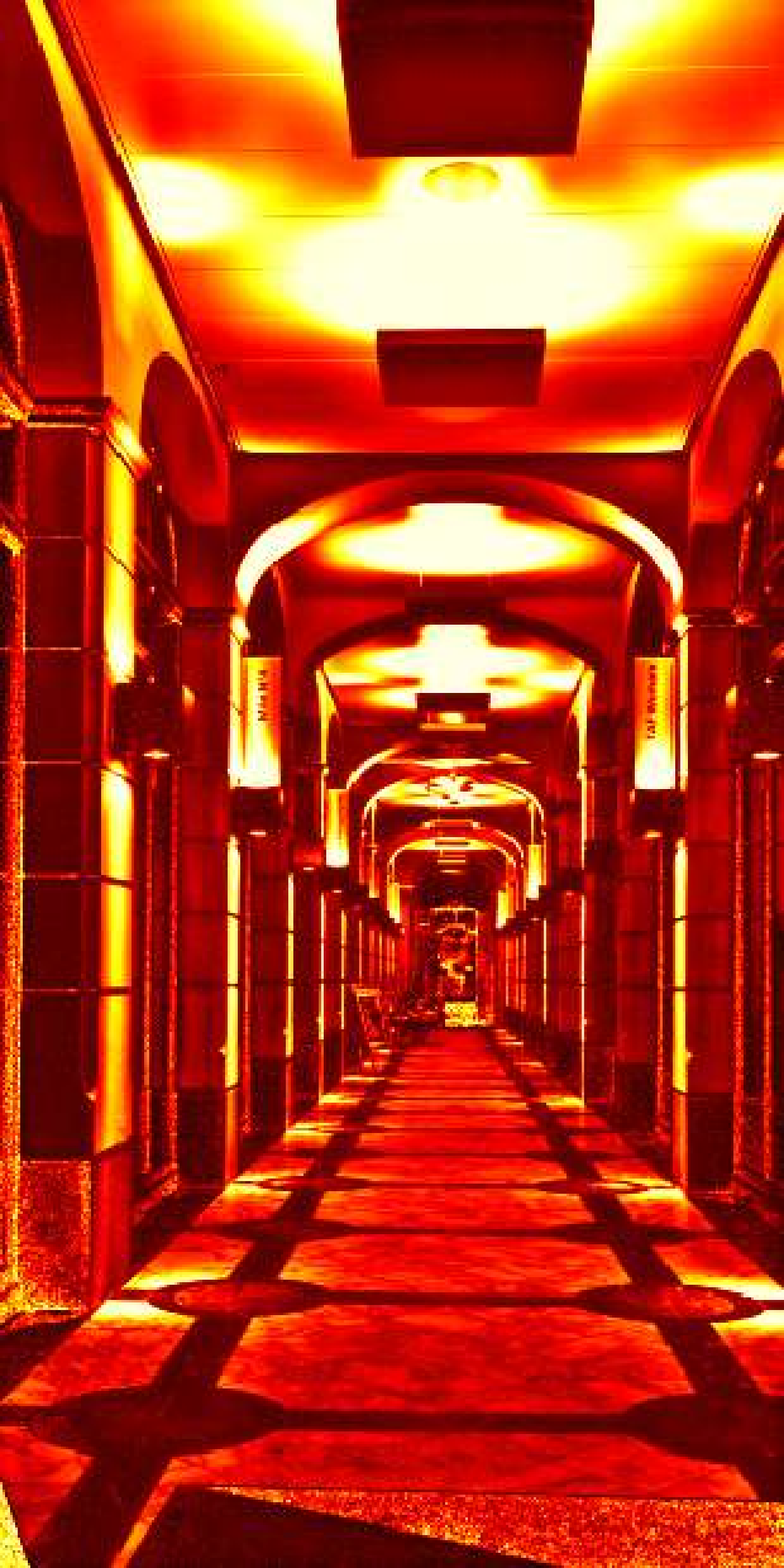}
		\includegraphics[width=0.09\linewidth,height=0.15\linewidth, frame]{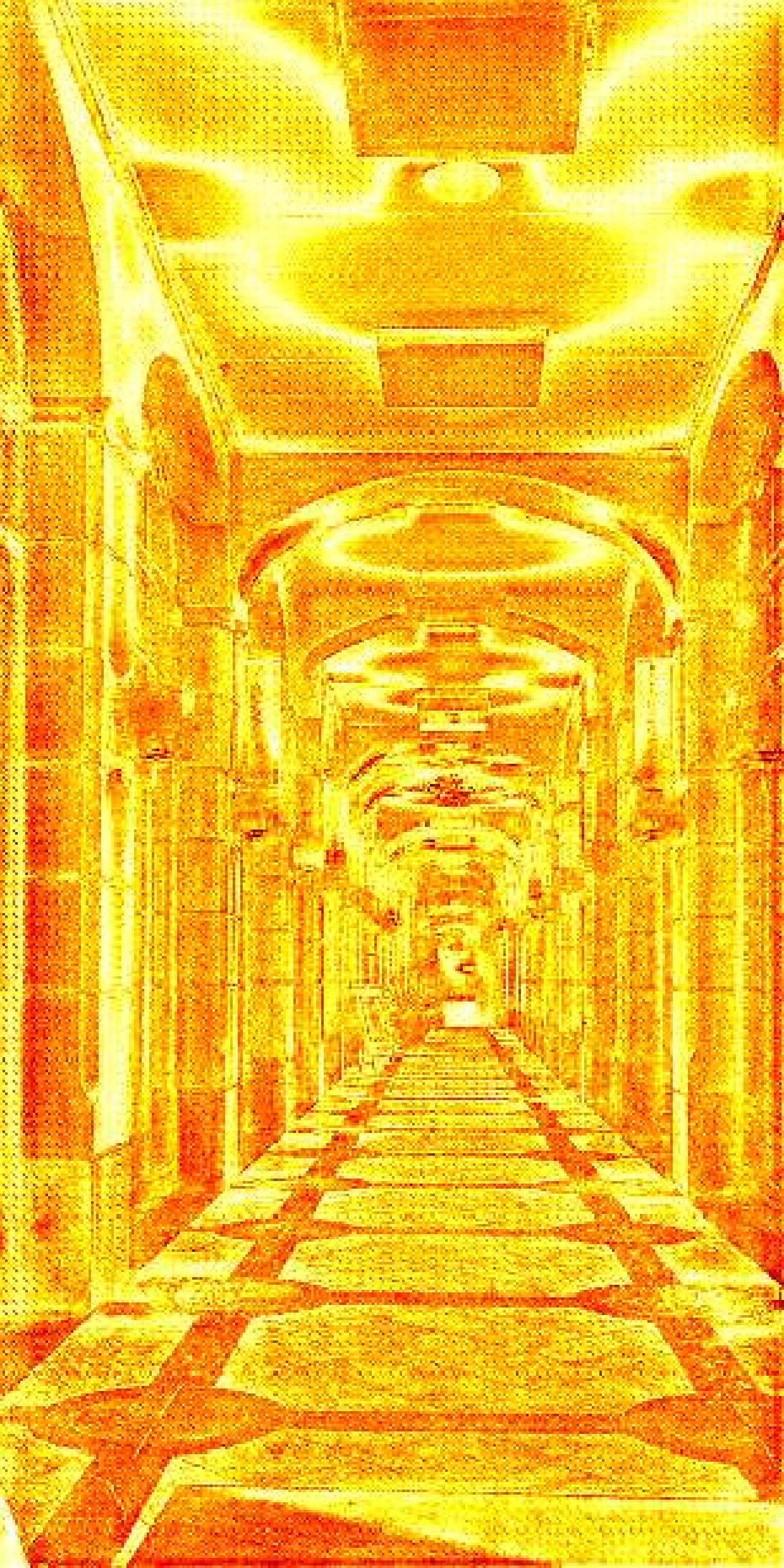}
		\includegraphics[width=0.09\linewidth,height=0.15\linewidth, frame]{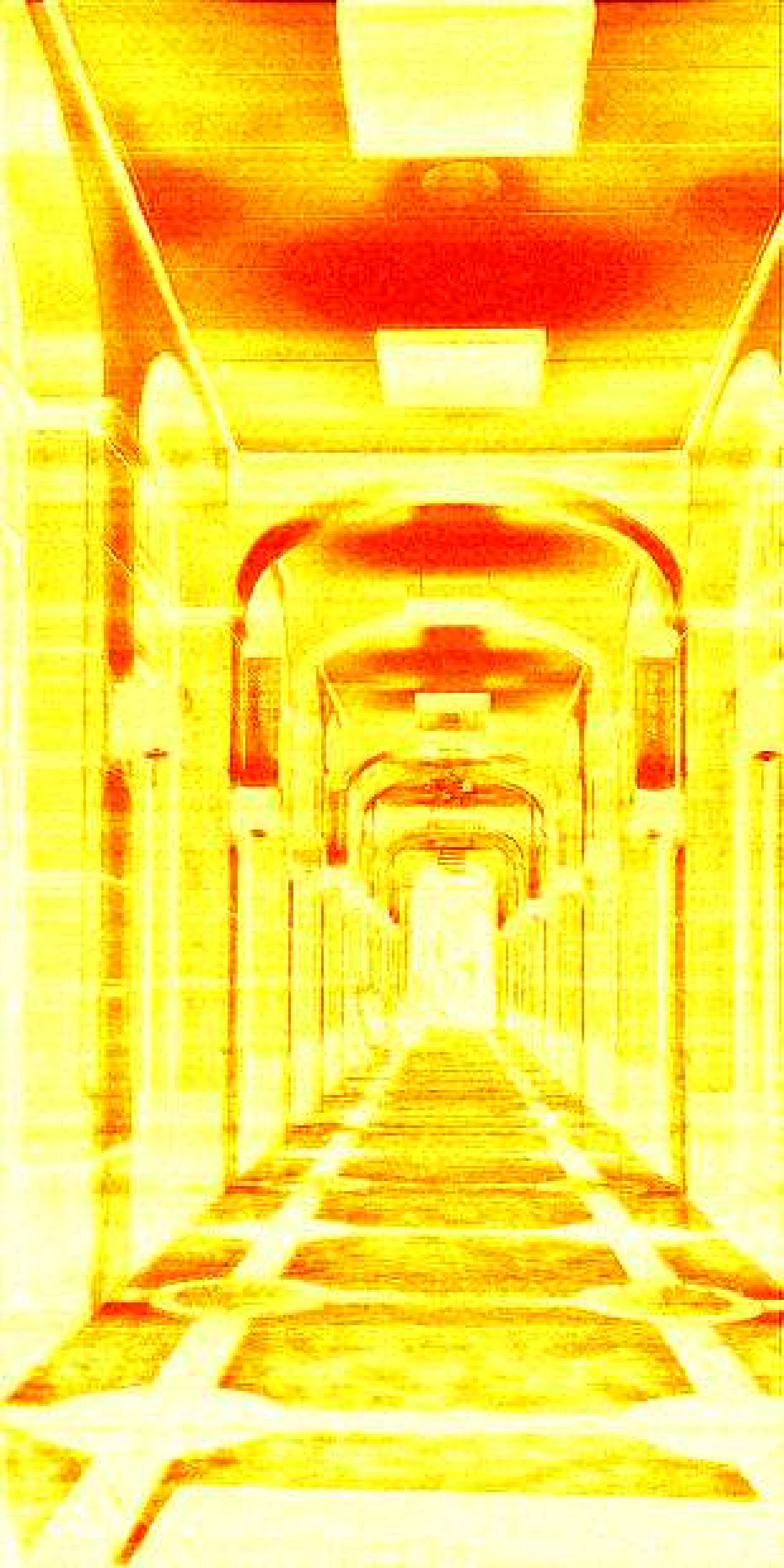}
		\includegraphics[width=0.09\linewidth,height=0.15\linewidth, frame]{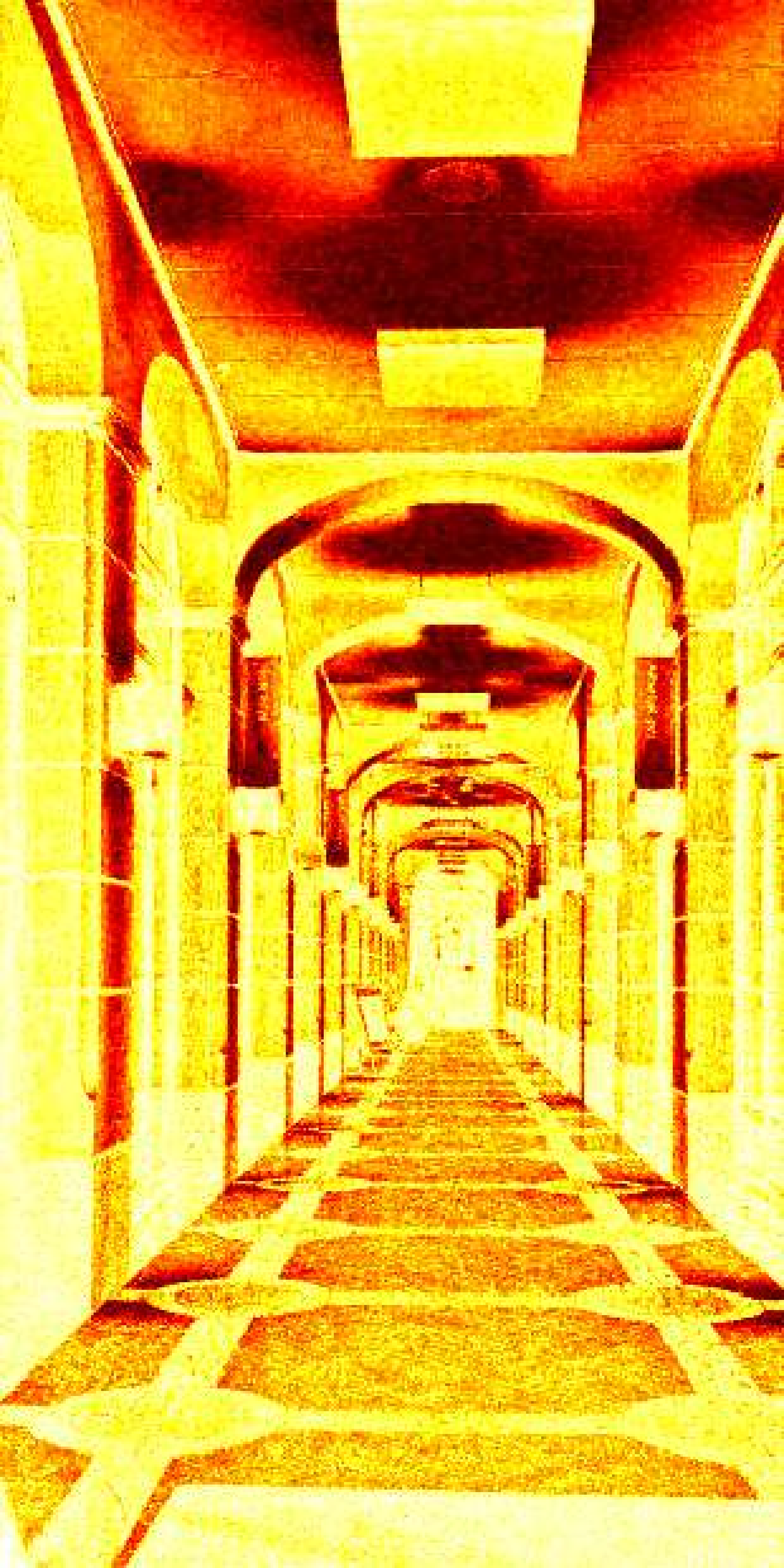}
		\includegraphics[width=0.09\linewidth,height=0.15\linewidth, frame]{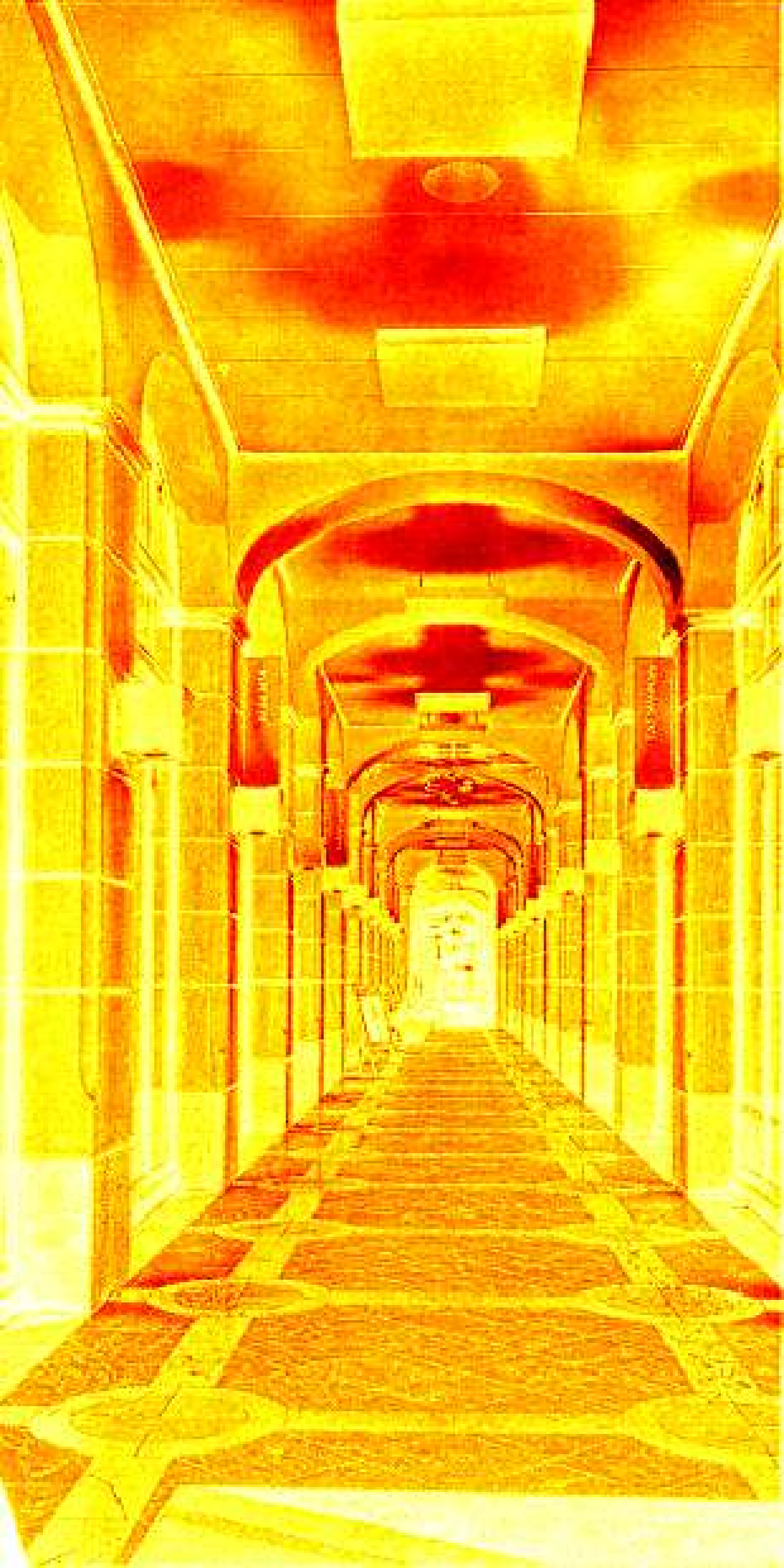}
		\includegraphics[width=0.09\linewidth,height=0.15\linewidth, frame]{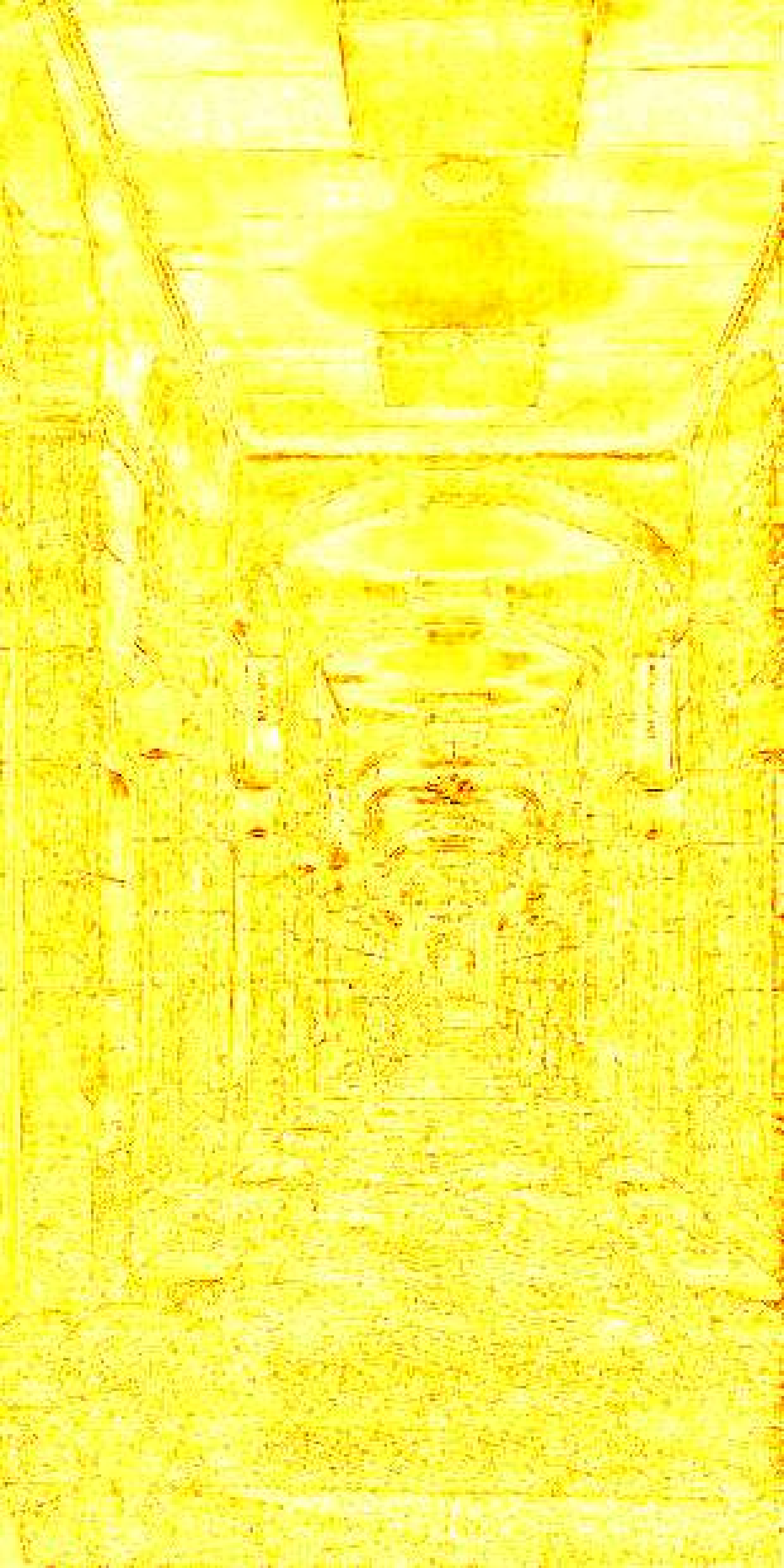}
	\end{minipage}\\
	\begin{minipage}{1\linewidth}
		\vspace{2pt}
		\hspace{4mm}
		\begin{minipage}{0.09\linewidth} \centering \footnotesize Input\\(PSNR/SSIM) \end{minipage}\hspace{1pt}
		\begin{minipage}{0.09\linewidth} \centering \footnotesize GT\\($+\infty$/1.000) \end{minipage}\hspace{1pt}
		\begin{minipage}{0.09\linewidth} \centering \footnotesize ZeroDCE\\(8.49/0.616) \end{minipage}\hspace{1pt}
		\begin{minipage}{0.09\linewidth} \centering	\footnotesize iPASSR\\(28.56/0.954) \end{minipage}\hspace{1pt}
		\begin{minipage}{0.09\linewidth} \centering	\footnotesize ZeroDCE++\\(7.27/0.540) \end{minipage}\hspace{1pt}
		\begin{minipage}{0.09\linewidth} \centering \footnotesize DVENet\\(25.63/0.925) \end{minipage}\hspace{1pt}
		\begin{minipage}{0.09\linewidth} \centering \footnotesize NAFSSR\\(24.03/0.946) \end{minipage}\hspace{1pt}
		\begin{minipage}{0.09\linewidth} \centering	\footnotesize NAFSSR-F\\(27.43/0.978) \end{minipage}\hspace{1pt}
		\begin{minipage}{0.09\linewidth} \centering	\footnotesize SNR\\(22.5 /0.944) \end{minipage}\hspace{1pt}
		\begin{minipage}{0.09\linewidth} \centering	\footnotesize \textbf{DCI-Net}\\(34.47/0.977) \end{minipage}
	\end{minipage}
	\begin{minipage}{1\linewidth}
		\centering
		\rotatebox{90}{\scriptsize{~~~~~~~~RGB Image (\textcolor{blue}{Right})}}
		\includegraphics[width=0.09\linewidth,height=0.15\linewidth, frame]{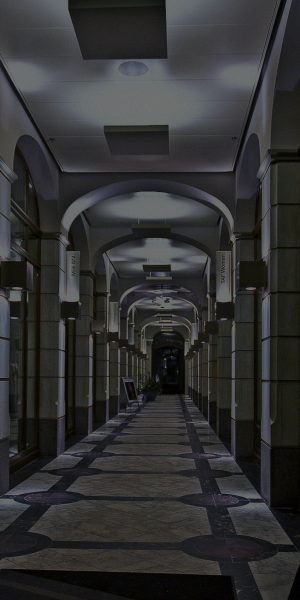}
		\includegraphics[width=0.09\linewidth,height=0.15\linewidth, frame]{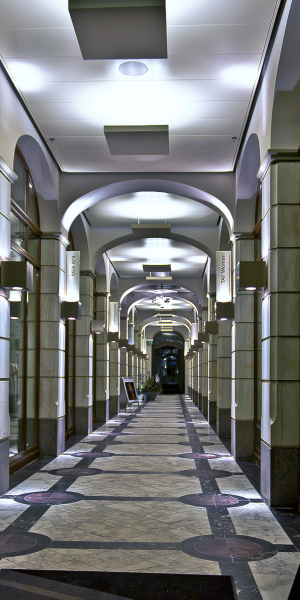}
		\includegraphics[width=0.09\linewidth,height=0.15\linewidth, frame]{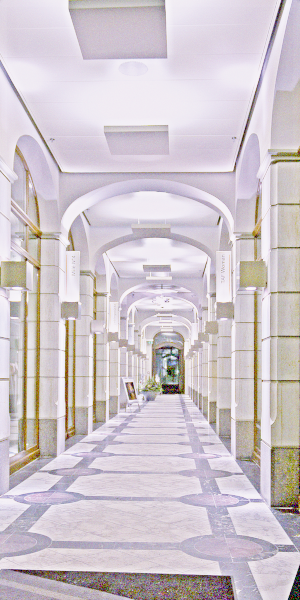}
		\includegraphics[width=0.09\linewidth,height=0.15\linewidth, frame]{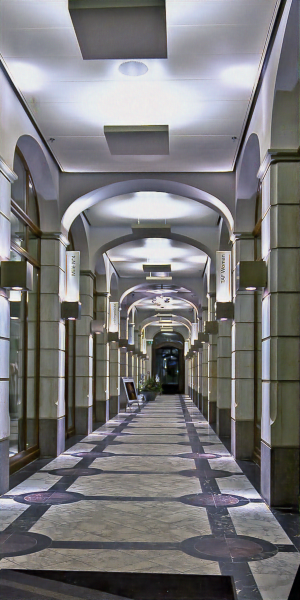}
		\includegraphics[width=0.09\linewidth,height=0.15\linewidth, frame]{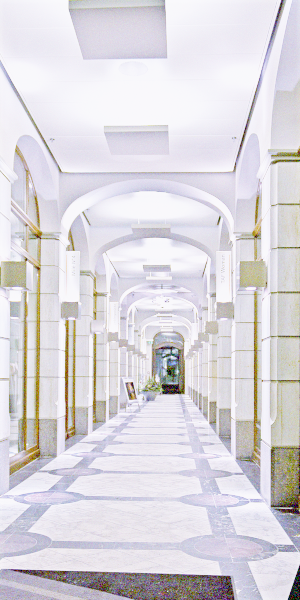}
		\includegraphics[width=0.09\linewidth,height=0.15\linewidth, frame]{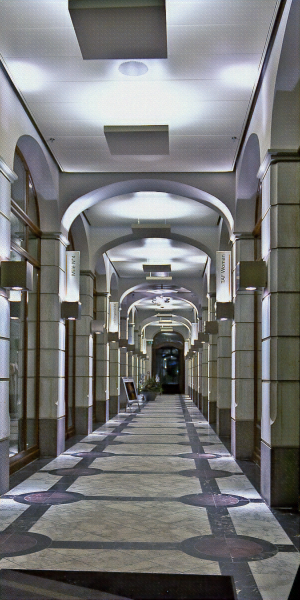}
		\includegraphics[width=0.09\linewidth,height=0.15\linewidth, frame]{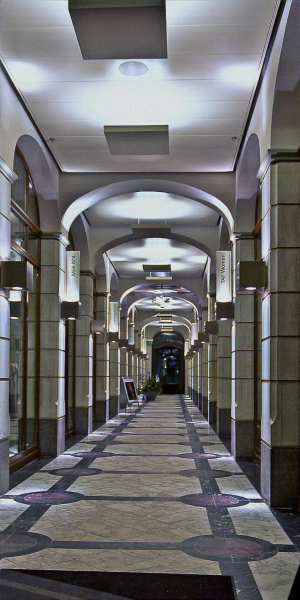}
		\includegraphics[width=0.09\linewidth,height=0.15\linewidth, frame]{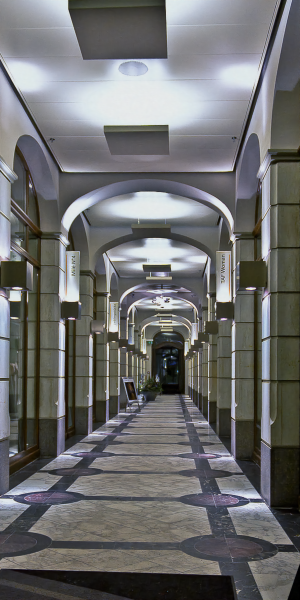}
		\includegraphics[width=0.09\linewidth,height=0.15\linewidth, frame]{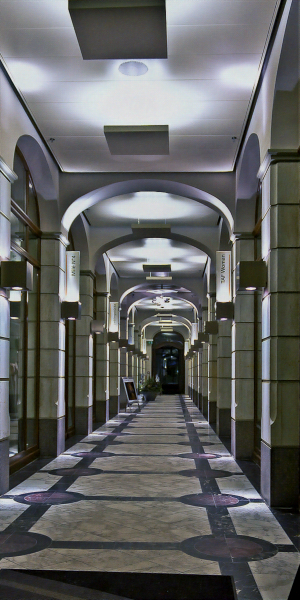}
		\includegraphics[width=0.09\linewidth,height=0.15\linewidth, frame]{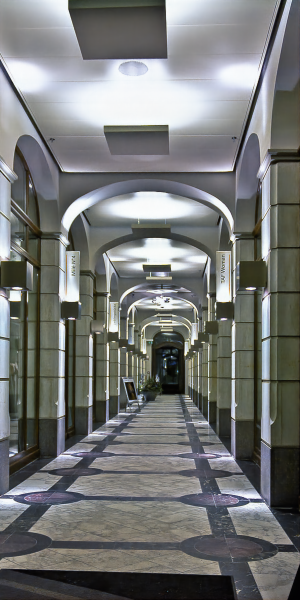}\\
		\rotatebox{90}{\scriptsize{~~~~~~~~Error Map  (\textcolor{blue}{Right})}}
		\includegraphics[width=0.09\linewidth,height=0.15\linewidth, frame]{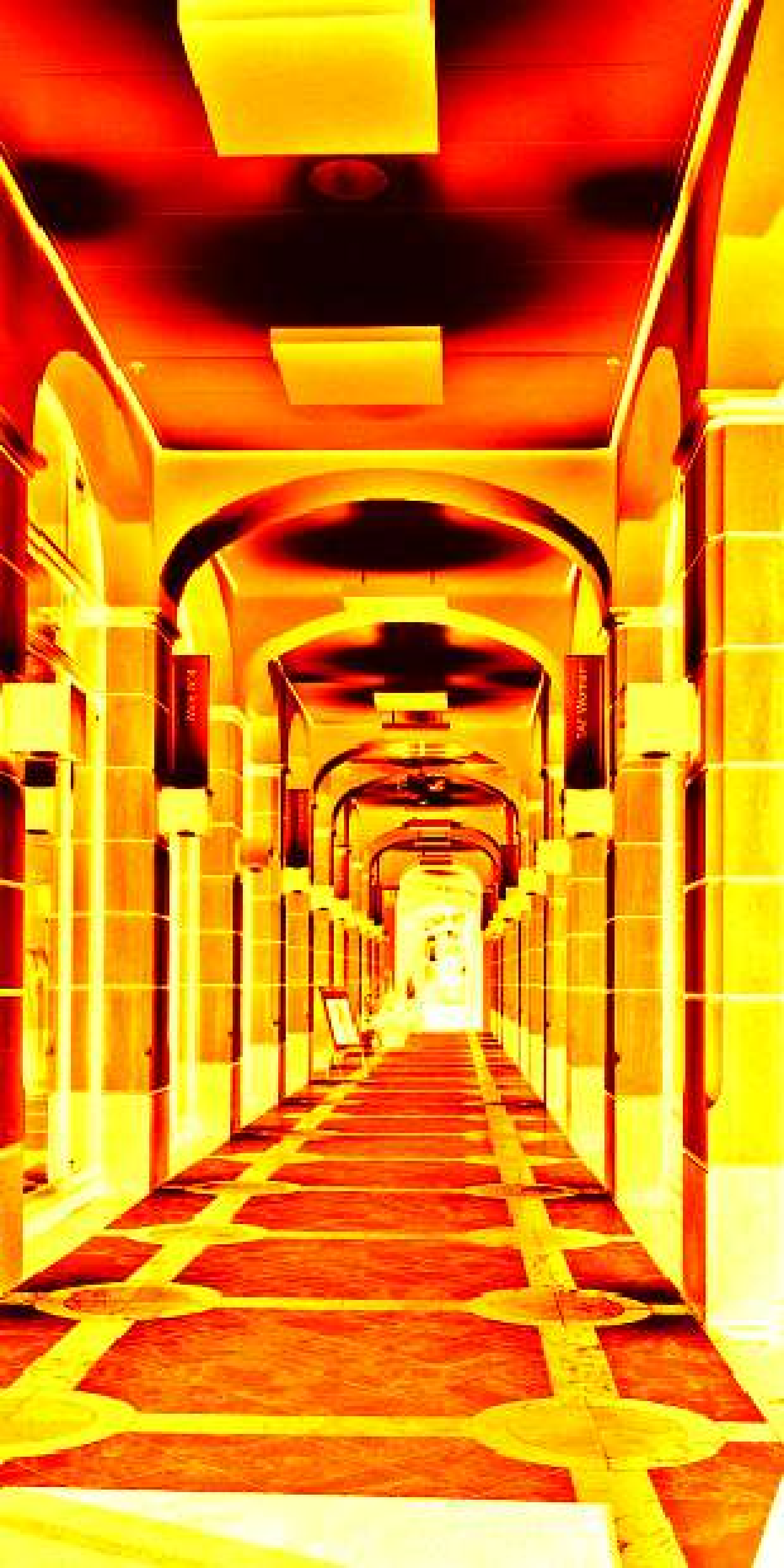}
		\includegraphics[width=0.09\linewidth,height=0.15\linewidth, frame]{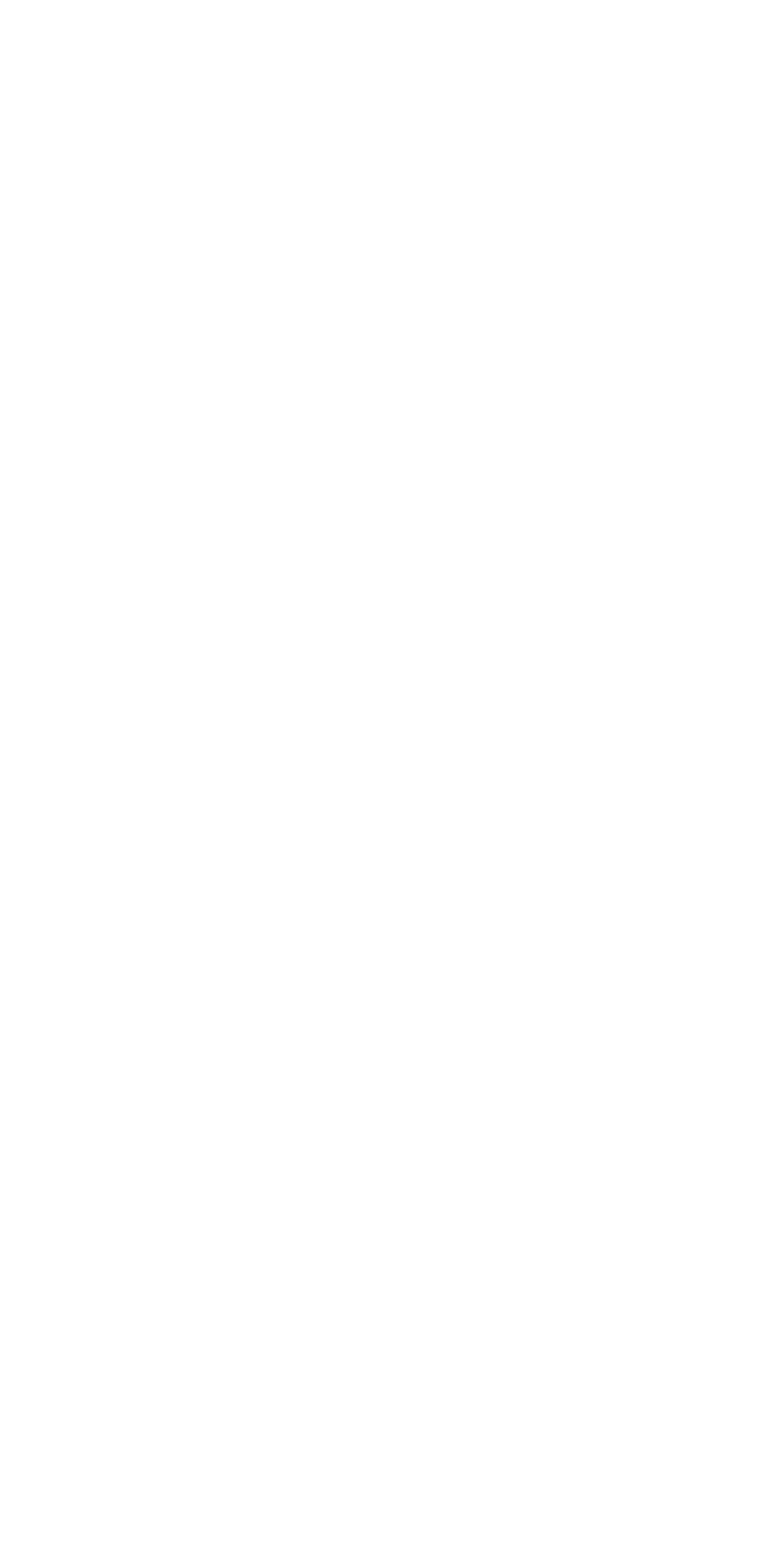}
		\includegraphics[width=0.09\linewidth,height=0.15\linewidth, frame]{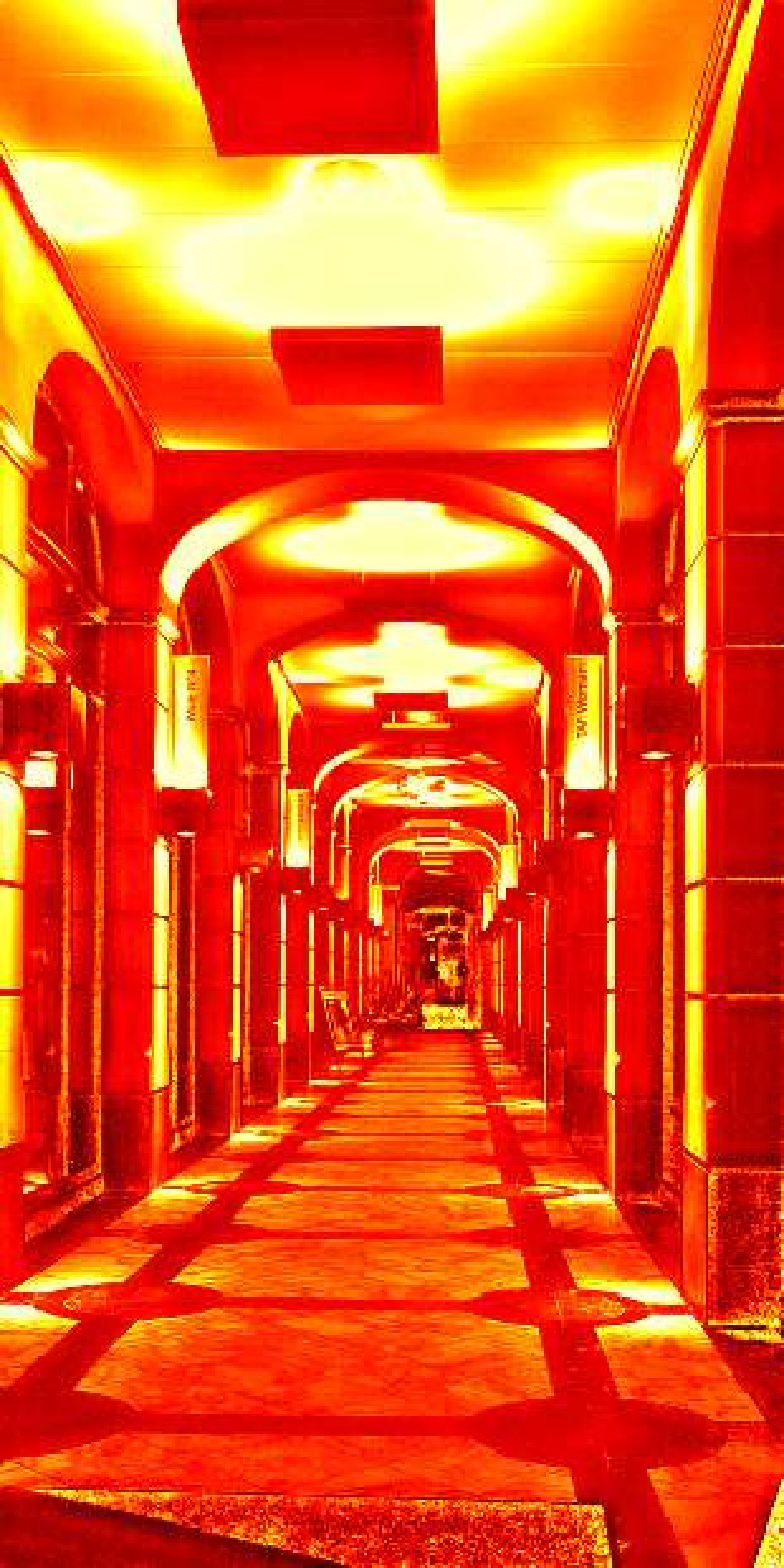}
		\includegraphics[width=0.09\linewidth,height=0.15\linewidth, frame]{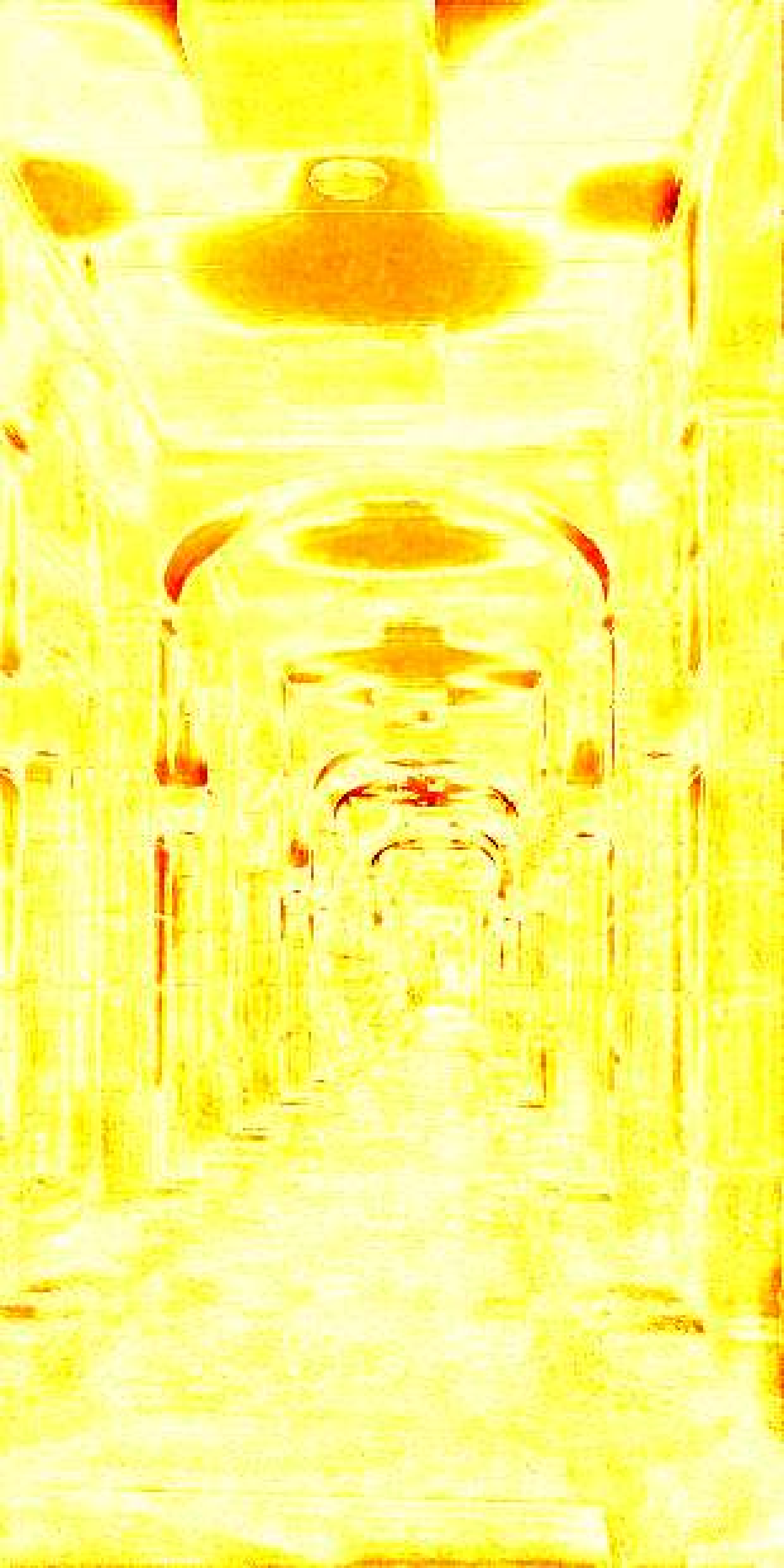}
		\includegraphics[width=0.09\linewidth,height=0.15\linewidth, frame]{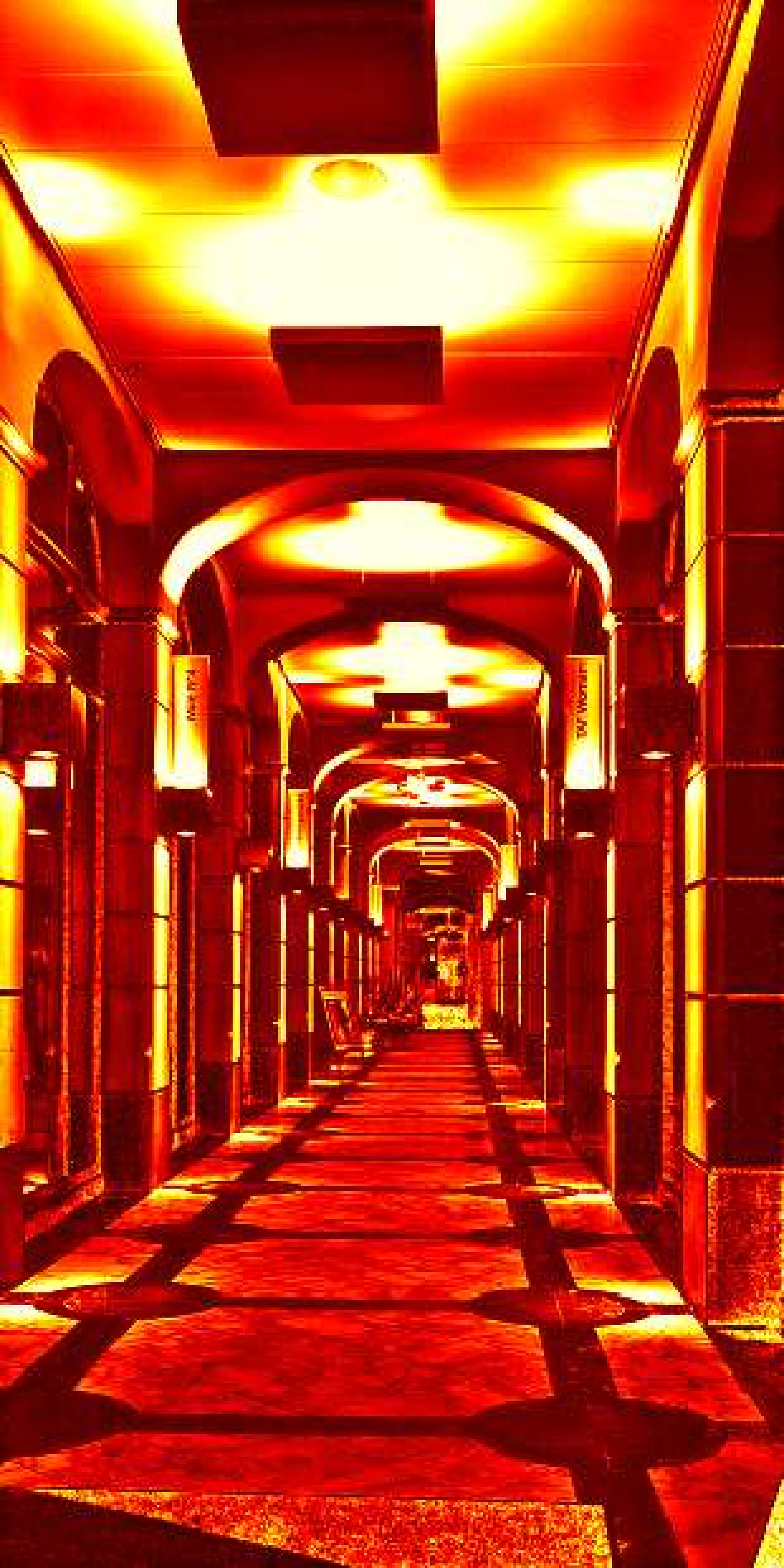}
		\includegraphics[width=0.09\linewidth,height=0.15\linewidth, frame]{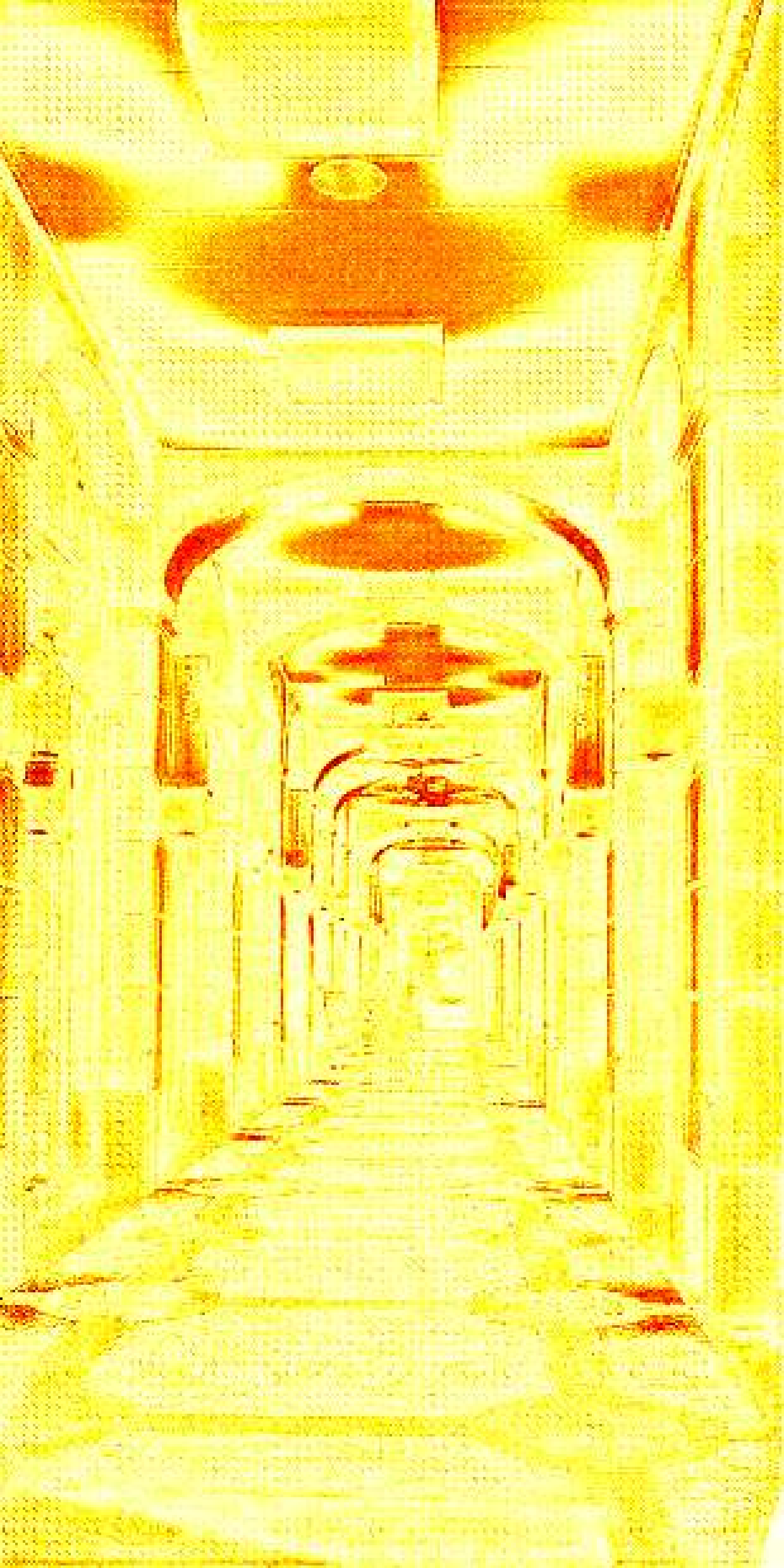}
		\includegraphics[width=0.09\linewidth,height=0.15\linewidth, frame]{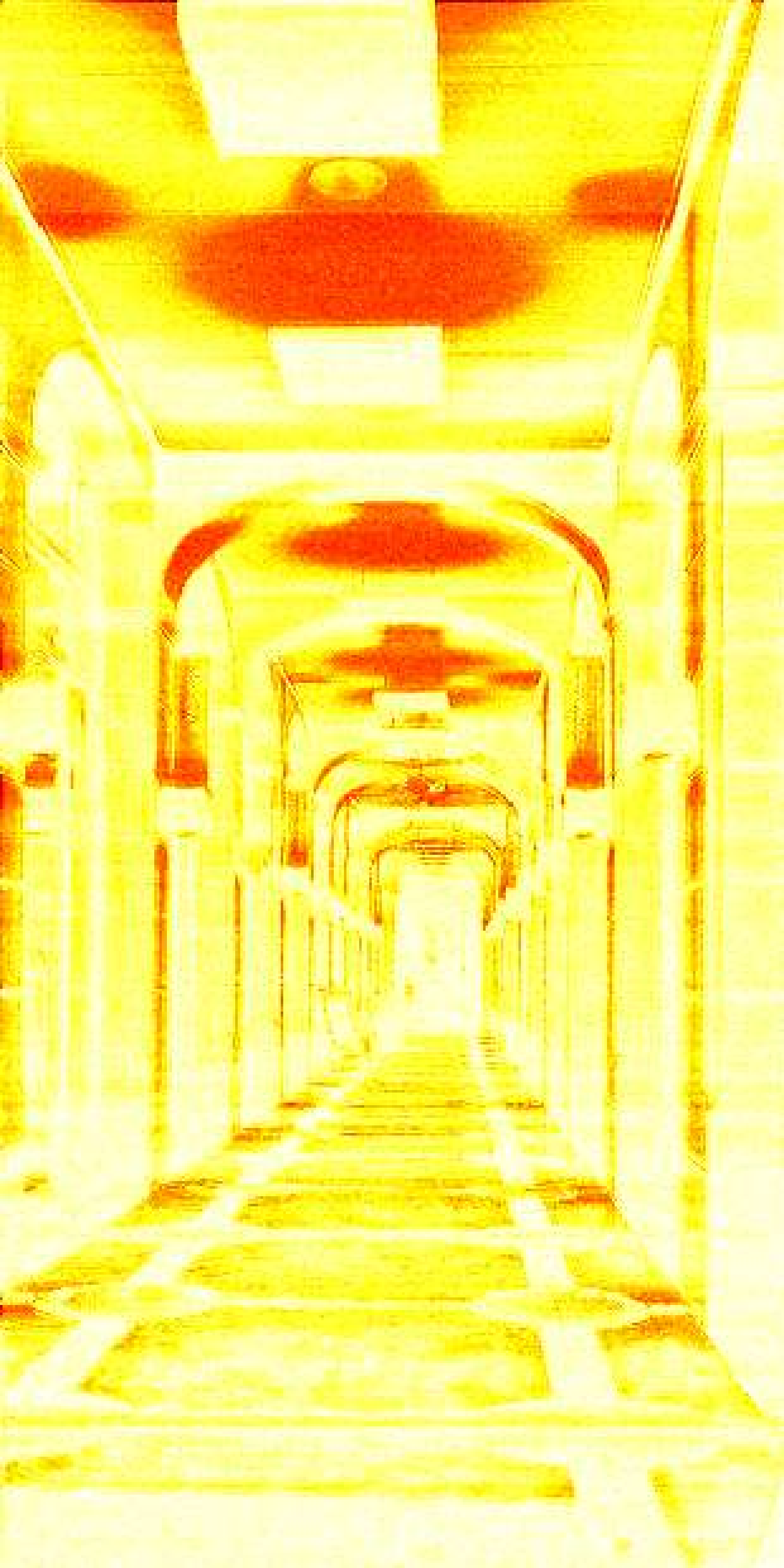}
		\includegraphics[width=0.09\linewidth,height=0.15\linewidth, frame]{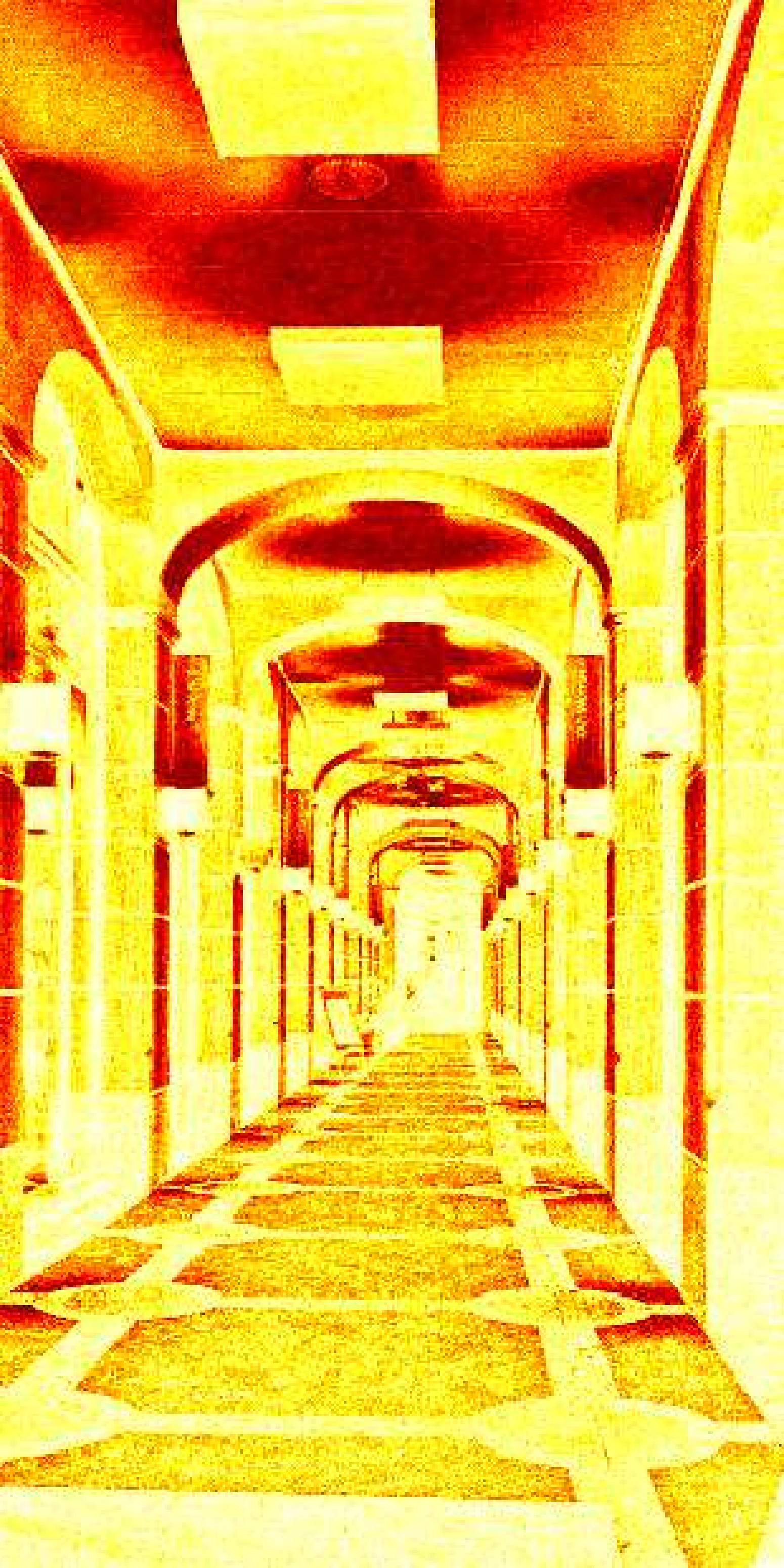}
		\includegraphics[width=0.09\linewidth,height=0.15\linewidth, frame]{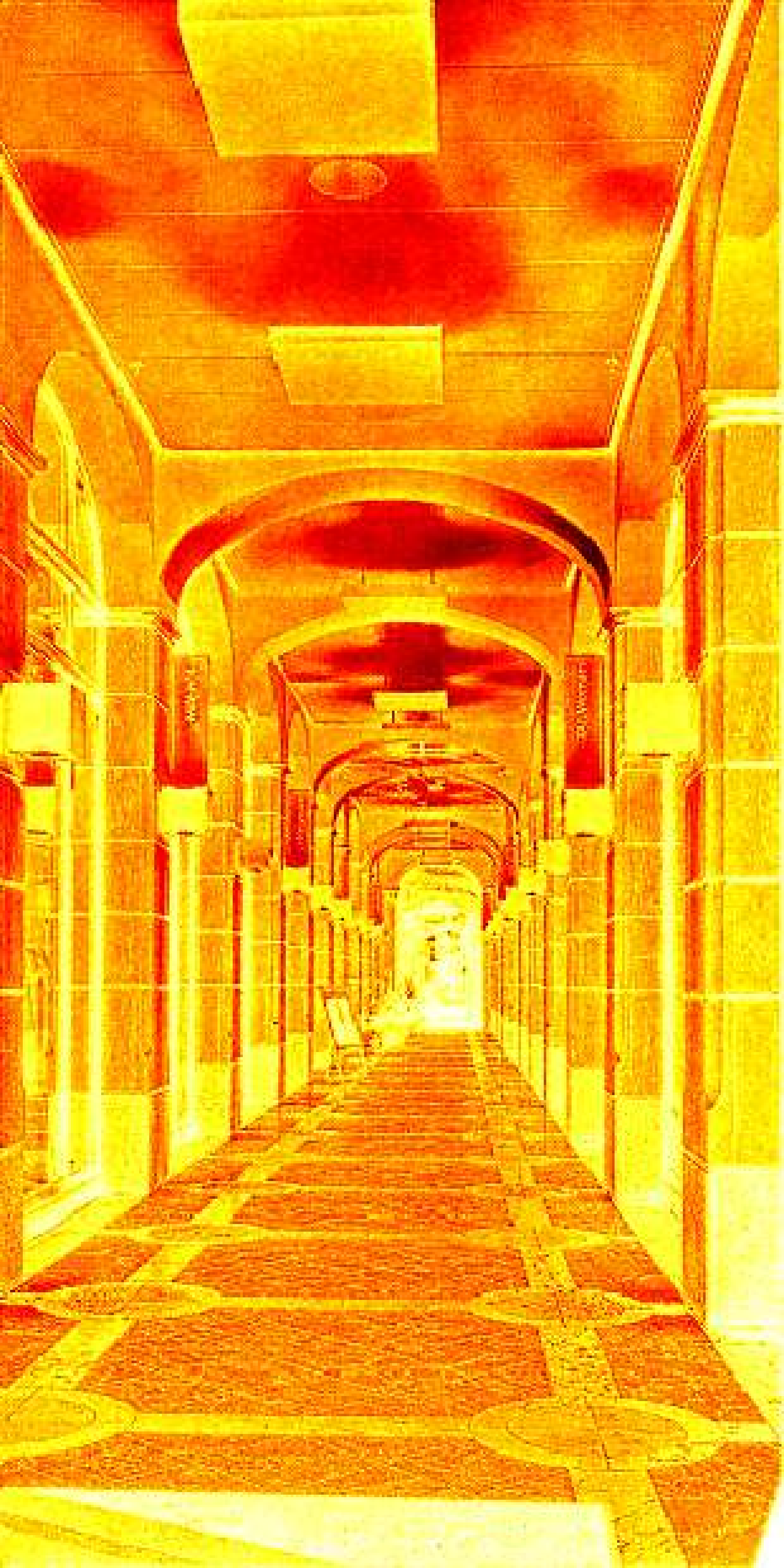}
		\includegraphics[width=0.09\linewidth,height=0.15\linewidth, frame]{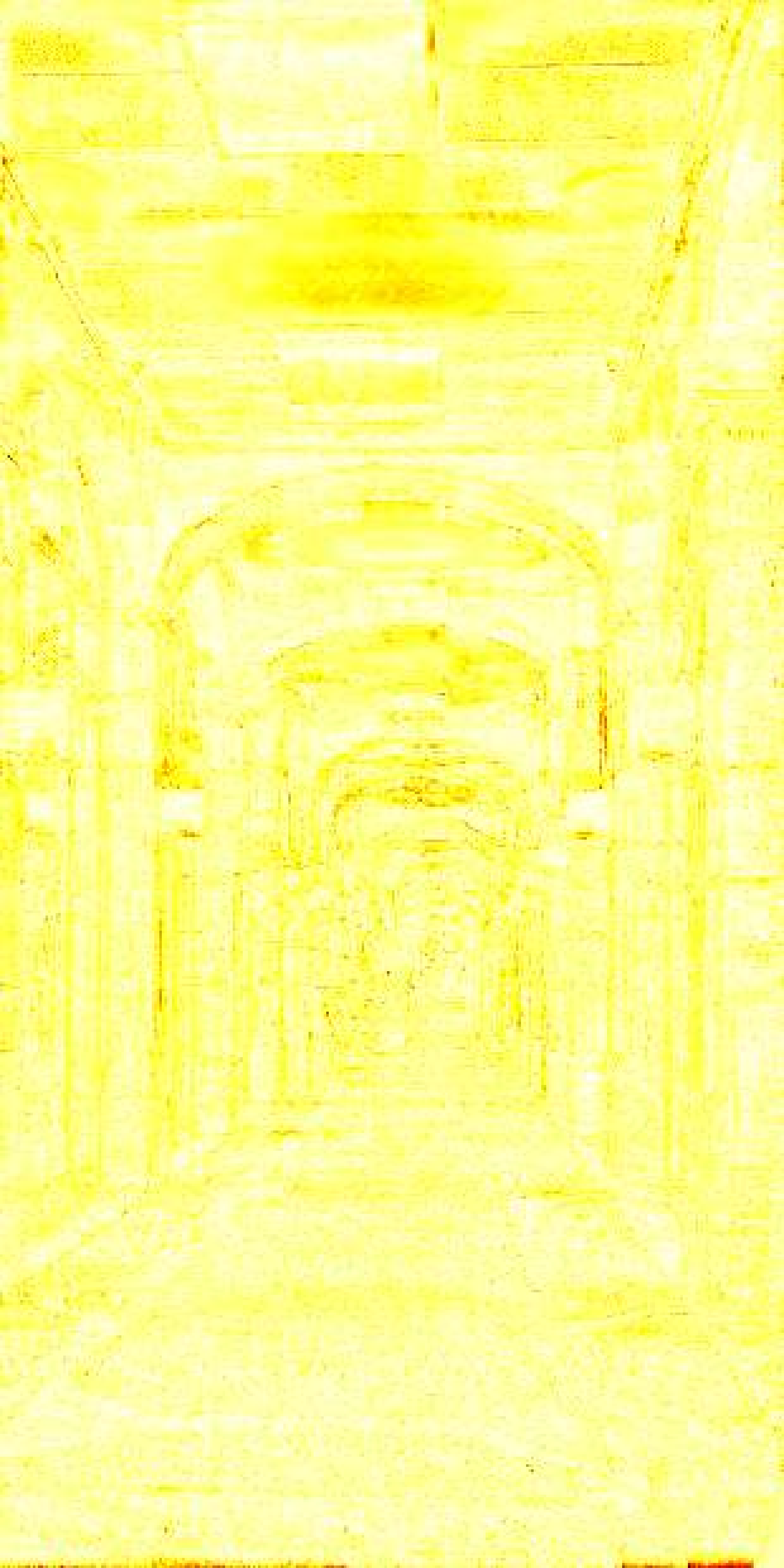}
	\end{minipage}\\
	\begin{minipage}{1\linewidth}
		\vspace{2pt}
		\hspace{4mm}
		\begin{minipage}{0.09\linewidth} \centering \footnotesize Input\\(PSNR/SSIM) \end{minipage}\hspace{1pt}
		\begin{minipage}{0.09\linewidth} \centering \footnotesize GT\\($+\infty$/1.000) \end{minipage}\hspace{1pt}
		\begin{minipage}{0.09\linewidth} \centering \footnotesize ZeroDCE\\(8.53/0.616) \end{minipage}\hspace{1pt}
		\begin{minipage}{0.09\linewidth} \centering	\footnotesize iPASSR\\(29.21/0.954) \end{minipage}\hspace{1pt}
		\begin{minipage}{0.09\linewidth} \centering	\footnotesize ZeroDCE++\\(7.28/0.534) \end{minipage}\hspace{1pt}
		\begin{minipage}{0.09\linewidth} \centering \footnotesize DVENet\\(29.12/0.932) \end{minipage}\hspace{1pt}
		\begin{minipage}{0.09\linewidth} \centering \footnotesize NAFSSR\\(25.65/0.951) \end{minipage}\hspace{1pt}
		\begin{minipage}{0.09\linewidth} \centering	\footnotesize NAFSSR-F\\(27.77/0.973) \end{minipage}\hspace{1pt}
		\begin{minipage}{0.09\linewidth} \centering	\footnotesize SNR\\(21.83/0.939) \end{minipage}\hspace{1pt}
		\begin{minipage}{0.09\linewidth} \centering	\footnotesize \textbf{DCI-Net}\\(35.87/0.972) \end{minipage}
	\end{minipage}
	\vspace{-4mm}
	\caption{Visualization of the enhanced images and corresponding error maps of each method based on Flickr1024 dataset, including the results of  NAFSSR \cite{Chu2022NAFSSRSI}, NAFSSR-F \cite{Chu2022NAFSSRSI}, iPASSRNet \cite{Wang2021SymmetricPA}, SNR \cite{Xu2022SNRAwareLI}, DVENet \cite{Huang2022LowLightSI}, ZeroDCE \cite{guo2020zero}, ZeroDCE++ \cite{li2021learning} and our DCI-Net. Whiter and brighter pixels in the error maps indicate smaller errors. Clearly, our DCI-Net achieves better illumination adjustment and color correction, and obtains smaller errors and superior performance than other methods, which can also be seen from the shown metrics.}
	\label{fig:5}
	\vspace{-2mm}
\end{figure*}

\subsection{Spatial-channel information mining block (SIMB)}
Recalling that previous single low-light image enhancement and stereo image restoration methods usually adopt CNNs with a kernel size of 3$\times$3 to construct a deep neural network. As a result, the long-range dependency cannot be well captured. Recently, transformer-based models have achieved impressive performance in diverse computer vision tasks due to the capacity of building long-range dependency \cite{dosovitskiy2020image, liu2021swin}. But vision transformer is computationally expensive. Some recent works have shown that large kernel convolutional layers can also obtain long-range correlations \cite{ding2022scaling}. Besides, PoolFormer shows that the overall architecture plays an important role in vision transformer \cite{yu2022metaformer}. Inspired by these works, we therefore propose SIMB. Although SIMB inherits the structure of vision transformer, it replaces the multi-head self-attention with large kernel convolutional layers, and further explores the information flow in channel dimension. The structure of SIMB is shown on the right of Fig. \ref{fig:2}. As can be seen, three are two core steps in SIMB: long-range dependency capture (LRDC) and expanded channel information refinement (ECIR). 

\begin{figure*}[t]
	\centering  
	\begin{minipage}{1\linewidth}
		\centering
		\includegraphics[width=0.24\linewidth]{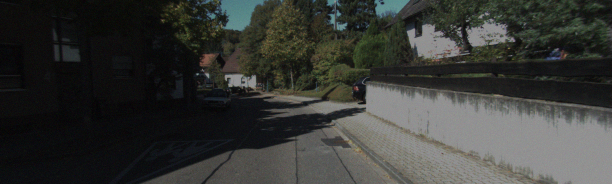}
		\includegraphics[width=0.24\linewidth]{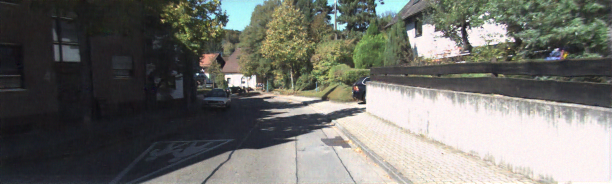}
		\includegraphics[width=0.24\linewidth]{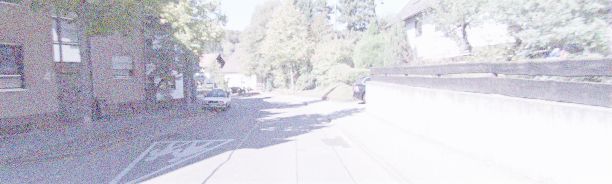}
		\includegraphics[width=0.24\linewidth]{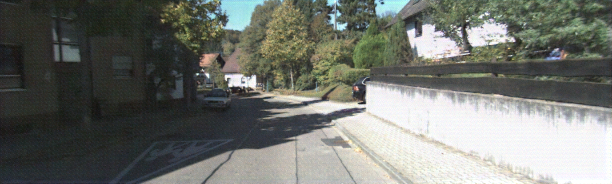}\\
		\includegraphics[width=0.24\linewidth]{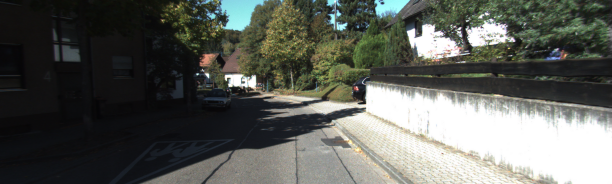}
		\includegraphics[width=0.24\linewidth]{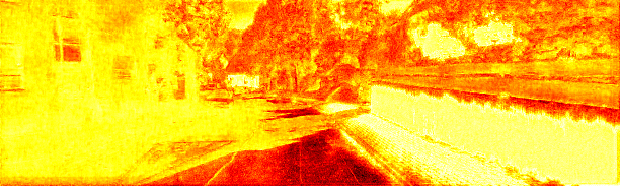}
		\includegraphics[width=0.24\linewidth]{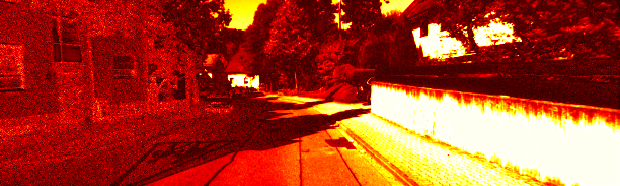}
		\includegraphics[width=0.24\linewidth]{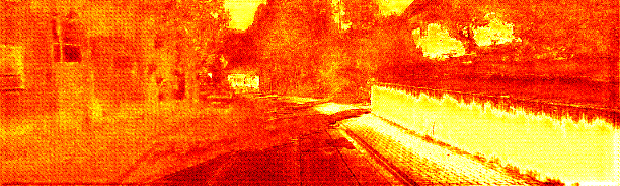}\\
	\end{minipage}\\
	\begin{minipage}{1\linewidth}
		\centering\vspace{1pt}
		\begin{minipage}{0.24\linewidth} \centering \footnotesize Input/GT(PSNR/SSIM) \end{minipage}
		\begin{minipage}{0.24\linewidth} \centering \footnotesize iPASSR(20.57/0.833) \end{minipage}
		\begin{minipage}{0.24\linewidth} \centering	\footnotesize ZeroDCE++(5.83/0.324) \end{minipage}
		\begin{minipage}{0.24\linewidth} \centering	\footnotesize DVENet(20.00/0.771) \end{minipage}
	\end{minipage}	
	
	\begin{minipage}{1\linewidth}
		\centering
		\includegraphics[width=0.24\linewidth]{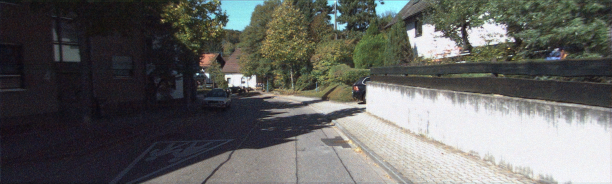}
		\includegraphics[width=0.24\linewidth]{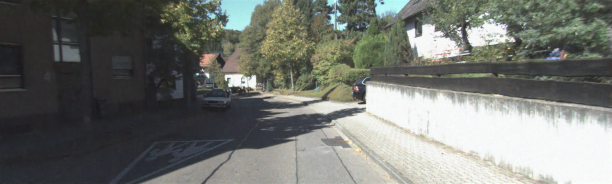}
		\includegraphics[width=0.24\linewidth]{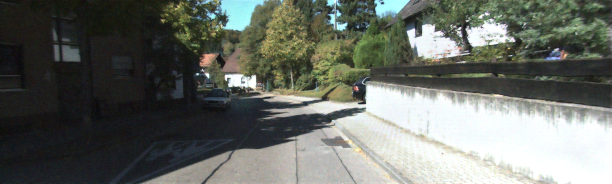}
		\includegraphics[width=0.24\linewidth]{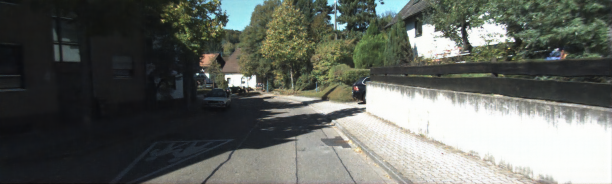}\\
		\includegraphics[width=0.24\linewidth]{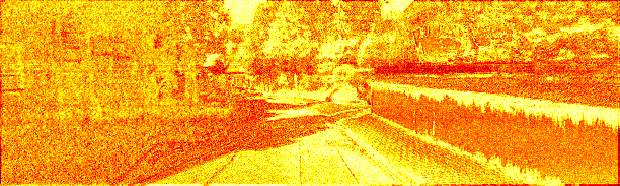}
		\includegraphics[width=0.24\linewidth]{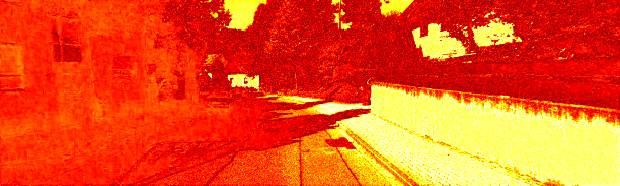}
		\includegraphics[width=0.24\linewidth]{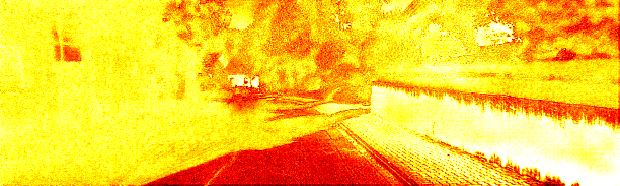}
		\includegraphics[width=0.24\linewidth]{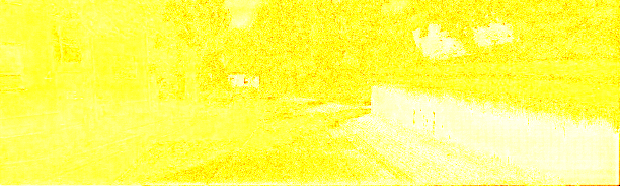}\\
	\end{minipage}\\
	\begin{minipage}{1\linewidth}
		\centering\vspace{1pt}
		\begin{minipage}{0.24\linewidth} \centering \footnotesize NAFSSR(25.58/0.856) \end{minipage}
		\begin{minipage}{0.24\linewidth} \centering \footnotesize NAFSSR-F(19.69/0.811) \end{minipage}
		\begin{minipage}{0.24\linewidth} \centering	\footnotesize SNR(23.87/0.871) \end{minipage}
		\begin{minipage}{0.24\linewidth} \centering	\footnotesize \textbf{DCI-Net}(28.02/0.932) \end{minipage}
	\end{minipage}
	
	\begin{minipage}{1\linewidth}
		\centering
		\includegraphics[width=0.24\linewidth]{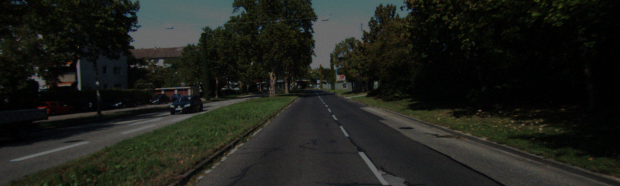}
		\includegraphics[width=0.24\linewidth]{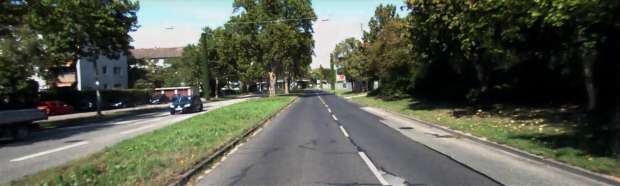}
		\includegraphics[width=0.24\linewidth]{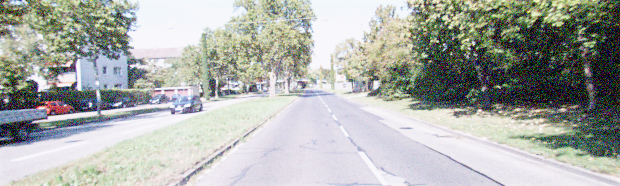}
		\includegraphics[width=0.24\linewidth]{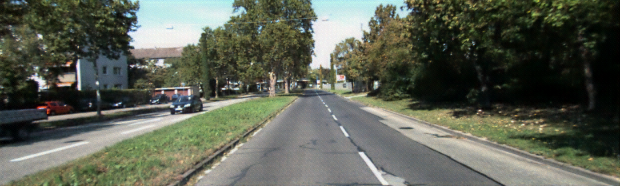}\\
		\includegraphics[width=0.24\linewidth]{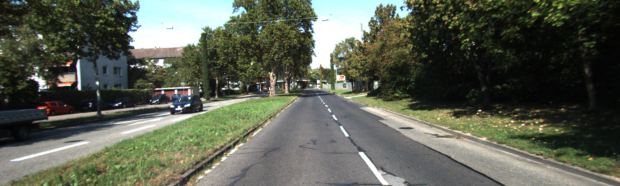}
		\includegraphics[width=0.24\linewidth]{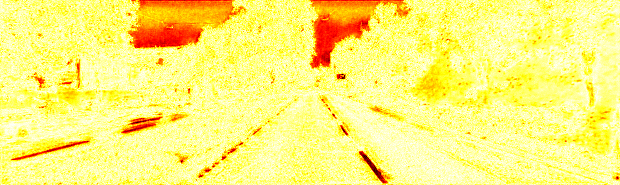}
		\includegraphics[width=0.24\linewidth]{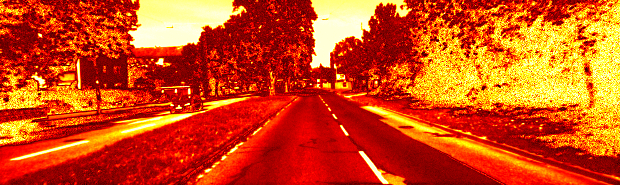}
		\includegraphics[width=0.24\linewidth]{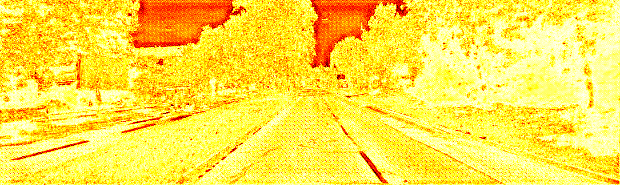}\\
	\end{minipage}\\
	\begin{minipage}{1\linewidth}
		\centering\vspace{1pt}
		\begin{minipage}{0.24\linewidth} \centering \footnotesize Input/GT(PSNR/SSIM) \end{minipage}
		\begin{minipage}{0.24\linewidth} \centering \footnotesize iPASSR(30.13/0.903) \end{minipage}
		\begin{minipage}{0.24\linewidth} \centering	\footnotesize ZeroDCE++(8.91/0.465) \end{minipage}
		\begin{minipage}{0.24\linewidth} \centering	\footnotesize DVENet(28.78/0.909) \end{minipage}
	\end{minipage}
	
	\begin{minipage}{1\linewidth}
		\centering
		\includegraphics[width=0.24\linewidth]{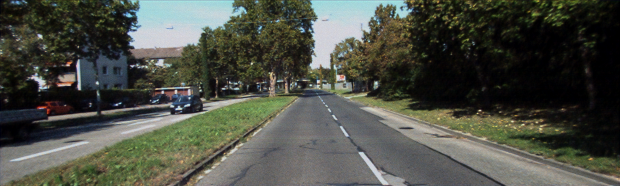}
		\includegraphics[width=0.24\linewidth]{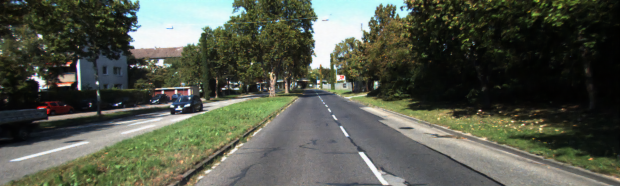}
		\includegraphics[width=0.24\linewidth]{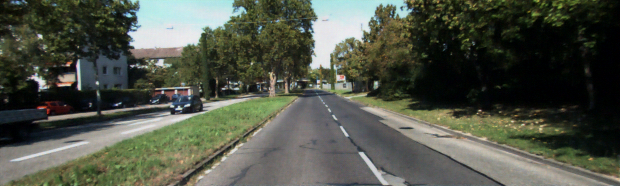}
		\includegraphics[width=0.24\linewidth]{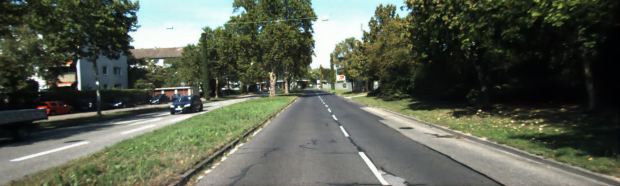}\\
		\includegraphics[width=0.24\linewidth]{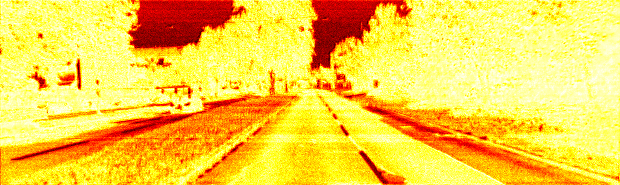}
		\includegraphics[width=0.24\linewidth]{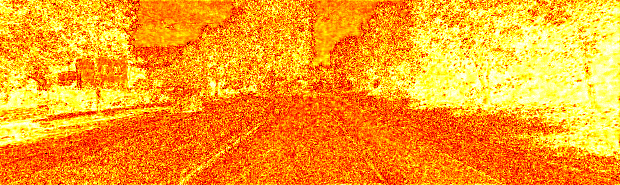}
		\includegraphics[width=0.24\linewidth]{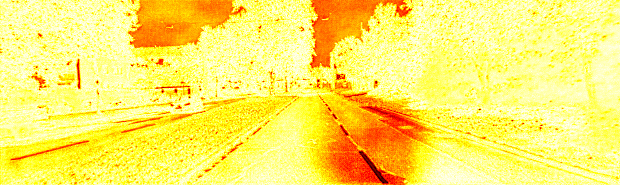}
		\includegraphics[width=0.24\linewidth]{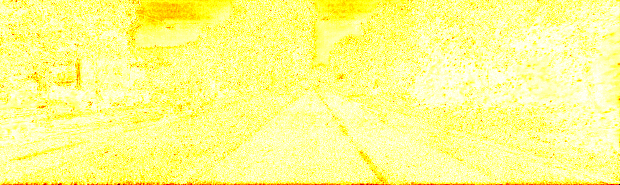}\\
	\end{minipage}\\
	\begin{minipage}{1\linewidth}
		\centering\vspace{1pt}
		\begin{minipage}{0.24\linewidth} \centering \footnotesize NAFSSR(23.11/0.892) \end{minipage}
		\begin{minipage}{0.24\linewidth} \centering \footnotesize NAFSSR-F(30.41/0.956) \end{minipage}
		\begin{minipage}{0.24\linewidth} \centering	\footnotesize SNR(26.06/0.888) \end{minipage}
		\begin{minipage}{0.24\linewidth} \centering	\footnotesize \textbf{DCI-Net}(35.09/0.959) \end{minipage}
	\end{minipage}
	\vspace{-6mm}
	\caption{Visualization of the enhanced images and corresponding error maps of each method based on the KITTI 2012 and KITTI 2015 datasets, including NAFSSR \cite{Chu2022NAFSSRSI}, NAFSSR-F \cite{Chu2022NAFSSRSI}, iPASSRNet \cite{Wang2021SymmetricPA}, SNR \cite{Xu2022SNRAwareLI}, DVENet \cite{Huang2022LowLightSI}, ZeroDCE++ \cite{li2021learning} and our DCI-Net.  Whiter and brighter pixels in the error maps indicate smaller errors. Compared with other methods, our DCI-Net obtains the smallest errors as can been seen from the error maps. This indicates that our proposed DCI-Net can better adjust the illumination and restore the details.}
	\label{fig:6}
	\vspace{-4mm}
\end{figure*}

\textbf{Long-range dependency capture (LRDC)}. To overcome the shortage that CNNs with small kernel size cannot build long-range relationship, we use large kernel convolutional layers for long-range dependency capture. The shifted kernel in CNN can be regarded as the shifted window in Swintransformer \cite{liu2021swin}. Nevertheless, it is computationally expensive for the vanilla CNN with increased kennel size. Hence, large kernel depth-wise convlolutional (DW-Conv) layer and MLP are used to approach the effect of vanilla CNN. There are two advantages for this design. Firstly, large kernel design can discover long-range correlations; secondly, DW-Conv layer can significantly reduce the computational cost. We set the kernel size of DW-Conv layer to $7$ that is consistent with the window size of Swintransformer \cite{liu2021swin}. From Fig. \ref{fig:2}, the transformation of LRDC can be described as follows:
\begin{equation}
	F_{lr} = {\rm MLP(DWConv(MLP(LN(}F{\rm ))))} + F,
\end{equation}
where ${\rm LN}(\cdot)$, ${\rm MLP}(\cdot)$ and ${\rm DWConv}(\cdot)$ denote the layer normalization (LN), MLP and DW-Conv, $F$ and $F_{lr}$ represent the input and output feature maps. 

\textbf{Expanded channel information refinement (ECIR)}. In the first step of SIMB, we mainly explore spatial information. But there less attention has been paid to channel information. Hence, we develop ECIR. The core idea of ECIR is simple yet effective, which completes channel information mixing in higher dimensional space. Note that this design can be easily implemented by incorporating channel attention (CA) into the second stage of vision transformer. As shown in Fig. \ref{fig:2}, the process of ECIR can be formulated by
\begin{equation}
	F_{sb} = {\rm MLP(CA(MLP(LN(}F_{lr}{\rm ))))} + F_{lr},
\end{equation}
where $\rm CA(\cdot)$ denotes the transformation of channel attention and $F_{sb}$ is the processed result of SIMB. It should be noted that the first MLP is used to expand the channel of feature maps to the higher dimension, while the second MLP re-maps the higher dimensional feature maps to the original channel size. Since CA can compress the spatial information and fully focuses on the channel information, it can be used to refine the channel information.

\subsection{Loss function}
The total loss function $\mathcal{L}$  of our DCI-Net contains two losses, i.e., frequency-domain reconstruction loss $\mathcal{L}_{fre}$ and smooth loss $\mathcal{L}_{tv}$, which are illustrated as follows:
\begin{equation}
	\mathcal{L} = \mathcal{L}_{fre} + \lambda \mathcal{L}_{tv}, 
\end{equation}
where $\lambda$ is a hyper-parameter that is set to 0.1 in this paper. To be specific, frequency-domain reconstruction loss is used to guide our method to reconstruct the normal-light stereo image, which can be described as
\begin{equation}
\begin{aligned}
	\mathcal{L}_{fre} = &\Vert {\rm FFT(}I^{L}_{high}{)} - {\rm FFT(}I^{L}_{gt}{)} \Vert_1 + \\
					  & \Vert {\rm FFT(}I^{R}_{high}{)} - {\rm FFT(}I^{R}_{gt}{)} \Vert_1,
\end{aligned}
\end{equation}
where $I^{L}_{gt}$ and $I^{R}_{gt}$ denote the ground-truth stereo normal-light images, and $\rm FFT(\cdot)$ denotes the fast fourier transform. The smooth loss is based on the total variation prior to obtain better and smoother results.

\section{Experimental Results and Analysis}
In this section, we first introduce the experimental setting. Then, the experimental results and detailed analysis of our DCI-Net will be illustrated.

\begin{figure*}[t]
	\centering  
	\rotatebox{90}{\scriptsize{RGB Image (\textcolor{magenta}{Left})}}\hspace{-2mm}
	\subfloat[]{
		\includegraphics[width=0.155\linewidth,height=0.11\linewidth, frame]{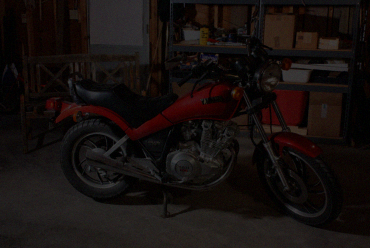}\hspace{-1mm}
	}
	\subfloat[]{
		\includegraphics[width=0.155\linewidth,height=0.11\linewidth, frame]{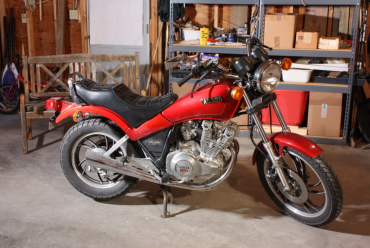}\hspace{-1mm}
	}
	\subfloat[]{
		\includegraphics[width=0.155\linewidth,height=0.11\linewidth, frame]{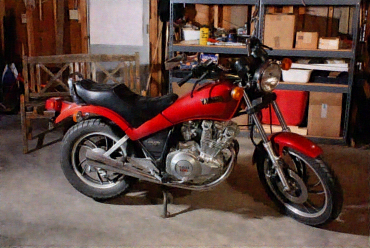}\hspace{-1mm}
	}
	\subfloat[]{
		\includegraphics[width=0.155\linewidth,height=0.11\linewidth, frame]{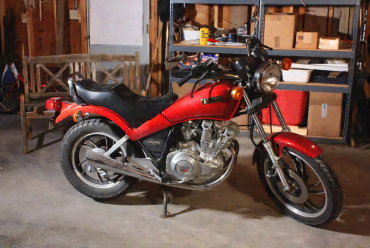}\hspace{-1mm}
	}
	\subfloat[]{
		\includegraphics[width=0.155\linewidth,height=0.11\linewidth, frame]{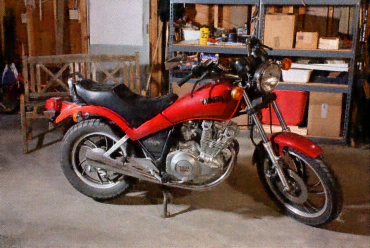}\hspace{-1mm}
	}
	\subfloat[]{
		\includegraphics[width=0.155\linewidth,height=0.11\linewidth, frame]{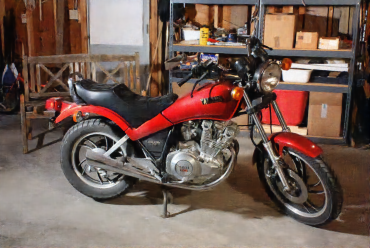}
	}
	\\\vspace{-4mm}
	\rotatebox{90}{\scriptsize{~Error Map (\textcolor{magenta}{Left})}}\hspace{-2mm}
	\subfloat[\scriptsize Input(PSNR/SSIM)]{
		\includegraphics[width=0.155\linewidth,height=0.11\linewidth, frame]{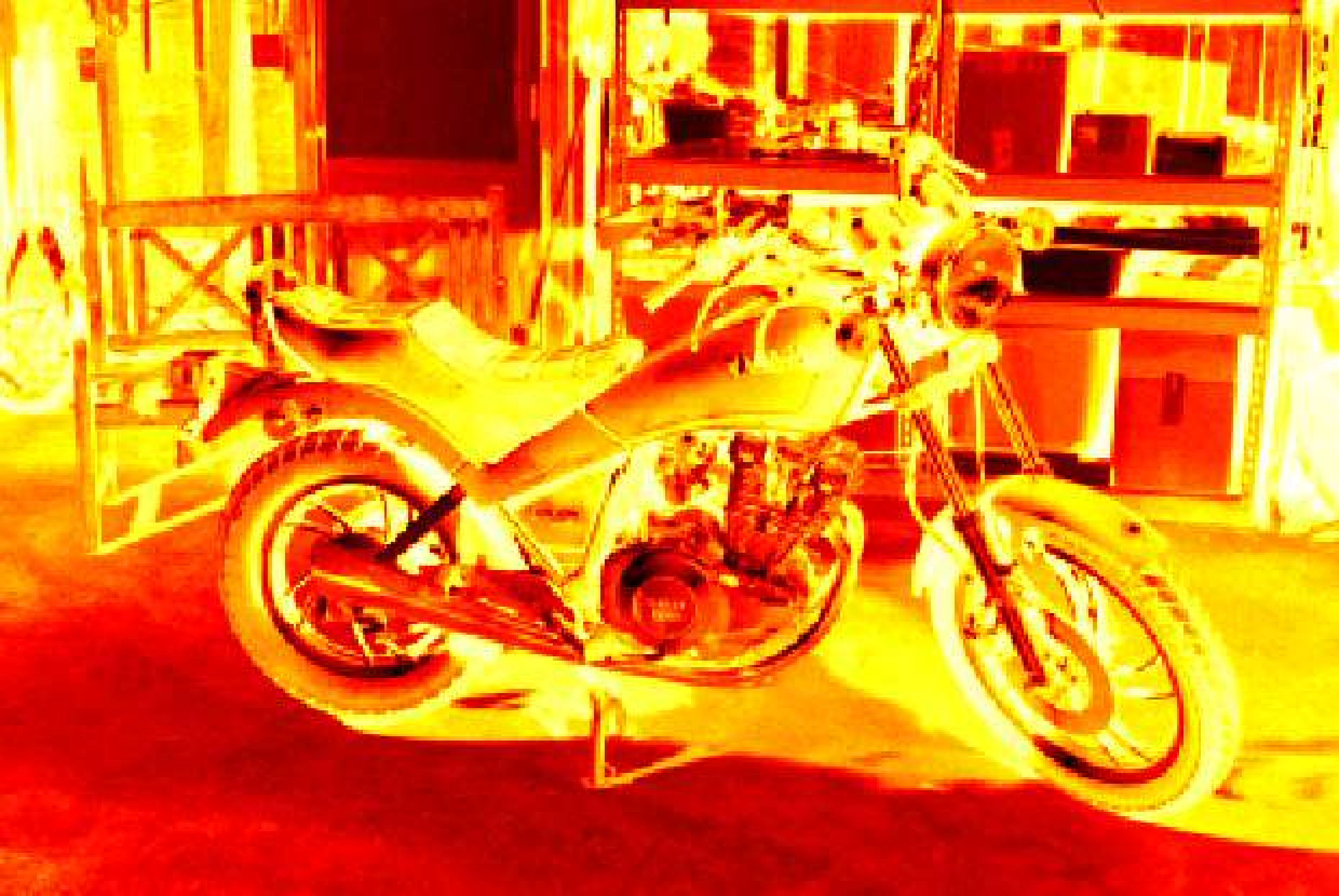}\hspace{-1mm}
	}
	\subfloat[\scriptsize GT($+\infty$/1.000)]{
		\includegraphics[width=0.155\linewidth,height=0.11\linewidth, frame]{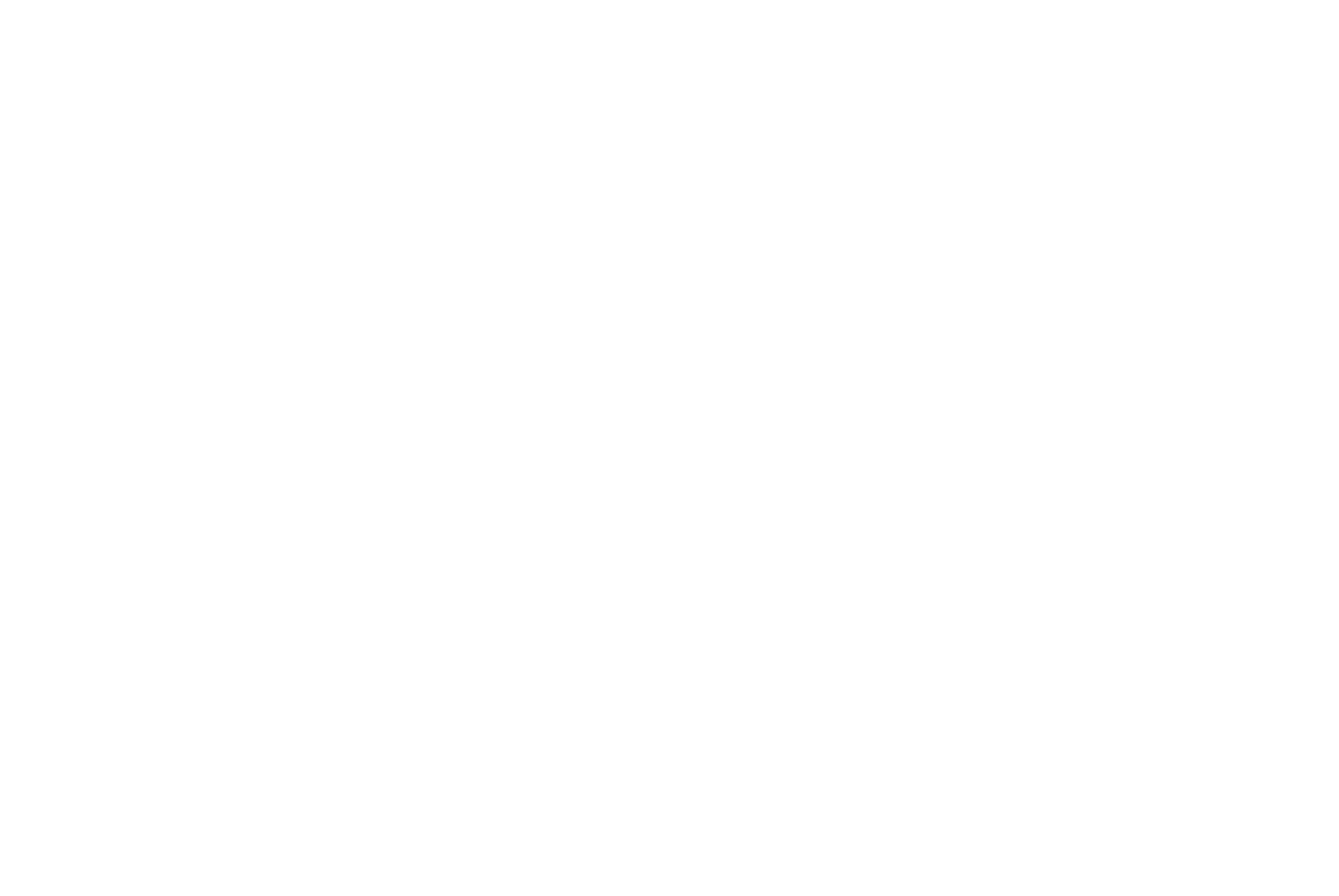}\hspace{-1mm}
	}
	\subfloat[\scriptsize SNR(19.00/0.796)]{
		\includegraphics[width=0.155\linewidth,height=0.11\linewidth, frame]{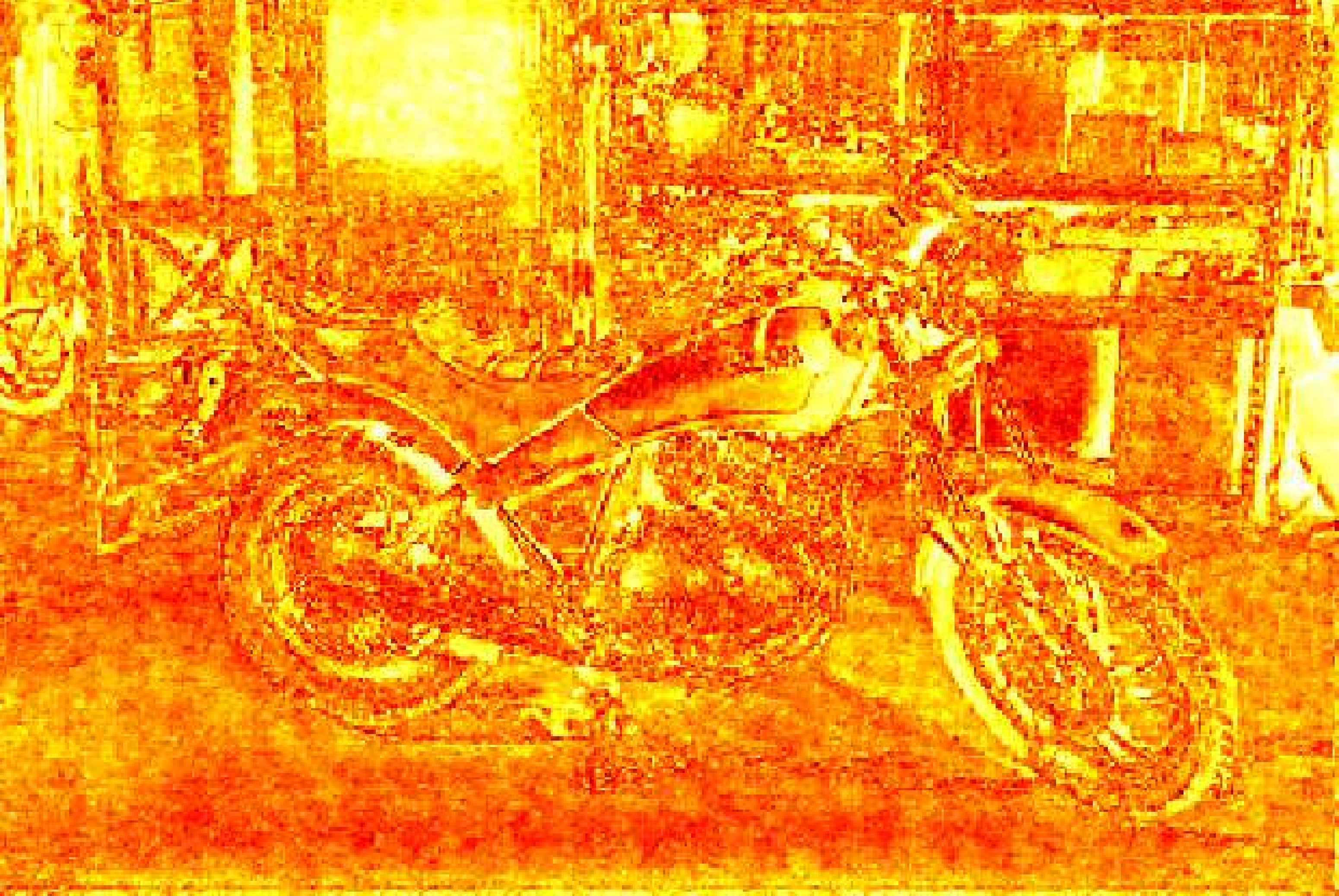}\hspace{-1mm}
	}
	\subfloat[\scriptsize NAFSSR-F(23.37/0.910)]{
		\includegraphics[width=0.155\linewidth,height=0.11\linewidth, frame]{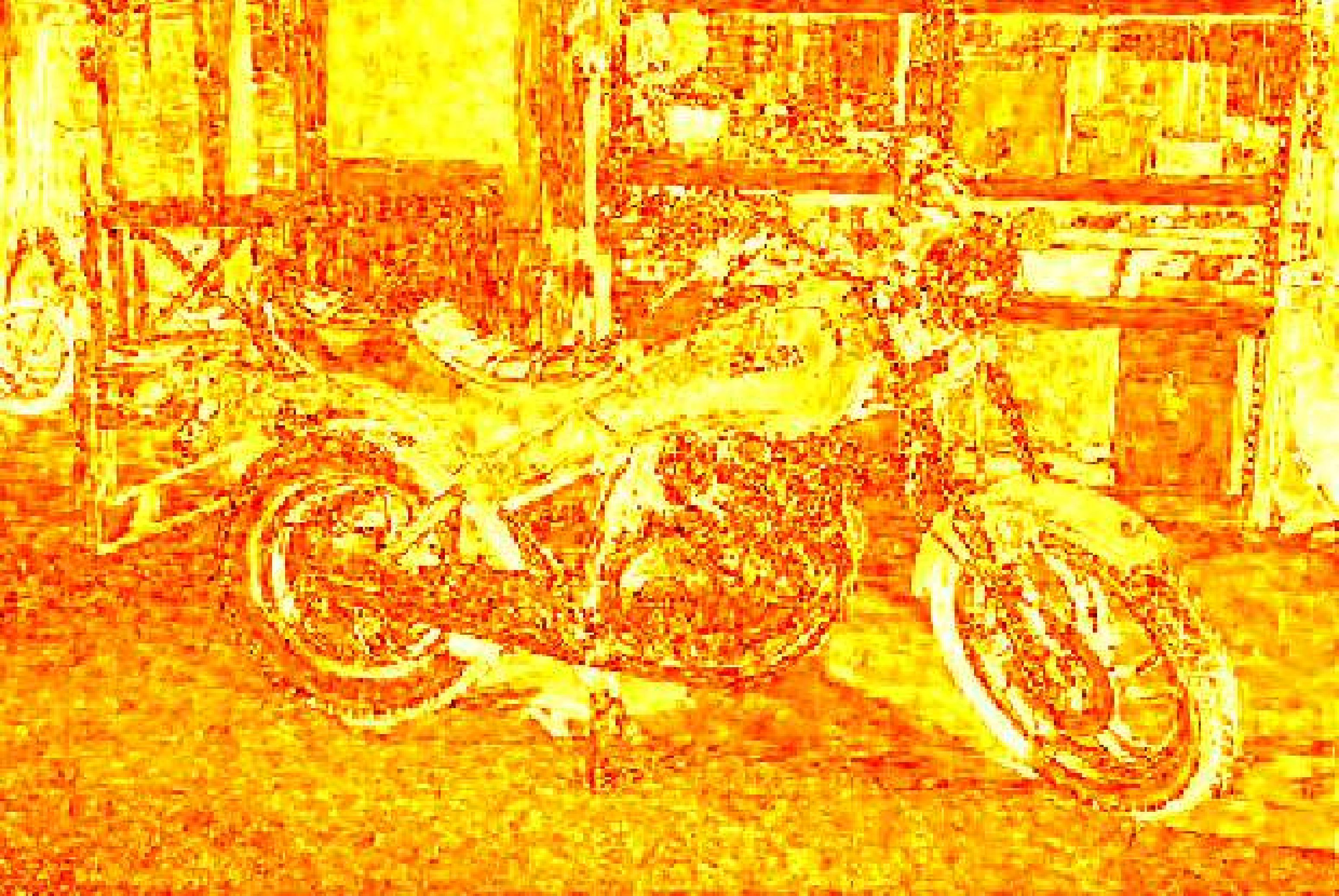}\hspace{-1mm}
	}
	\subfloat[\scriptsize DVENet(23.45/0.862)]{
		\includegraphics[width=0.155\linewidth,height=0.11\linewidth, frame]{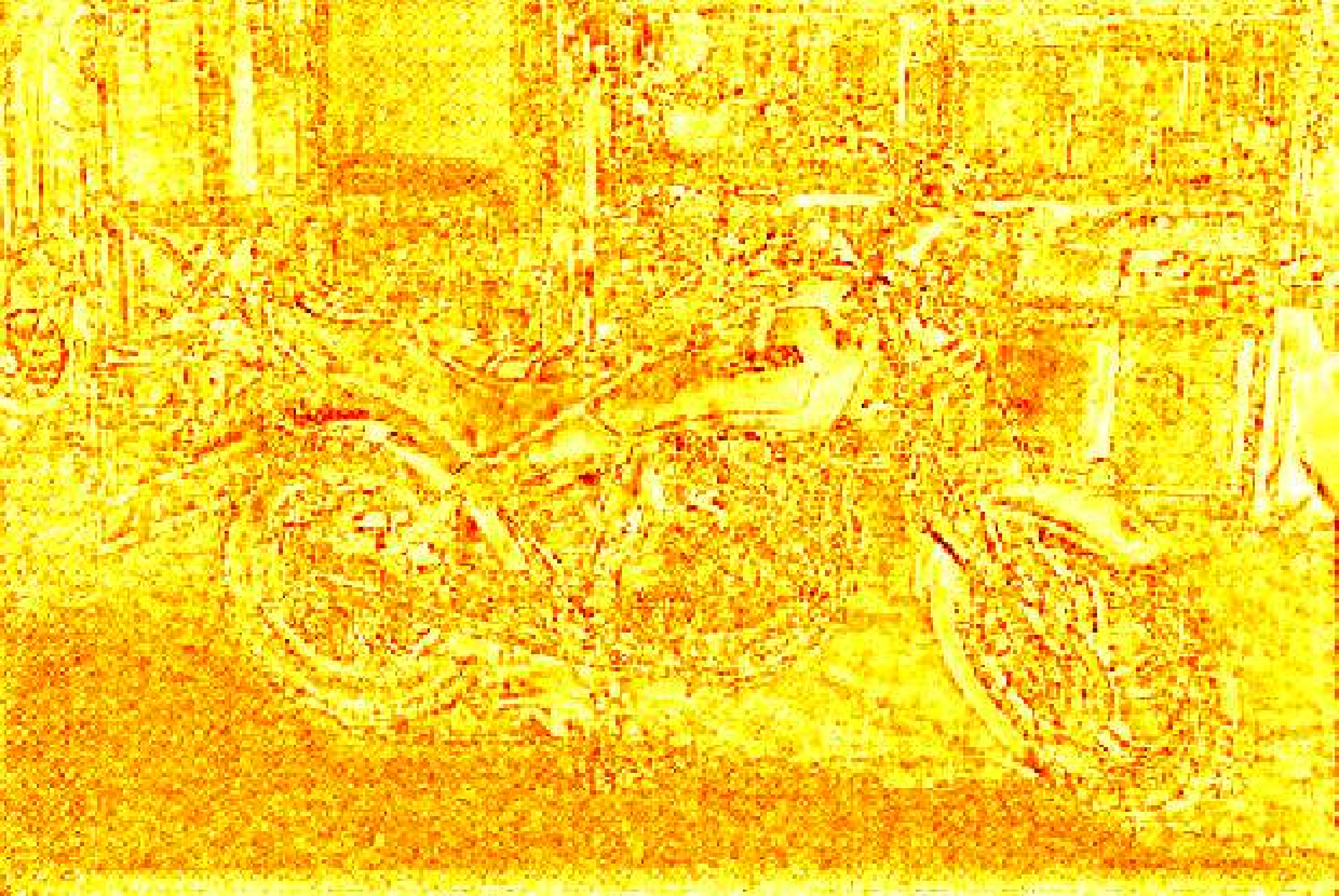}\hspace{-1mm}
	}
	\subfloat[\scriptsize \textbf{DCI-Net}(27.06/0.927)]{
		\includegraphics[width=0.155\linewidth,height=0.11\linewidth, frame]{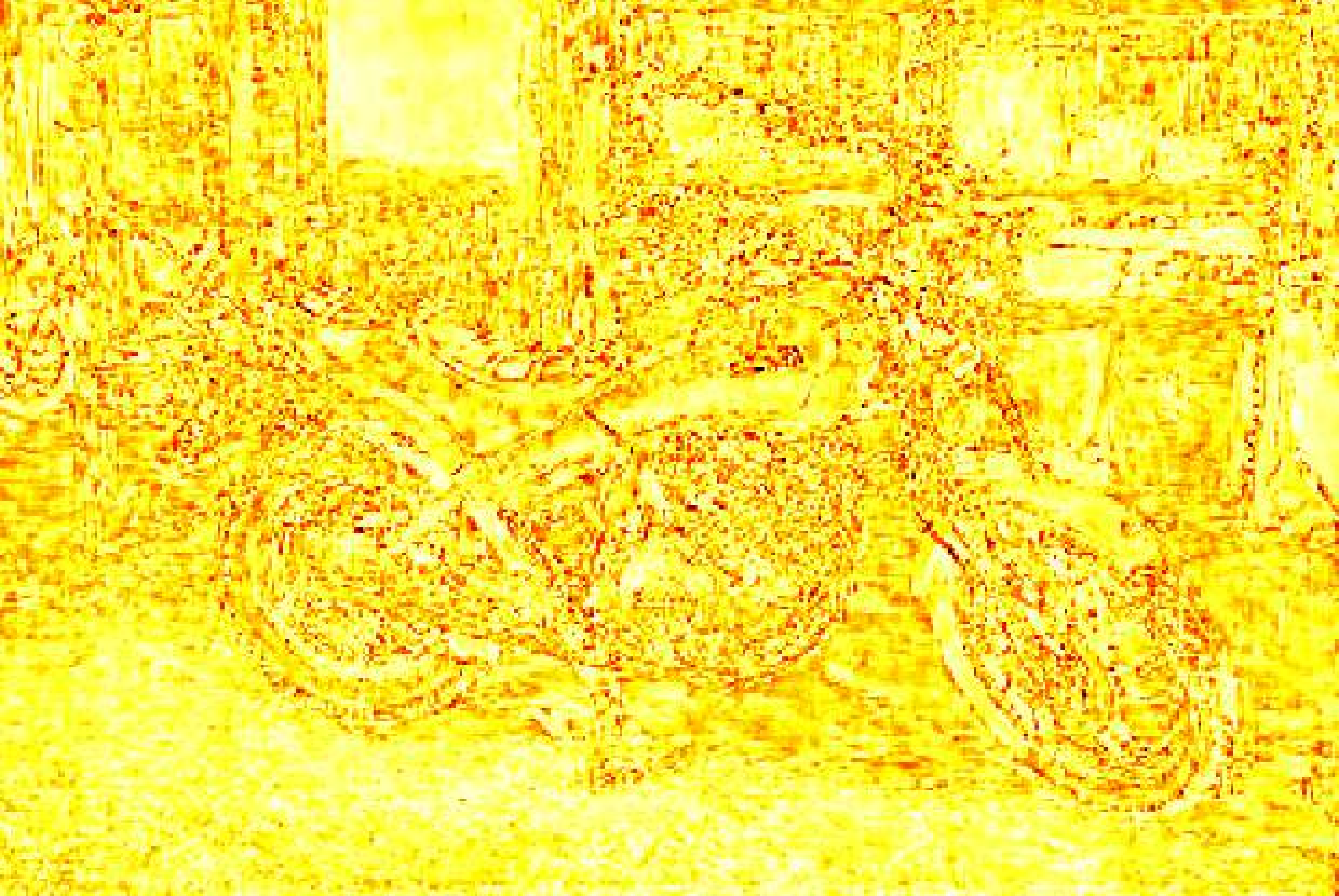}
	}\\
	\rotatebox{90}{\scriptsize{RGB Image (\textcolor{blue}{Right})}}\hspace{-2mm}
	\subfloat[]{
		\includegraphics[width=0.155\linewidth,height=0.11\linewidth, frame]{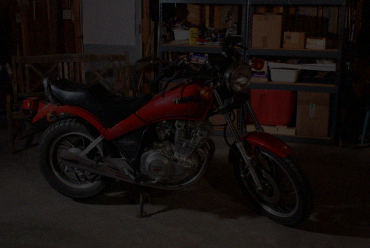}\hspace{-1mm}
	}
	\subfloat[]{
		\includegraphics[width=0.155\linewidth,height=0.11\linewidth, frame]{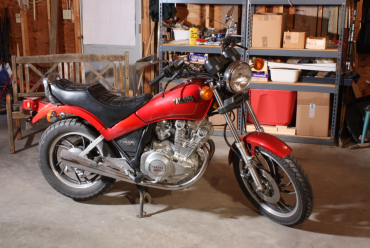}\hspace{-1mm}
	}
	\subfloat[]{
		\includegraphics[width=0.155\linewidth,height=0.11\linewidth, frame]{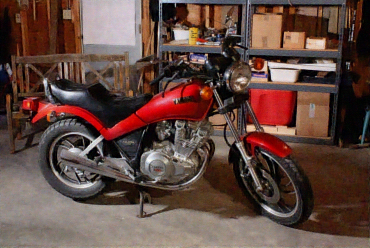}\hspace{-1mm}
	}
	\subfloat[]{
		\includegraphics[width=0.155\linewidth,height=0.11\linewidth, frame]{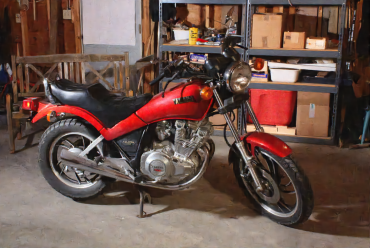}\hspace{-1mm}
	}
	\subfloat[]{
		\includegraphics[width=0.155\linewidth,height=0.11\linewidth, frame]{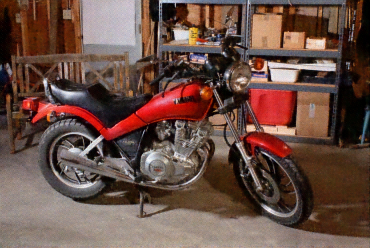}\hspace{-1mm}
	}
	\subfloat[]{
		\includegraphics[width=0.155\linewidth,height=0.11\linewidth, frame]{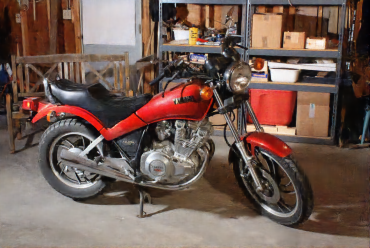}
	}
	\\\vspace{-4mm}
	\rotatebox{90}{\scriptsize{~Error Map (\textcolor{blue}{Right})}}\hspace{-2mm}
	\subfloat[\scriptsize Input(PSNR/SSIM)]{
		\includegraphics[width=0.155\linewidth,height=0.11\linewidth, frame]{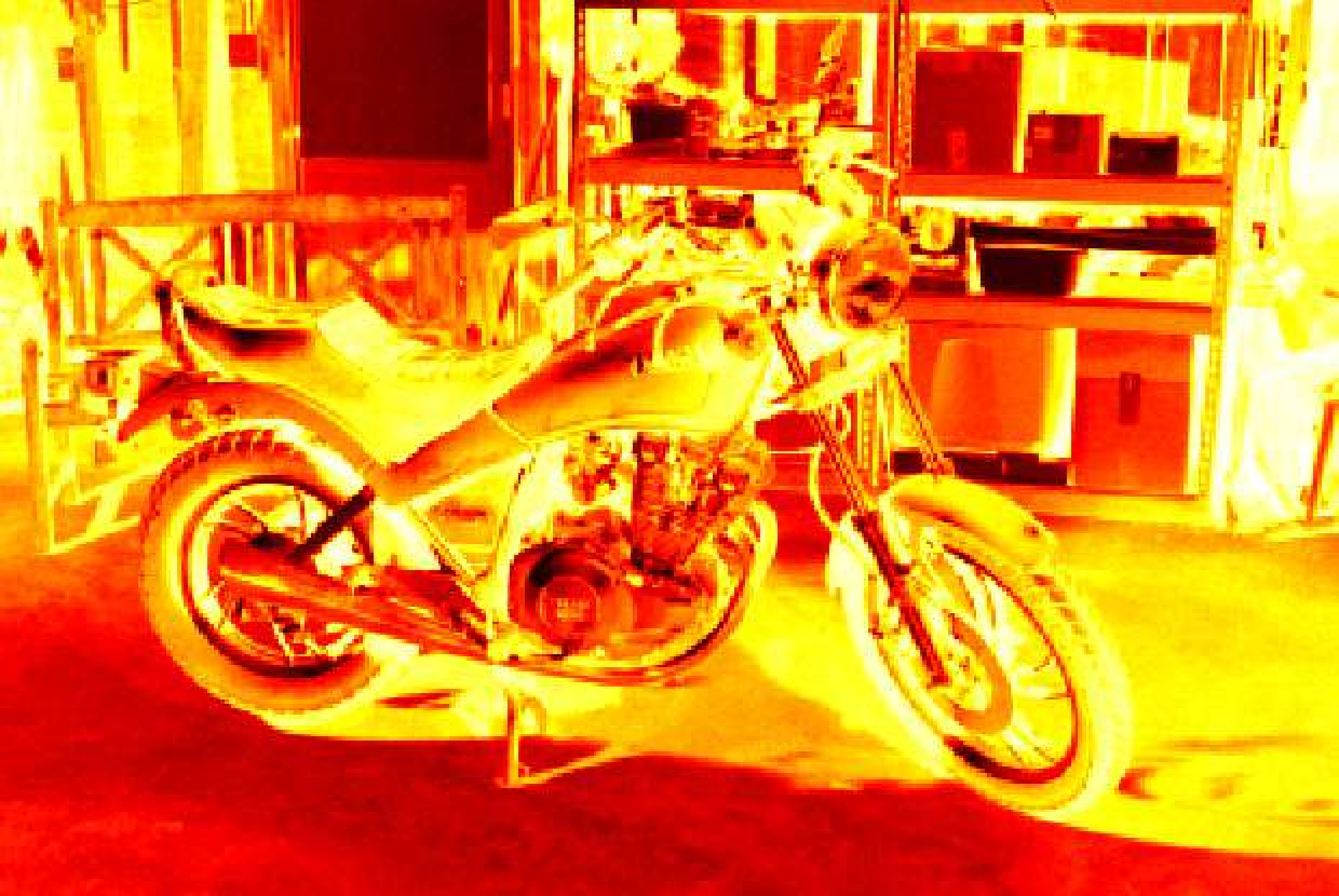}\hspace{-1mm}
	}
	\subfloat[\scriptsize GT($+\infty$/1.000)]{
		\includegraphics[width=0.155\linewidth,height=0.11\linewidth, frame]{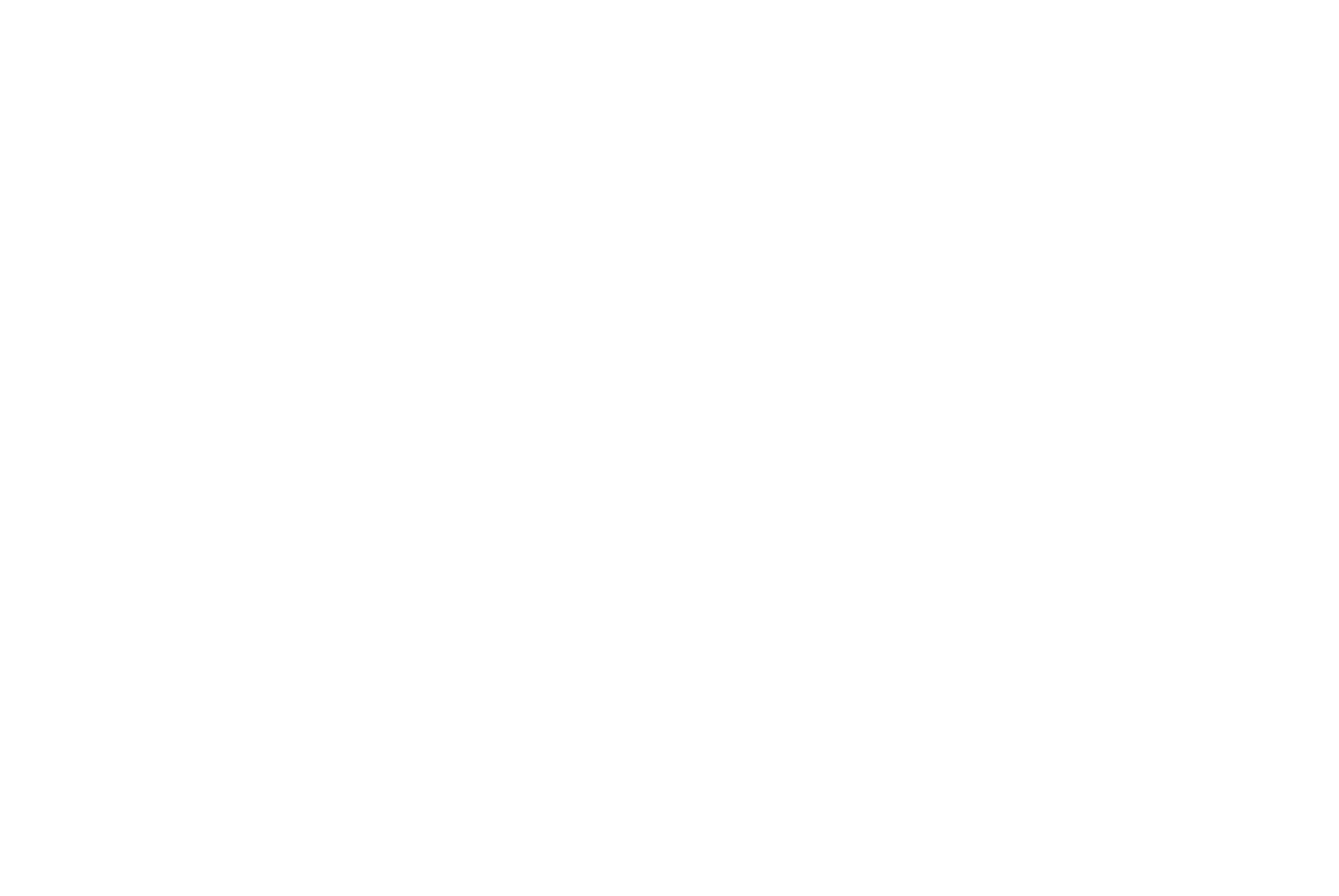}\hspace{-1mm}
	}
	\subfloat[\scriptsize SNR(19.69/0.812)]{
		\includegraphics[width=0.155\linewidth,height=0.11\linewidth, frame]{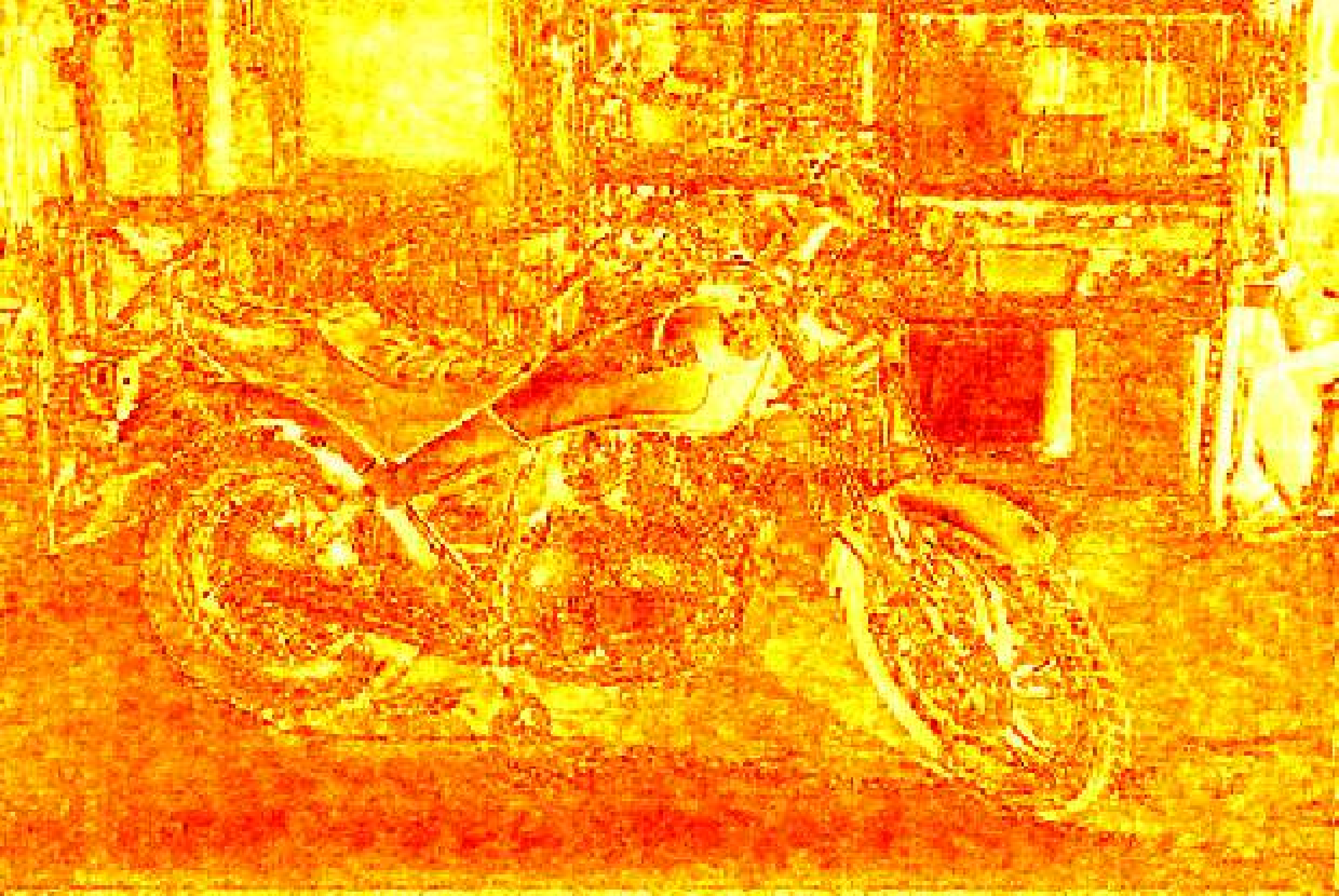}\hspace{-1mm}
	}
	\subfloat[\scriptsize NAFSSR-F(23.65/0.910)]{
		\includegraphics[width=0.155\linewidth,height=0.11\linewidth, frame]{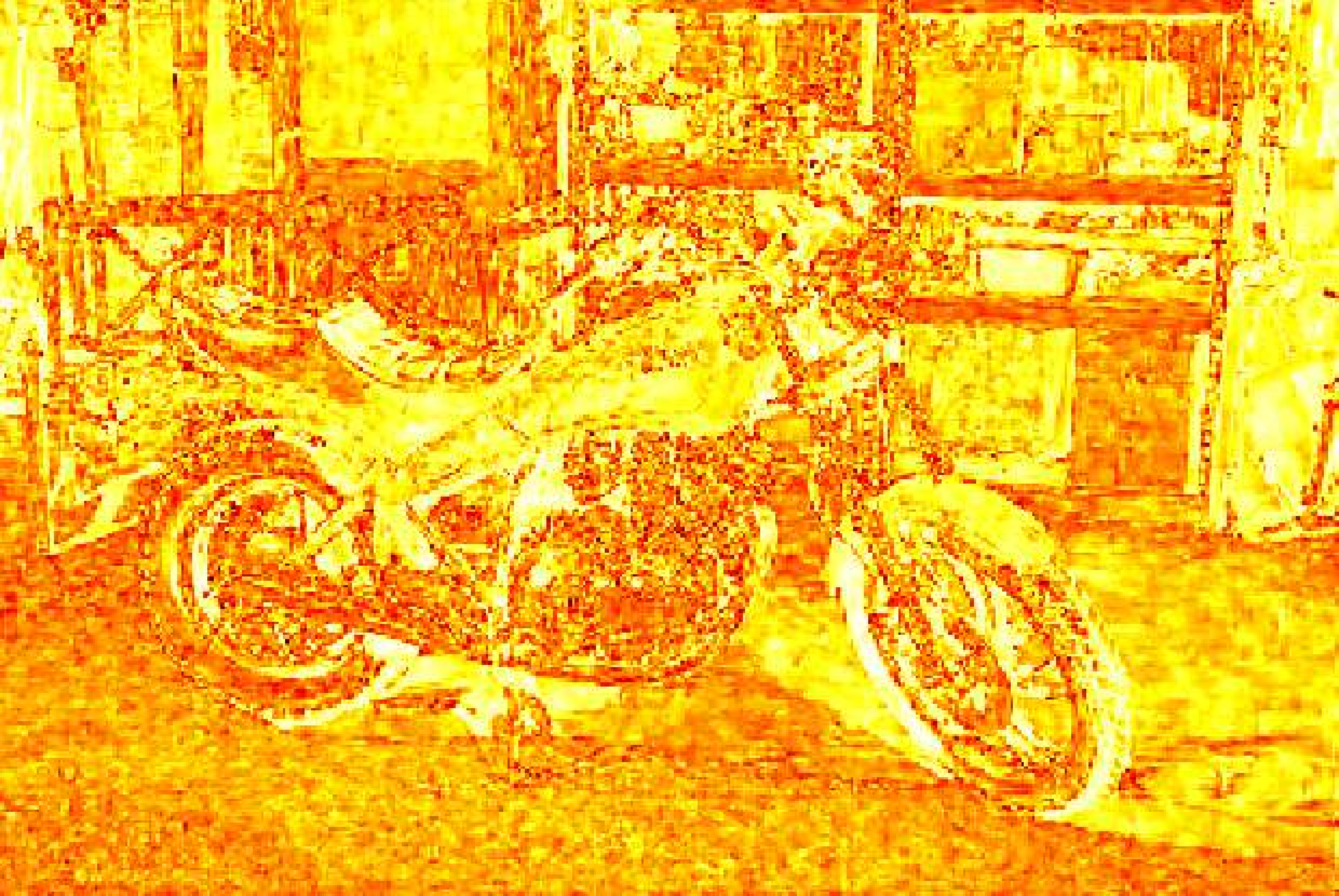}\hspace{-1mm}
	}
	\subfloat[\scriptsize DVENet(22.05/0.845)]{
		\includegraphics[width=0.155\linewidth,height=0.11\linewidth, frame]{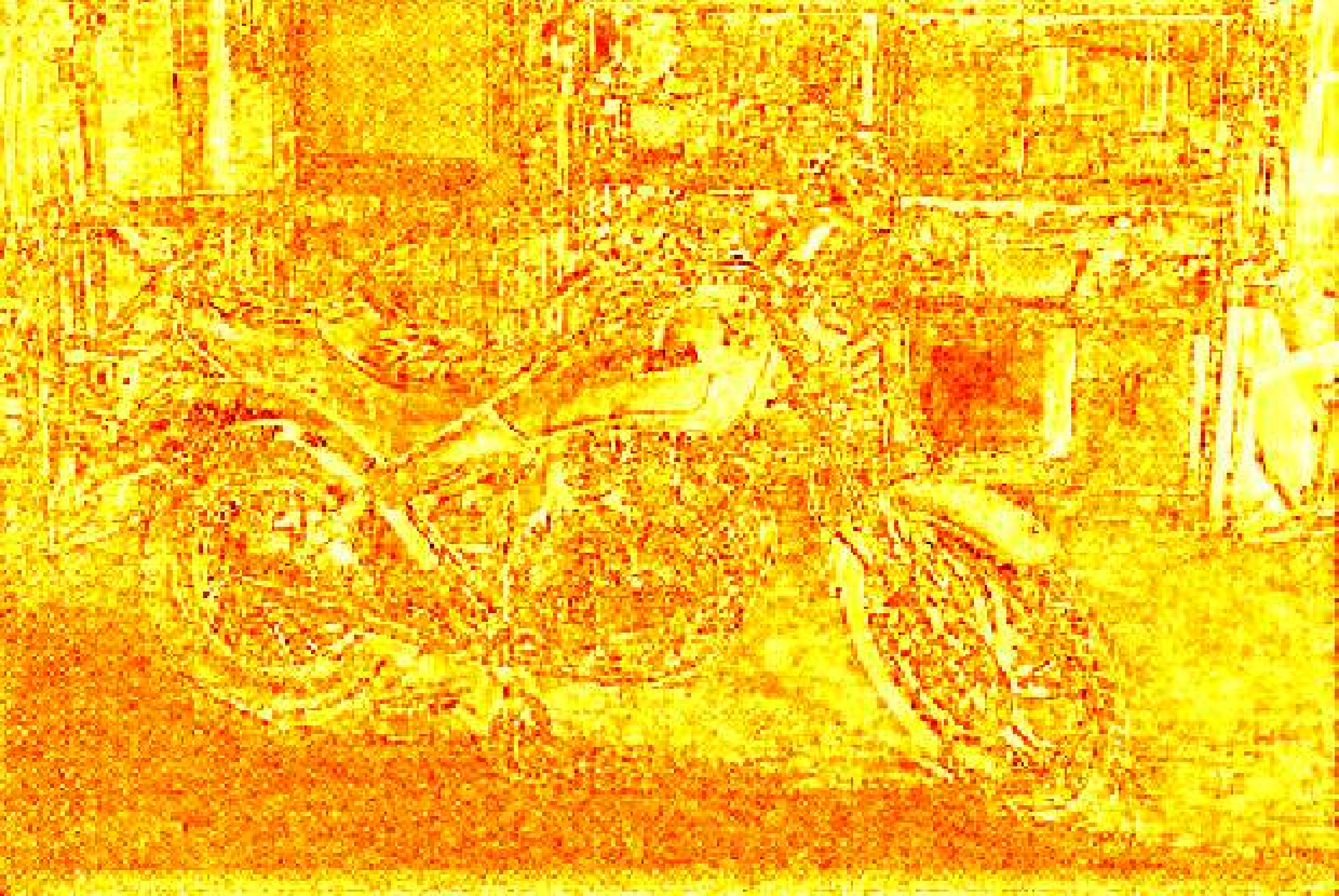}\hspace{-1mm}
	}
	\subfloat[\scriptsize \textbf{DCI-Net}(28.30/0.931)]{
		\includegraphics[width=0.155\linewidth,height=0.11\linewidth, frame]{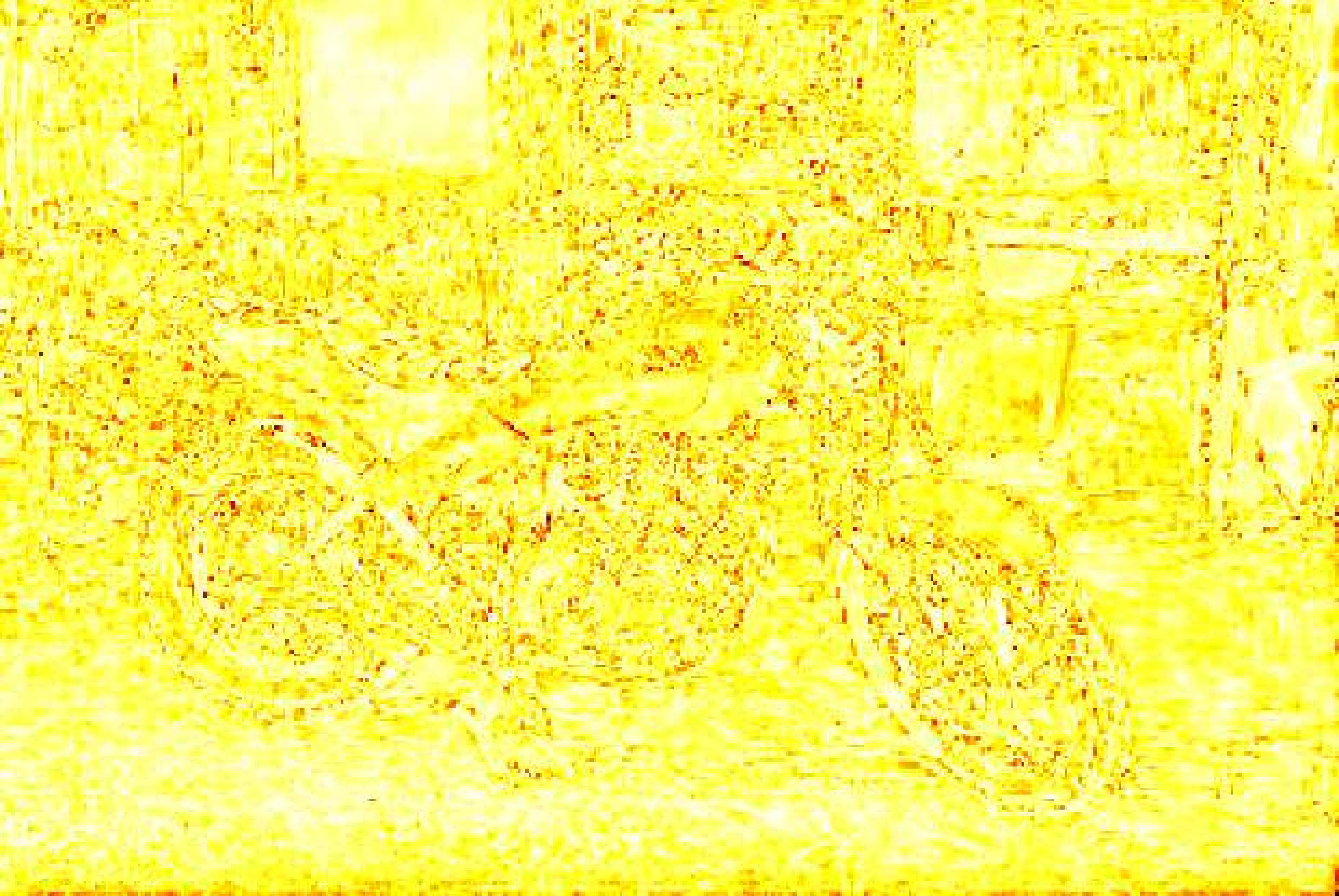}
	}\\\vspace{-2mm}
	\caption{Visual results of each method based on Middlebury dataset, including NAFSSR-F \cite{Chu2022NAFSSRSI}, SNR \cite{Xu2022SNRAwareLI}, DVENet \cite{Huang2022LowLightSI} and our DCI-Net. Whiter and brighter pixels in the error maps indicate smaller errors. We can see that there is a motorcycle in the error maps of all compared methods. But in contrast, it is hard to see the shape of the motorcycle for our DCI-Net, i.e., less information is lost in our method.}
	\label{fig:7}
	\vspace{-3mm}
\end{figure*}

\begin{table*}[t]
	\vspace{2.5mm}
	\centering
	\caption{Evaluation results of each method on Flickr1024, KITTI 2012, KITTI 2015 and Middlebury datasets. Note that PSNR/SSIM values achieved on both the left images (i.e., Left) and a pair of stereo images (i.e., (Left + Right) /2) are reported. The \textbf{bold} denotes the best. It is clear that our DCI-Net achieves SOTA performance among all compared methods.} 
	\vspace{-2.5mm}
	\footnotesize
	\setlength\tabcolsep{5pt} 
	\begin{tabular}{cccccccccc}
		\toprule[1pt]
		\multirow{2.5}{*}{Method} & \multirow{2.5}{*}{Venue} & \multicolumn{4}{c}{Left} & \multicolumn{4}{c}{(Left + Right) / 2} \\
		\cmidrule(lr){3-6}\cmidrule(lr){7-10}
		& & Flickr1024 & KITTI 2012 & KITTI 2015 & Middlebury & Flickr1024 & KITTI 2012 & KITTI 2015 & Middlebury\\
		\midrule
		ZeroDCE \cite{guo2020zero}           & cvpr'20  & 10.49/0.420 & 9.21/0.403 & 10.17/0.441 & 11.34/0.591 & 10.49/0.420 & 9.18/0.401 & 10.17/0.439 & 11.27/0.592\\
		iPASSRNet \cite{Wang2021SymmetricPA} & cvprw'21 & 22.27/0.777 & 23.32/0.820 & 21.76/0.802 & 21.50/0.872 & 22.20/0.777 & 23.49/0.827 & 21.69/0.804 & 21.67/0.871 \\
		ZeroDCE++ \cite{li2021learning}      & tpami'21 & 10.74/0.418 & 9.15/0.388 & 10.08/0.427 & 10.28/0.563 & 10.74/0.418 & 9.14/0.386 & 10.07/0.425 & 10.19/0.563 \\
		DVENet \cite{Huang2022LowLightSI}    & tmm'22   & 21.82/0.751 & 22.14/0.787 & 21.05/0.767 & 20.75/0.855 & 21.66/0.751 & 21.96/0.787 & 21.05/0.769 & 21.09/0.856 \\
		NAFSSR \cite{Chu2022NAFSSRSI}        & cvprw'22 & 21.68/0.734 & 21.43/0.775 & 20.68/0.767 & 22.25/0.857 & 21.74/0.735 & 21.70/0.778 & 20.64/0.771 & 22.39/0.858 \\
		NAFSSR-F \cite{Chu2022NAFSSRSI}      & cvprw'22 & 24.21/0.815 & 24.22/0.856 & 22.29/0.837 & 23.49/0.920 & 24.26/0.816 & 23.93/0.854 & 22.45/0.842 & 23.43/0.921 \\
		SNR \cite{Xu2022SNRAwareLI}          & cvpr'22  & 21.34/0.749 & 22.11/0.794 & 20.93/0.769 & 21.43/0.851 & 21.40/0.751 & 22.16/0.793 & 20.90/0.770 & 21.65/0.856 \\
		
		\midrule
		DCI-Net   & - & \textbf{25.47}/\textbf{0.832} & \textbf{26.69}/\textbf{0.883} & \textbf{26.26}/\textbf{0.862} & \textbf{24.27}/\textbf{0.931} & \textbf{25.36}/\textbf{0.832} & \textbf{26.90}/\textbf{0.884} & \textbf{26.03}/\textbf{0.864} & \textbf{24.26}/\textbf{0.929}\\
		\bottomrule[1pt]
	\end{tabular}	
	\vspace{-3mm}	
	\label{table:1}
\end{table*}

\subsection{Experimental setting}
\textbf{Datasets and evaluation metrics}. Following the setting of iPASSRNet \cite{Wang2021SymmetricPA}, we use 60 stereo image pairs from Middlebury \cite{scharstein2014high} and 800 stereo image pairs from Flickr1024 \cite{Wang2019LearningPA} as the training set. For testing set, we pick 112 stereo image pairs from Flickr1024, 20 stereo image pairs from KITTI 2015 \cite{menze2015object}, 20 stereo image pairs from KITTI 2012 \cite{geiger2012we} and 5 stereo image pairs from Middlebury. Note that we clearly follow \cite{zhang2021learning} to synthesize the low-light stereo images. For evaluations, we employ two widely used image quality assessment metrics: PSNR and SSIM. The greater PSNR and SSIM, the better quality of the enhanced results. As for the structure of our DCI-Net, we set $N_1 = 4$, $N_2 = 4$, $N_3 = 2$, $N_4 = 2$, and $N_5 = 4$.

\textbf{Implementation details}. All experiments are conducted by using Pytorch with RTX 2080Ti GPUs. We employ Adam optimizer with batch size of 16. All the training images are randomly cropped into $128\times 128$ pixels. The initial learning rate is set to 0.0002 and reduced by half per 500 epochs. DCI-Net is trained for 2000 epochs totally. Some related methods are compared, including NAFSSR \cite{Chu2022NAFSSRSI}, NAFSSR-F \cite{Chu2022NAFSSRSI}, iPASSRNet \cite{Wang2021SymmetricPA}, SNR \cite{Xu2022SNRAwareLI}, DVENet \cite{Huang2022LowLightSI}, ZeroDCE \cite{guo2020zero} and ZeroDCE++ \cite{li2021learning}. For iPASSRNet, NAFSSR and NAFSSR-F, we set the upscale factor to 1. Note that we use frequency-domain reconstruction loss to retrain NAFSSR, which is denoted as NAFSSR-F. 

\subsection{Quantitative results}
We evaluate the performance of our DCI-Net on four datasets, i.e., Flickr1024, KITTI 2012, KITTI 2015 and Middlebury datasets. Table \ref{table:1} shows the numerical results of both single view and a pair of stereo images. We see that our DCI-Net obtains better performance on all evaluated datasets than other compared methods. Overall, the two zero-shot methods ZeroDCE and ZeroDCE++ obtain the worst results, since no supervised data drives the training process. For supervised single low-light image enhancement method, SNR is inferior to the stereo image restoration methods, because single LLIE methods do not consider the cross-view cues. To be specific, greater SSIM values suggests that our DCI-Net can better recover the structural information, compared with other competitors. The best PSNR demonstrates that our DCI-Net is capable of restoring the details to reconstruct the normal-light stereo images.

\subsection{Visual analysis results}
For better comparison, we visualize the  enhanced images and corresponding error maps of each method on the Flickr1024 dataset in Fig.\ref{fig:5}. Note that the process of computing the error maps can be referred to \cite{zheng2021windowing}. We see that our DCI-Net is capable of recovering more consistent color and accurate illumination than other competitors. Based on the error maps, DCI-Net is the lightest one, indicating that our model can recover more details. Fig.\ref{fig:6} shows the visual comparison on KITTI 2012 and KITTI 2015 datasets. According to the displayed PSNR and SSIM, the proposed DCI-Net obtains maximum values, which suggests that the enhanced images of our model are of higher quality. Similar results can also be found from the error maps. The errors produced by our method are obviously smaller than others. The visual results on Middlebury dataset are shown in Fig. \ref{fig:7}. We can find that our DCI-Net can better reconstruct the normal-light image with less information loss.

\subsection{Ablation study}
We perform ablation studies to demonstrate the effectiveness of the designed modules, losses and kernel sizes in our DCI-Net. The experiments are performed on Flickr1024 dataset. The numerical results are shown in Table \ref{table:2}.

\textbf{Effect of DIM}. To show the effectiveness of DIM, we design three models as shown in Table \ref{table:2}. Specifically, W/o CVI and W/o CSI denote removing CVI and CSI from our DCI-Net respectively. W/o DIM denotes removing DIM from DCI-Net, which is also regarded as removing CVI and CSI simultaneously. From Table \ref{table:2}, we see that there is significant performance reduction when we remove any of them. Because the proposed DIM has the ability to complete the cross-scale cross-view information interaction.

\textbf{Effect of SIMB}. We further verify the role of SIMB by dropping LRDC and ECIM from SIMB in our DCI-Net, which are denoted as W/o LRDC and W/o ECIM in Table \ref{table:2}. When LRDC is deleted, the numerical results decrease significantly. The performance degradation suggests that LRDC is an important component that can capture long-range dependency. Removing ECIM also makes worse performance since ECIM is able to mine the channel information and refine the feature representation.

\textbf{Effect of losses}. We evaluate the performance with different losses. As shown in Table \ref{table:2}, models W/ $\mathcal{L}_1$ loss, W/ $\mathcal{L}_2$ loss and W/ $\mathcal{L}_{ssim}$ denotes using $l_1$ loss, $l_2$ loss and $ssim$ loss to replace the frequency-domain reconstruction loss $\mathcal{L}_{fre}$ to train our model. The performance degrades for the three models, which means the frequency-domain reconstruction loss is superior to others for our model.

\textbf{Effect of kernel size}. We finally study the impact of different kernel sizes of the DW-Conv layer in SIMB on the results. We test the performance with kernel size of $3\times3$ and $5\times5$, denoted as KS $3\times3$ and KS $5\times5$. Note that the kernel size in our DCI-Net is set to $7\times7$. From Table \ref{table:2}, we see that the performance gets better with larger kernels, and our proposed DCI-Net obtains the best result.

\begin{table}[t]
	\centering
	\caption{Ablation study on the effects of losses, kernel size, DIM and SIMB over the Flickr2014 dataset. \textbf{Bold} denotes the best.} 
	\footnotesize
	\setlength\tabcolsep{8pt} 
	\begin{tabular}{ccccc}
		\toprule[1pt]
		\multirow{2.5}{*}{Model} & \multicolumn{2}{c}{Left} & \multicolumn{2}{c}{(Left + Right) / 2} \\
		\cmidrule(lr){2-3}\cmidrule(lr){4-5}
		& PSNR & SSIM & PSNR & SSIM \\
		\midrule
		W/o DIM & 24.67 & 0.809 & 24.63 & 0.811\\
		W/o CVI & 25.26 & 0.828 & 25.23 & 0.829\\
		W/o CSI & 25.07 & 0.824 & 25.08 & 0.826\\
		\midrule
		W/o LRDC & 22.07 & 0.795 & 22.13 & 0.796\\
		W/o ECIM & 25.01 & 0.817 & 25.06 & 0.818\\
		\midrule
		W/ $\mathcal{L}_1$ & 24.17 & 0.788 & 24.08 & 0.789\\
		W/ $\mathcal{L}_2$ & 21.56 & 0.655 & 21.54 & 0.655\\
		W/ $\mathcal{L}_{ssim}$ & 24.01 & 0.819 & 23.93 & 0.821\\
		\midrule
		KS of $3\times3$ & 24.91 & 0.814 & 24.85 & 0.814\\
		KS of $5\times5$  & 25.07 & 0.823 & 25.01 & 0.823\\
		\midrule
		DCI-Net & \textbf{25.47} & \textbf{0.832} & \textbf{25.36} & \textbf{0.832}\\
		\bottomrule[1pt]
	\end{tabular}		
	\label{table:2}
	\vspace{-2mm}
\end{table}

\section{Conclusion}
We explored effective strategies to address the issues of weak cross-view information interaction and lacking of long-range dependencies in intra-view learning to deal with the spatial long-range effects for stereo image enhancement in the dark. Technically, we proposed a novel decoupled cross-scale cross-view interaction network (DCI-Net). To be specific, we present a decoupled interaction module to improve the information flow via exploring cross-scale cross-view information interaction in a decoupling manner, namely, cross-view interaction at different scales and cross-scale interaction. In addition, we further propose a spatial-channel information mining block to capture long-range dependency and improve the feature representation. Extensive experiments show that our DCI-Net achieves state-of-the-art quantitative performance and obtains better visual results in terms of more accurate texture-recovery and less detail loss, compared to other related methods. In the future, designing more effective and more efficient cross-view interaction method is worth studying. Besides, incorporating the low-light stereo image enhancement into some high-level vision tasks is also an interesting future work.

\section{Acknowledgment}
This work is partially supported by the National Natural
Science Foundation of China (62072151, 61932009,
61822701, 62036010, 72004174), and the Anhui Provincial
Natural Science Fund for the Distinguished Young Scholars
(2008085J30). Zhao Zhang is the corresponding author.

{\small
\bibliographystyle{ieee_fullname}
\bibliography{egbib}
}

\end{document}